\documentclass[11pt]{article}

\usepackage[final]{acl}

\usepackage{times}
\usepackage{latexsym}
\usepackage[T1]{fontenc}
\usepackage[utf8]{inputenc}
\usepackage{microtype}
\usepackage{inconsolata}

\usepackage{tocloft}
\usepackage[toc,page,header]{appendix}
\usepackage{adjustbox}
\usepackage{minitoc}
\usepackage{graphicx}
\usepackage{wrapfig}
\usepackage{multirow}
\usepackage{makecell}
\usepackage{tabularx}
\usepackage{array}
\usepackage{amsmath,amssymb,amsfonts}
\usepackage{amsthm}
\usepackage{mathtools}
\usepackage{xcolor}
\usepackage{textcomp}
\usepackage{booktabs}
\usepackage{algorithm}
\usepackage{algpseudocode}
\usepackage{listings}
\usepackage{xspace}
\usepackage{slashed}
\usepackage{siunitx}
\usepackage{adjustbox}
\usepackage{scalerel}
\usepackage{enumitem}
\usepackage[normalem]{ulem}
\useunder{\uline}{\ul}{}
\usepackage{epigraph}
\usepackage[
  align=center,
  shadow=true,
  shadowsize=4.5pt,
  nobreak=true,
  framemethod=tikz,
  skipabove=9.5pt,
  skipbelow=9pt,
  innertopmargin=5pt,
  innerbottommargin=5pt,
  innerleftmargin=5pt,
  innerrightmargin=5pt,
  leftmargin=1.5pt,
  rightmargin=1.5pt
]{mdframed}
\usepackage{tikz}
\usetikzlibrary{shadows}
\usepackage{placeins}
\usepackage{float}
\usepackage{bbm}
\usepackage{needspace}

\usepackage{caption}
\usepackage[pagebackref=true,breaklinks=true,colorlinks=true,bookmarks=false]{hyperref}
\definecolor{deepred}{HTML}{940000}
\hypersetup{linkcolor=deepred}
\hypersetup{urlcolor  = [rgb]{0.4,0.15,0.95}}
\hypersetup{citecolor=[rgb]{0.4,0.15,0.95}}
\usepackage{url}

\theoremstyle{plain}

\theoremstyle{definition}

\theoremstyle{remark}

\newcolumntype{C}{>{\centering\arraybackslash}X}
\definecolor{Gray}{gray}{0.94}
\colorlet{mydarkgreen}{green!60!black}

\newcommand{\envname}{\textsc{BesiegeField}\xspace}

\definecolor{NMIHeader}{RGB}{223,220,203}
\definecolor{NMIPanel}{RGB}{238,236,224}
\definecolor{NMIHuman}{RGB}{246,244,238}

\newcommand{\shline}{\noalign{\global\savewidth\arrayrulewidth\global\arrayrulewidth 1pt}\hline\noalign{\global\arrayrulewidth\savewidth}}
\newlength\savewidth
\providecommand{\parttoc}{}

\title{Compositional Machine Design as Program Synthesis with LLMs}

\author{Wenqian Zhang\textsuperscript{1}~~~~Yangyi Huang\textsuperscript{2}~~~~Weiyang Liu\textsuperscript{2,*}~~~~Zhen Liu\textsuperscript{1,*,\textdagger}\\[1.5mm]
  \textsuperscript{1}The Chinese University of Hong Kong, Shenzhen~~~~\textsuperscript{2}The Chinese University of Hong Kong\\[1mm]
  \textsuperscript{*}Equal Advising~~~~
  \textsuperscript{\textdagger}Corresponding Author
  \\[3.85mm]
  \href{https://besiegefield.github.io/}{\tt besiegefield.github.io}
  \vspace{-7mm}
  }

\newcommand{\paperabstract}{
Large language models (LLMs) have shown strong abilities in writing and revising programs, yet many program-synthesis benchmarks still evaluate programs in symbolic or digital environments. We introduce \textbf{compositional machine design}, a physically grounded form of program synthesis where machines are written as programs that compose standardized parts, and success is determined by simulated physical behavior. To study this problem, we present \envname, a testbed built on the machine-building game \textit{Besiege}. In \envname, LLM agents generate machine programs from textual functional demands, execute the resulting machines in simulation, and receive rewards and state feedback. We benchmark LLM agents across representative machine-design tasks under single-agent generation, iterative editing, and hierarchical workflows. Strong models recover task-relevant structures and sometimes achieve nontrivial physical performance, but often struggle with spatially precise assembly, mechanism-level planning, and translating feedback into useful structural edits. We further finetune Qwen2.5-14B, an open-source LLM, with reinforcement learning from simulation-derived rewards. We find that, under a fixed generation budget, RL improves the best machine discovered. We additionally evaluate human performance to provide a reference point for task difficulty. These results establish compositional machine design as a testbed for studying LLM agents that synthesize executable machine programs and improve them through physical feedback.

}

\begin{document}
\maketitle

\doparttoc 
\faketableofcontents
\begin{abstract}
\paperabstract
\end{abstract}

% \footnote[2]{Corresponding Author}

% \setlength\epigraphwidth{5cm}
% \epigraph{\emph{``Man is a tool-making animal.''}}{--- \textup{Benjamin Franklin}}

\begin{figure*}[h!]
  \centering
  \includegraphics[width=\linewidth]{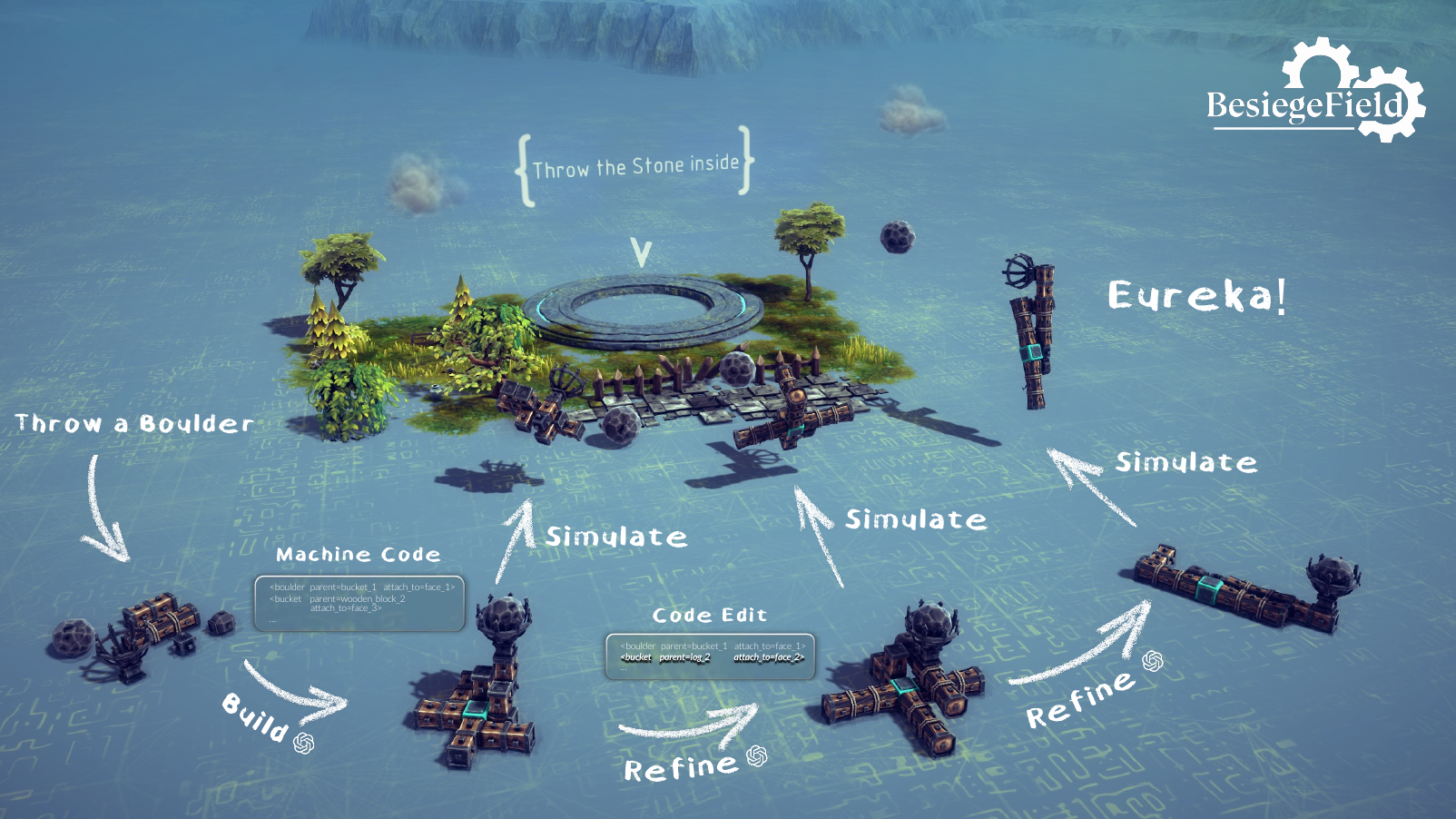}
  \caption{\footnotesize The task of compositional machine design is illustrated in our \envname environment. The figure shows a high-level sketch of the agentic workflow (w/ Gemini 2.5 Pro), along with the resulting machines and their simulated performance. The design objective is to create a machine that throws boulders long distances.}
  \label{fig:NMI-fig1c}
  \vspace{-1mm}
\end{figure*}

\section{Introduction}

The history of human progress is, in many ways, the history of machines, from the Antikythera mechanism that predicted eclipses to Leonardo da Vinci's sketches of flying machines. As large language models (LLMs) are increasingly studied for goal-directed reasoning and creation, from mathematical problem solving to scientific discovery, a natural question arises: 

\begin{center}
\textit{Can LLMs also reason about and design functional physical artifacts, such as machines that satisfy specified functional demands?}
\end{center}

Rather than treating machine design as an application of general-purpose tool use, we focus on a minimal setting that directly tests whether LLMs can turn functional intent into spatial composition under physical feedback. A mechanical clock, a vehicle, a launcher, or a lifting device is not defined by its parts alone, but by how those parts are arranged, constrained, actuated, and exposed to external inputs such as gravity, contact, load, or environmental forces. This makes compositional machine design close to program synthesis: the design is a structured object that can be written by a model, while its correctness is decided only when the specified composition is executed in physics.

\begin{center}
\textit{Can LLMs use physical feedback to write machine programs that compose standardized parts into functional machines?}
\end{center}

We name this minimalist setting \emph{compositional machine design}. A candidate machine is represented as a construction program that specifies components, attachments, orientations, and actuation. The program is executed in a physical simulator, where the resulting behavior provides task rewards and feedback. While abstracting away many details of real engineering, the setting focuses on the core challenge: translating textual intent into a search over spatial programs guided by physical feedback.

To study compositional machine design, we introduce \envname, an interactive environment adapted from the game \textit{Besiege}. \envname exposes a library of semantically meaningful mechanical components, a construction DSL based on a construction-tree representation, customizable physical scenarios, simulation-derived rewards, and rollout state feedback. LLM agents write construction programs, execute the corresponding machines in simulation, observe their behavior, and refine designs. This supports evaluation of both inference-time agentic refinement and model-level learning from physical rewards.

Using \envname, we evaluate LLM agents under single-agent generation, iterative editing, and hierarchical design workflows across seven task families. Strong models can produce recognizable and sometimes functional machines, but performance depends strongly on the construction representation, the feedback exposed after simulation, and the workflow used to revise failed designs. A recurring failure is the gap between a plausible mechanism-level plan and the exact attachments, orientations, supports, and actuation needed to make that plan work in simulation. We further study RL fine-tuning with simulation-derived rewards, finding that it improves the best machine discovered under a fixed generation budget. A small human reference study provides an additional scale for task difficulty. Our contributions are summarized as follows:

\begin{itemize}[leftmargin=*]
    \item We formulate \emph{compositional machine design} as an LLM-agent problem, where models synthesize construction programs for machines evaluated by physical simulation.
    \item We introduce \envname, an interactive testbed with standardized mechanical components, programmatic construction, scripted physical scenes, rollout feedback, and simulation-derived rewards.
    \item We benchmark LLM agents across seven machine-design task families, comparing single-agent generation, iterative editing, and hierarchical design workflows.
    \item We analyze the roles of construction representation, simulation feedback, and RL fine-tuning, including their effects on task performance and mechanism-level exploration.
    \item We conduct a short-exposure human reference study and characterize differences between human and LLM design behaviors.
\end{itemize}

\section{Related Works} 

\textbf{Programmatic 3D generation and machine construction.}
There is a long history in 3D asset generation and engineering design of representing the construction of a target instance as a program or a sequence of operations in a domain-specific language~\citep{ritchie2023neurosymbolic,sun20253d,deng2022unsupervised}, here called 3D graphics codes~\citep{qiu2025sgpbench, chen2025symbolic}. Unlike point clouds or meshes, these codes describe objects at a higher semantic level, capturing part composition, design constraints, and user operations in modeling software. Recent work studies CAD scripts~\citep{wu2023cad,alrashedy2025generating,li2025cad}, Blender macros~\citep{huang2024blenderalchemy}, LEGO assemblies~\citep{pun2025generating}, Minecraft structures~\citep{fan2022minedojo,liu2024odyssey}, and procedural scene generation~\citep{sun20253d,chen2025symbolic,jones2025shapelib,yuan2024cadtalk}. These works show that discrete construction programs can provide a useful interface for generative models. However, most of them focus on geometry, visual plausibility, or scene composition, rather than whether the generated structure functions under physical execution. Compositional machine design instead requires heterogeneous components with mechanical roles, such as wheels, gears, springs, hinges, and propellers, whose arrangement determines task behavior in simulation. Robot morphology design is also related~\citep{ha2016task,carlone2019robot,belmonte2022meta,ghansah2023humanoid,yuan2022transform2act,ringel2025text2robot,qiu2024robomorph}, but it is usually framed as morphology--control co-design. Our focus is broader machine assembly from standardized mechanical parts, with many tasks evaluated under fixed controls or no learned controller. A concurrent benchmark, BuildArena~\cite{xia2026buildarena}, likewise evaluates LLMs on physics-based engineering construction, defining Support, Transport, and Lift tasks organized by difficulty under a shared agentic workflow. \envname differs in scope and emphasis: machines are executable construction-tree programs whose success is a function of simulated physical behavior rather than a single target structure, and our experiments center on program representation, simulation-feedback search, RLVR post-training, and a human reference study.

\textbf{LLM agents.}
LLM agents organize language models into iterative loops of perception and action~\citep{yao2023react,minaee2024large,hu2024scenecraft}. They use external tools~\citep{schick2023toolformer,liu2024toolnet,kim2024leveraging,qin2024toolllm}, interact with simulated or real environments~\citep{savva2019habitat,shridhar2020alfworld}, refine outputs through reflection~\citep{hu20243d,alrashedy2025generating,shinn2023reflexion,yu2025generating}, and can be organized into multi-agent workflows~\citep{li2023camel,chen2024agentverse,zhang2024aflow}. Search and reflection have improved performance on many complex generation tasks~\citep{yao2023tree,putta2024agent,koh2024tree,besta2024graph,deng2024multi,renze2024self,xiao2025verbalized}. LLM agents have also been applied to code synthesis~\citep{gao2023pal,novikov2025alphaevolve,madaan2023self}, CAD design~\citep{alrashedy2025generating}, and game environments~\citep{wang2023voyager,fan2022minedojo}. Closest to our setting, \citet{makatura2023can} proposed an agent-based framework for generating mechanical structures from text prompts, but their system treats structure generation as a one-shot process and delegates geometric and physical parameter search to external optimization tools. Concurrent work, VLMgineer~\citep{gao2026vlmgineer}, also studies foundation-model-driven physical design, but focuses on robotic tool and action co-design rather than executable machine construction programs. In contrast, \envname studies whether LLM agents can directly revise executable machine programs using simulation feedback, and compares single-agent generation, iterative editing, and hierarchical construction in the same environment. 

\textbf{Reinforcement learning with verifiable rewards (RLVR).}
Recent studies show that RL fine-tuning with verifiable rewards from simulators or verifiers can improve reasoning behavior~\citep{shao2024deepseekmath,guo2025deepseek,bai2022constitutional}, even when fine-tuning uses a single prompt~\citep{wang2025reinforcement}. At the same time, reward optimization can reduce output diversity as entropy collapses during training. Mitigation methods include entropy or KL regularization~\citep{cui2025entropy,ouyang2022training}, Pass@\(k\) training~\citep{tang2025optimizing,chen2025pass}, and distribution-matching objectives such as generative flow networks~\citep{zhu2025flowrl,hu2023amortizing}. \envname provides simulation-derived rewards for machine programs, allowing us to test whether RLVR can improve physically grounded design and whether reward maximization narrows mechanism search.

\textbf{Benchmark suites for LLM agents.}
Recent LLM-agent evaluations range from tool use and API orchestration~\citep{qin2024toolllm,patil2024gorilla} to software engineering~\citep{jimenez2024swebench}, embodied environments~\citep{procthor}, web navigation~\citep{zhouwebarena}, and general agent benchmarks~\citep{liuagentbench}. These benchmarks test important forms of tool use, grounding, and long-horizon decision making, but they do not focus on generative mechanical reasoning under physical simulation. MineDojo~\citep{fan2022minedojo} and LEGO-Puzzles~\citep{tang2025lego} involve spatial construction or embodied interaction, but the generated structures are not evaluated as machines whose function depends on simulated mechanical behavior. \envname fills this gap: the generation quality is evaluated with the simulated physical behaviors.

\section{\envname: Testbed for Physics-Driven Machine Design}
\label{sec:besiegefield}

To make compositional machine design executable for LLM agents, we build \envname on top of the machine-building game \textit{Besiege}. An agent receives a task objective, a scene description, and a component library, and outputs a construction program specifying component choices, attachment relations, orientations, and actuation settings. The program is checked for validity, converted into a game machine, placed in a scripted physical scene, and executed by the simulator. The rollout returns task rewards and state information that can be used for later revisions. This section describes the environment interface; the benchmark tasks, reward definitions, and evaluation metrics are introduced in Section~\ref{sec:benchmark}.

\begin{figure}[t!]
  \centering
  \includegraphics[width=0.98\linewidth]{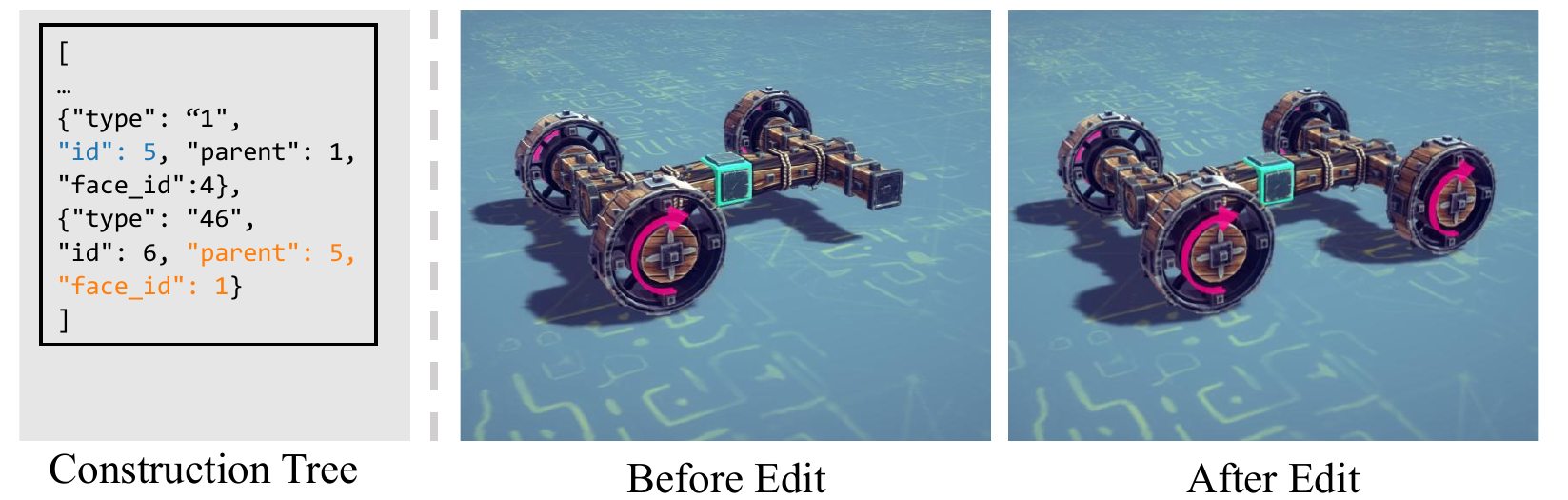}
  \captionsetup{font=footnotesize} %
  \caption{\footnotesize The construction-tree DSL. For the illustrated local edit: parent block info is in blue and child info is in yellow.}
  \label{fig:XML-intro}
  \vspace{-3mm}
\end{figure}

\paragraph{Construction programs.}
The native \textit{Besiege} machine file is coordinate-based: it lists blocks with their positions and rotations, while connectivity is inferred by the game engine from spatial contact. This format is difficult for LLM agents because small coordinate errors can break attachments and the intended part relations are not explicit. \envname therefore exposes a compact construction DSL (Fig.~\ref{fig:XML-intro}) based on a construction-tree representation. A program starts from a root block and then adds components one by one through local attachment commands. Each command specifies the new component type and identifier, the parent component, the parent face used for attachment, and the orientation of the new component. Components with two endpoints, such as springs or braces, specify two parent attachments. The DSL is then compiled back into the native machine format for simulation.

\paragraph{Physical execution.}
\envname exposes standardized mechanical components, including structural blocks, braces, wheels, hinges, rotating blocks, springs, grabbers, containers, projectiles, ballast weights, propellers, and cogs. These components have simple physical roles: structural blocks provide support, hinges permit rotation, springs exert elastic forces, containers hold objects, weights affect balance, and powered components such as wheels, cogs, and rotating blocks provide actuation. After a valid construction program is converted to the game format, the machine is simulated in a scripted scene. During rollout, the environment records task-relevant state such as positions, velocities, orientations, object displacement, projectile trajectories, and broken components. These records are used both for scoring and for feedback in agentic refinement workflows.

\begin{figure}[t]
  \centering
  \includegraphics[width=0.98\linewidth]{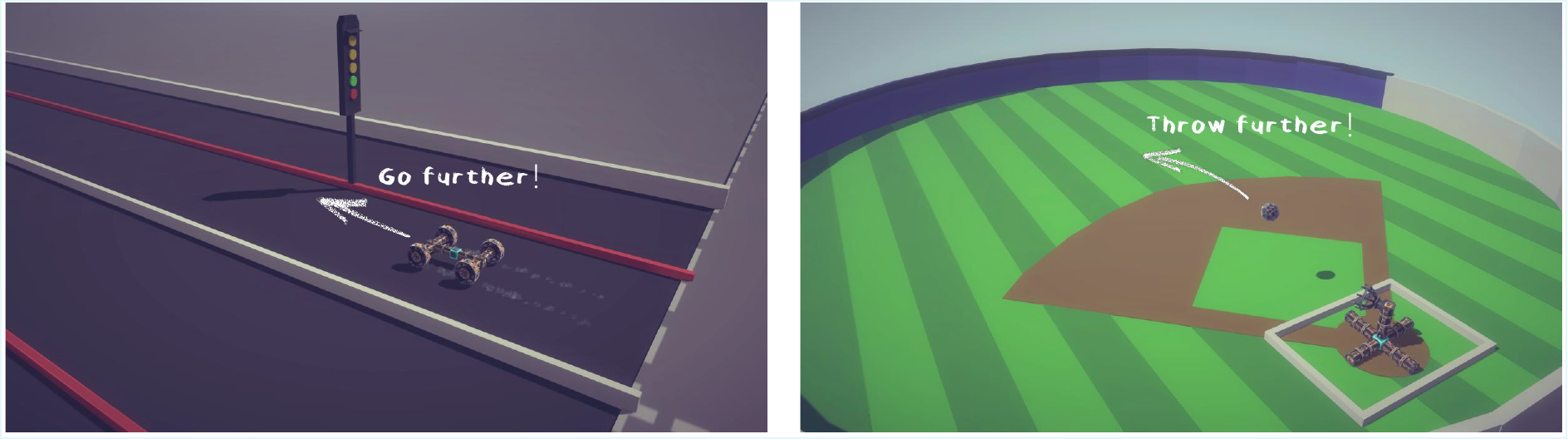}
  \captionsetup{font=footnotesize} %
  \caption{\footnotesize Demonstration of the machine design tasks in our experiments. (Left: Movement; Right: Throwing).}
  \label{fig:besiegefield_intro}
  \vspace{-3mm}
\end{figure}

\section{Benchmarking LLMs for Compositional Machine Design}
\label{sec:benchmark}

We first evaluate whether LLM agents can write construction programs whose physical rollouts satisfy textual functional demands. Unlike ordinary code generation, success here is measured not only by whether the output parses, but by whether the assembly behaves as intended under simulation.

\subsection{Benchmark Setting}
\label{sec:benchmark_setting}

\textbf{Tasks.}
We evaluate LLM agents on seven task families in \envname: \textsc{Movement}, \textsc{Throwing}, \textsc{Delivery}, \textsc{Jumping}, \textsc{Lifting}, \textsc{Flying}, and \textsc{Gearing}. These tasks cover locomotion, object transport, projectile launching, vertical motion, aerial movement, and transmission-like mechanisms. \textsc{Picking} is supported by the environment but excluded from the main benchmark because it couples mechanism design with control-policy design. Each task is specified by a textual functional demand and a scripted physical scene; rewards are computed from rollout outcomes such as travel distance, projectile distance, object displacement, lifted height, or success counts.

\textbf{Representative analysis tasks.}
For detailed analysis, we focus on \textsc{Movement} and \textsc{Throwing}, two compact tasks that stress complementary forms of reasoning. \textsc{Movement} primarily tests stable spatial assembly: correct wheel orientation, balance, symmetry, and support. \textsc{Throwing} tests coordinated dynamics: components must interact over time to transfer motion to a projectile. As shown in Fig.~\ref{fig:besiegefield_intro}, these two tasks are simple enough for repeated LLM generation, yet rich enough to expose failures in spatial assembly, mechanism planning, and physical prediction.

\textbf{Control protocol.}
To isolate machine structure and assembly from controller search, we use a fixed open-loop control protocol. After a valid machine is loaded into the scene, it first settles under gravity; powered components are then activated according to a fixed schedule. Performance differences therefore mainly reflect the generated construction program rather than a learned task-specific controller. The protocol is disclosed in every prompt. Allowing a separately optimized controller per machine would introduce an additional coupled optimization problem and make scores incomparable across models; we therefore leave joint morphology–control optimization to future work.

\begin{figure*}[t]
  \centering
  \includegraphics[width=\linewidth]{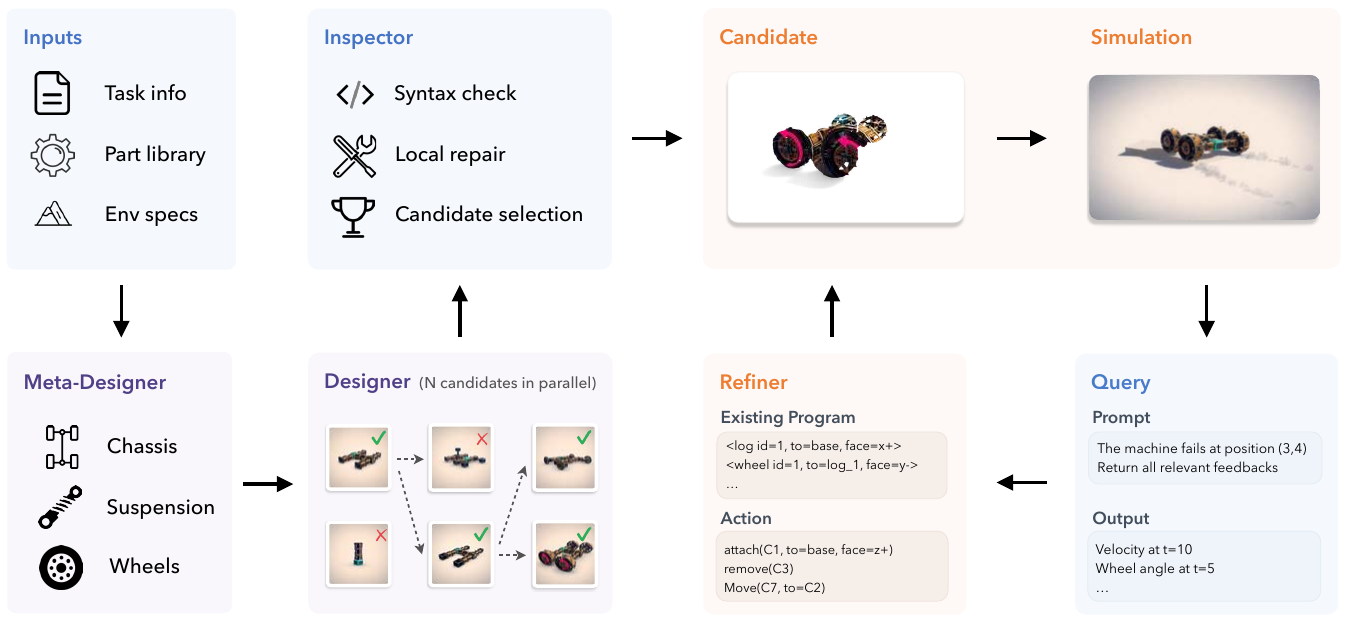}
    \caption{
        \footnotesize \textbf{Agentic workflows in \envname.}
        Hierarchical agentic workflow. A meta-designer decomposes the task into functional modules; designer agents construct candidate machines through module-wise search; an inspector identifies structural or conceptual issues; and simulation-guided refinement improves candidates using selected state feedback.
    }
  \label{fig:NMI-pipeline}
  \vspace{-2mm}
\end{figure*}

\begin{figure}[t]
  \centering
  \includegraphics[width=\linewidth]{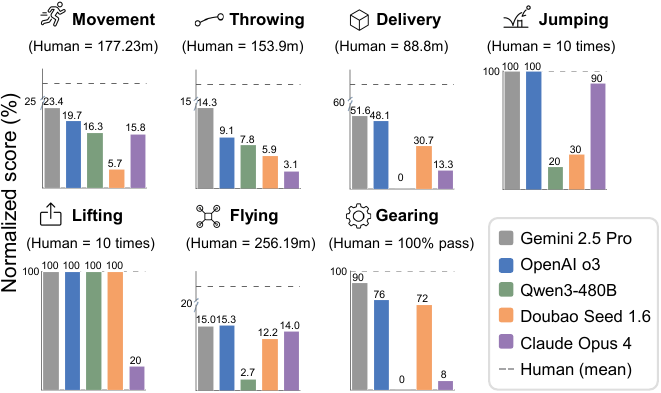}
    \caption{
        \footnotesize \textbf{LLM benchmark across evaluated task families.}
        Bars show LLM best performance task scores normalized by experienced-player mean performance where available, across Movement,Throwing, Delivery, Jumping, Lifting, Flying and Gearing. The experienced-player reference is used only as a scale reference and is not part of the human study in Section~\ref{sec:human}.
    }
  \label{fig:EMNLP-LLMBenchmark}
  \vspace{-2mm}
\end{figure}

\textbf{Metrics.}
We report parse validity, spatial validity, machine validity, mean simulation score, and maximum simulation score. Parse validity measures whether the generated output can be decoded into the construction DSL. Spatial validity measures whether the assembly satisfies attachment constraints and avoids severe spatial conflicts. Machine validity requires both parse and spatial validity; invalid machines receive zero reward. Mean score measures reliability under a fixed generation budget, while maximum score measures whether an agent discovers at least one strong design.

\subsection{Agentic Workflow Design}
\label{sec:agentic_workflow}

\textbf{Single-agent generation.}
The single-agent workflow asks one LLM to produce a complete construction program from the task objective, component descriptions, construction syntax, and environment documentation. The model generates a design chain-of-thought (CoT) and then emits the machine program. This is the direct text-to-machine-program setting.

\textbf{Iterative editing.}
A one-shot design often fails because of a small but decisive physical error: a wheel faces the wrong axis, a projectile holder attaches to the wrong parent, or a powered block rotates against the intended motion. We therefore use an iterative editing workflow. A designer first proposes a candidate machine. The environment-query module executes the simulation and summarizes task-relevant feedback, such as travel distance, projectile trajectory, global orientation, and broken components. A refiner then edits the current construction program rather than starting over.

The workflow includes an inspector before simulation-guided refinement. The inspector critiques the design from a structural perspective, and only the first refiner receives this critique. Later refiners use the program, edit history, and simulation feedback. At each step, the refiner proposes multiple candidate edits; candidates are evaluated by rollout, and the best branch is selected through search.

\textbf{Hierarchical construction.}
The hierarchical workflow separates functional decomposition from low-level construction. As shown in Fig.~\ref{fig:NMI-pipeline}, a meta-designer first decomposes the target machine into functional modules and specifies their intended interactions. Designer agents then construct the machine module by module. For each module, multiple candidates are generated in parallel; invalid partial machines are discarded, and valid candidates are retained for further expansion in a beam-search-style procedure. After the full machine is assembled, the same inspection and refinement loop is applied. This workflow tests whether functional decomposition helps LLMs bridge mechanism-level plans and construction-level operations.

\begin{figure*}[h!]
  \centering
  \includegraphics[width=\linewidth, clip, trim={0 0.7cm 0 0}]{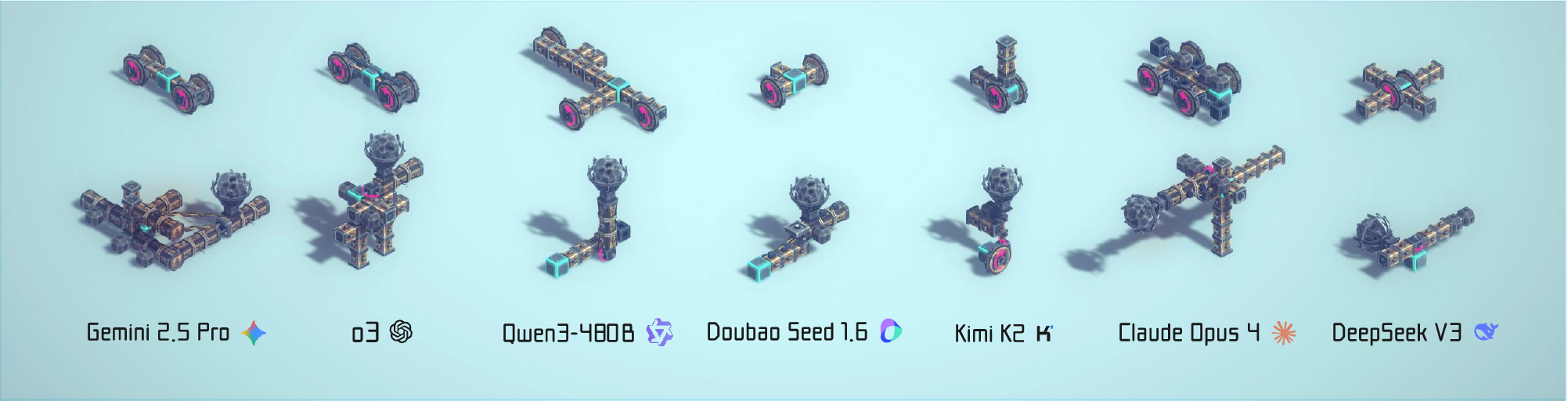}
  \caption{
    Representative machines generated by LLM agents on \textsc{Movement} and \textsc{Throwing}.
Top: \textsc{Movement}; bottom: \textsc{Throwing}. The examples show that LLMs can produce recognizable task-relevant structures, though physical validity and rollout performance vary across models and workflows.
    }
  \label{fig:generated_machine_gallery}
  \vspace{-.5mm}
\end{figure*}

\begin{figure*}[h!]
  \centering
  \includegraphics[width=\linewidth]{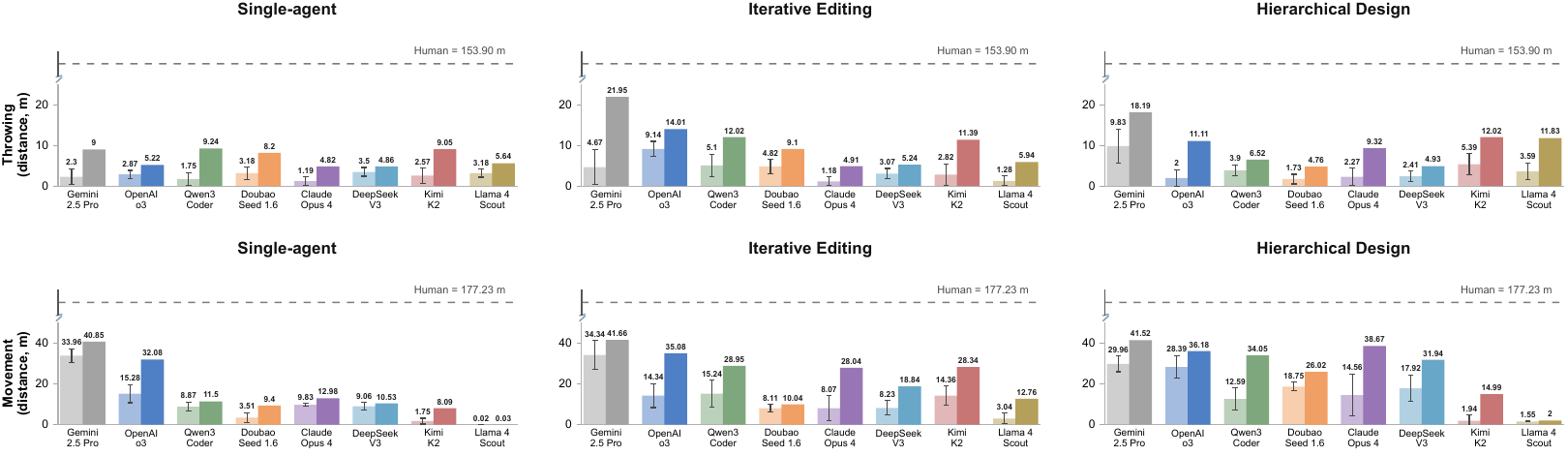}
      \caption{\footnotesize \textbf{Agentic workflow performance on representative Throwing and Movement tasks.}
For each model, bars report mean and maximum simulation scores under the fixed evaluation budget; whiskers on mean bars indicate 0.5$\times$std. The dashed broken-axis line shows an experienced-player community mean as a scale reference.
}
  \label{tab:agentic}
  \vspace{-2mm}
\end{figure*}

\subsection{Main Results and Observations}
\label{sec:agent_results}

\textbf{LLMs can synthesize nontrivial machine programs.}
Detailed results are reported in Fig. ~\ref{tab:agentic}. Compositional machine design is challenging, but strong models can recover task-relevant structures and sometimes achieve nonzero rollout performance, as shown in Figs.~\ref{fig:EMNLP-LLMBenchmark} and~\ref{fig:generated_machine_gallery}. In \textsc{Movement}, successful designs often contain stable bases, support structures, and powered wheels in useful orientations. In \textsc{Throwing}, agents sometimes produce launcher-like assemblies with projectile holders, pivots, supports, and counterweights. These results show that current LLMs can do more than emit syntactically valid programs: they can sometimes map textual functional demands to plausible physical compositions.

\textbf{Failures are mainly structural and physical.}
Failure cases are rarely just formatting errors. Appendix Fig.~\ref{fig:failure-mode} shows common patterns, including incorrect part orientations, wrong parent attachments, missing supports, instruction-following failures, and flawed mechanism choices. Some machines contain plausible components but lack the geometry needed to transfer motion or remain stable. Others expose a mismatch between CoT and program: the model describes a reasonable mechanism but emits a construction tree that does not realize it.

\textbf{Simulation feedback and edit history help refinement.}
Simulation feedback improves performance over one-shot generation, especially when it exposes editable failures such as motion in the wrong direction, failed projectile release, orientation drift, or broken components (Fig.~\ref{tab:rl_env_feedback}). Edit history reduces repeated invalid attempts and helps preserve useful partial structure (Appendix Table~\ref{tab:edit_history}). Scalar reward alone is less useful than state feedback: it ranks candidates, but often does not reveal which attachment, orientation, or support caused the failure.

\textbf{Hierarchy helps when the high-level plan is reliable.}
Hierarchical construction improves mean performance when the meta-designer produces a mechanically meaningful decomposition, as in the stronger Gemini 2.5 Pro setting (Fig.~\ref{tab:agentic}). It also tends to reduce variance, consistent with a more structured search space. The same commitment can hurt weaker models: a flawed high-level blueprint may cause downstream agents to refine the wrong mechanism rather than replace it. Hierarchy is thus useful as a search bias, but only when the abstract plan remains physically grounded.

\textbf{Representation and spatial information matter.}
The construction-tree representation substantially outperforms coordinate-based generation (Appendix Table~\ref{tab:rep_comparison}). This supports using local attachment relations as the LLM-facing interface: parent choices, attachment faces, and orientations are more directly tied to machine structure than independent global positions and rotations. Parsed 3D summaries also improve performance in several settings (Appendix Table~\ref{tab:abl:machine-3d-info}), suggesting that current LLMs still struggle to infer spatial relations from construction programs alone.

\textbf{Design CoT helps planning more than construction.}
High-level reasoning improves semantic plausibility, but it does not remove the construction bottleneck. When other models are conditioned on Gemini-generated CoT, their outputs become more aligned with intended mechanisms and often improve in performance (Appendix Fig.~\ref{fig:LLM-feed-gemini-cot}). Yet the gap remains: knowing that a catapult needs a lever or counterweight is different from compiling that plan into correct attachments, orientations, and actuation settings. The CoT examples in Appendix Figs.~\ref{fig:gemini-cot-example} and~\ref{fig:o3-cot-example} illustrate both the promise and brittleness of such spatial reasoning.

\section{Reinforcement Learning from Simulation Rewards}
\label{sec:rl}

The agent workflows above use physical feedback at inference time. We next ask whether simulation-derived rewards can improve the generator through fine-tuning.

\subsection{Cold-start and RL Setup}
\label{sec:rl_setup}

We evaluate RL on representative \textsc{Movement} and \textsc{Throwing} tasks. Before RL, we build a cold-start dataset to give the model an initial distribution over valid and mechanically plausible construction programs. We collect textual descriptions of machine functions from \textit{Besiege} player communities and additional task-oriented prompts, then prompt Gemini 2.5 Pro to generate design CoT and machine programs. After filtering for parse validity, machine validity, and functional behavior, we obtain 9,982 valid machine--CoT pairs with no exact or paraphrased duplicates of the benchmark prompts. We use this dataset to fine-tune Qwen2.5-14B-Instruct; details are provided in Appendix~\ref{sec:cold_start_details}.

For RL, each sampled construction program is executed in \envname and assigned reward \(R=\mathbbm{1}_{\mathrm{valid}}\cdot P\), where \(\mathbbm{1}_{\mathrm{valid}}\) indicates whether the generated machine satisfies validity constraints and \(P\) is the task-specific performance score. For \textsc{Movement}, \(P\) is travel distance in the target direction. For \textsc{Throwing}, \(P\) depends on projectile distance and maximum height, with constraints that penalize degenerate solutions such as rolling or carrying the boulder. We fine-tune with GRPO and compare the standard advantage estimator with a Pass@\(k\)-style variant. Implementation details are given in Appendix~\ref{sec:rl_details}.

\subsection{Reinforcement Learning Results}
\label{sec:rl_results}

\begin{figure}[t]
  \centering
  \includegraphics[width=\linewidth]{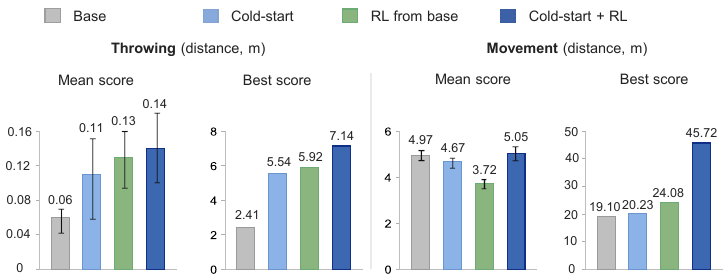}
    \caption{
        \footnotesize \textbf{RL fine-tuning performance on representative \textsc{Movement} and \textsc{Throwing} tasks.}
    \textit{Base} is Qwen2.5-14B-Instruct. \textit{Cold-start} is obtained by supervised fine-tuning on filtered machine--CoT pairs. \textit{RL from base} applies GRPO directly to the base model, while \textit{Cold-start + RL} applies GRPO after cold-start fine-tuning. Whiskers on mean bars indicate 0.05$\times$ std.}
  \label{fig:nmi-rlvr-b}
  \vspace{1mm}
\end{figure}

\begin{figure}[t]
  \centering
  \includegraphics[width=\linewidth]{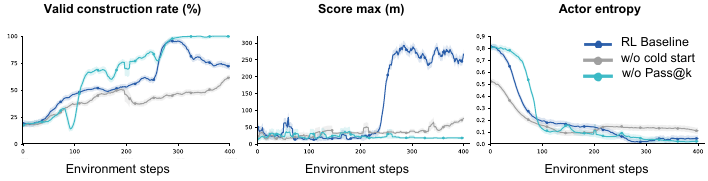}
    \caption{
        \footnotesize \textbf{RL fine-tuning dynamics on \textsc{Throwing}.}
    During RL fine-tuning, the validity rate increases, the maximum simulation score improves, and output token entropy decreases.
    }
  \label{fig:nmi-rlvr-c}
  \vspace{-2mm}
\end{figure}

\textbf{RL improves best-sample performance.}
The most robust gain is in best-sample discovery: under a fixed generation budget, RL substantially improves the best machine discovered, while mean score and validity changes are task-dependent (Table ~\ref{tab:rl_general}).
The best setting also improves maximum reward in the evaluation comparison (Fig.~\ref{fig:nmi-rlvr-b}). Cold-starting is important: RL from the cold-start model outperforms RL from the base instruction model, while cold-start fine-tuning alone remains insufficient. The training curves further show increasing validity, improving best score, and decreasing output entropy during RL (Fig.~\ref{fig:nmi-rlvr-c}).

\textbf{Pass@\(k\) better matches design search.}
High-reward machines are sparse, so the best candidate under a generation budget matters more than the average candidate. The Pass@\(k\)-style objective better matches this setting and discovers high-reward designs more effectively than the Pass@1-style estimator (Appendix Fig.~\ref{fig:RL-catapult-max-score}).

\textbf{RL mostly refines existing strategies.}
Qualitative evolution (Appendix Fig.~\ref{fig:nmi-rlvr-d}) shows that RL often improves local details while preserving the same high-level mechanism. Models adjust positions, supports, orientations, or nearby components, but rarely discover qualitatively different strategies once training has concentrated around a design family. This pattern is consistent with the simultaneous increase in reward and decrease in output entropy: simulation rewards help optimize designs, but current RL dynamics still favor refinement of existing templates over broad mechanism search.

\section{Human Reference Study}
\label{sec:human}

We conduct a short-exposure, non-expert human study to provide a reference point for the difficulty of \envname and for how designers use simulation feedback. Details are given in Appendix~\ref{app:user_study}. Participants receive the same component library and representative task objectives, together with brief instructions on the construction interface and the available rollout feedback. We record task performance and design behavior, including planning, testing, revision, and restarts (Appendix Fig.~\ref{fig:nmi-human-study}).

The tasks are difficult for participants after limited exposure to the environment: they do not immediately produce consistently high-performing machines. Their search behavior, however, differs from that of LLM agents in our experiments. Participants often alternate between planning, quick trial-and-error, and larger redesigns, while LLM agents tend to remain in local editing loops after proposing an initial mechanism. This suggests that access to feedback is not the only bottleneck; an effective designer must also decide when a failure calls for a local fix and/or a different mechanism.

\section{Discussion}
\label{sec:discussion}

\textbf{Capabilities for spatial program synthesis.}
Compositional machine design requires more than generating valid structured code. The program is a spatial object whose local edits change physical behavior: a wrong parent, face, axis, or support can invalidate an otherwise reasonable mechanism. The central difficulty is to keep three levels aligned: the functional demand, the mechanism-level plan, and the construction-level program. This also makes the agent interface part of the problem. Construction-tree representations, parsed spatial summaries, collision checks, simulator queries, visual inspection, and constrained edit operators all determine what the model can see and revise. Better LLM designers for this setting may thus come not only from stronger base models, but also from interfaces that expose spatial structure and physical failure modes in forms that models can act on.

\textbf{Diverse mechanism generation.}
Machine design should not be reduced to finding one high-scoring structure. Different mechanisms can satisfy the same function while trading off robustness, simplicity, controllability, efficiency, and ease of modification. Our RL experiments show that simulation rewards improve the best machine discovered under a fixed generation budget, but can also concentrate generations around narrower strategies. For design tasks, this means reward maximization alone may produce better templates without producing better designers. Future learning methods should treat structural diversity as part of the objective, encouraging models to maintain a repertoire of functional mechanisms rather than collapse onto the currently best-performing pattern.

\textbf{Toward general agentic design systems.}
\envname is much simpler than real engineering, but it isolates a problem that broader design agents will also face: turning a language-level functional demand into a structure whose value is determined by physical behavior. Real systems will need richer CAD-compatible representations, visual and geometric inspection, controllable actuation, more realistic simulation, and objectives beyond task score, including robustness, safety, fabrication cost, and ease of modification. Recent progress in text-to-CAD and part-assembly agents suggests that LLM-based systems are becoming better at producing plausible multi-part structures. 
\envname provides a controlled setting for studying how physics may enter this loop for complex assembly.

\section*{Limitations}

While \envname provides a controlled abstraction for studying compositional machine design, it should not be interpreted as direct evidence of real-world engineering capability. Its physics are simplified, its components are standardized, and the current benchmark omits key engineering constraints such as material properties, manufacturing processes, tolerances, safety margins and cost. 

The current formulation also focuses on text-based LLM agents, excluding visual inspection, sketching and embodied manipulation that human designers naturally use. Due to resource constraints and to ensure consistency with the human study, our evaluation focuses on mainstream LLMs rather than newly released frontier models. On the modeling side, current LLM agents often fail to ground high-level design CoT into spatially and mechanically valid assemblies; hierarchical workflows may propagate early errors and encourage local patching rather than mechanism-level redesign.

Reinforcement learning most robustly improves the best machine discovered within a fixed generation budget, while mean score and final-snapshot validity changes are task-dependent; in our experiments, it also mainly refines existing strategies instead of reliably discovering diverse mechanisms.
Moving closer to engineering practice will require richer multimodal and CAD-compatible representations, more realistic physics, fabrication-aware constraints, and learning methods that preserve structural diversity.

\section*{Ethical Considerations}

The human-participant study was approved by the Ethics Committee of the authoring institute. All participants provided written informed consent before participation, and all data were anonymized before analysis.

This work uses machine design as a controlled simulation benchmark for evaluating LLM agents' compositional and physical reasoning abilities. It does not involve real-world mechanical design, manufacturing, or safety-critical deployment, and the generated machines should not be interpreted as safe or reliable engineering designs. During the writing process, AI assistants were used only for language polishing and editing.

\section*{Acknowledgements} 

We sincerely thank the developers of Besiege for creating such an inspiring game and for fostering an open, vibrant player community, without which our exploration of this idea would not have been possible. We also extend our gratitude to the developers of BepInEx, whose work made it possible for us to unlock the full potential of Besiege. This work was supported by the National Natural Science Foundation of China (Grant No. 6250076194).

\bibliography{references/besiegefield}
\clearpage % flush pending floats and start appendices on a fresh page
\onecolumn
\addcontentsline{toc}{section}{Appendix} % Add the appendix text to the document TOC
\renewcommand \thepart{} % make "Part" text invisible
\renewcommand \partname{}
\part{\Large{\centerline{Appendix}}}
\parttoc
% \newpage 
\twocolumn
\appendix

\pagebreak
\clearpage
\newpage

\section{\textsc{\envname} environment}
\label{appendix:env}

This section describes the implementation of \textsc{\envname}, including the construction interface, simulation control, state recording and block library used in our experiments. \textsc{\envname} is designed to expose the machine-building process in \textit{Besiege} through programmatic interfaces, allowing LLM-generated machine programs to be parsed, instantiated, simulated and evaluated automatically.

\subsection{Environment interface}

\begin{figure}[h!]
  \centering
  \includegraphics[width=0.9\linewidth]{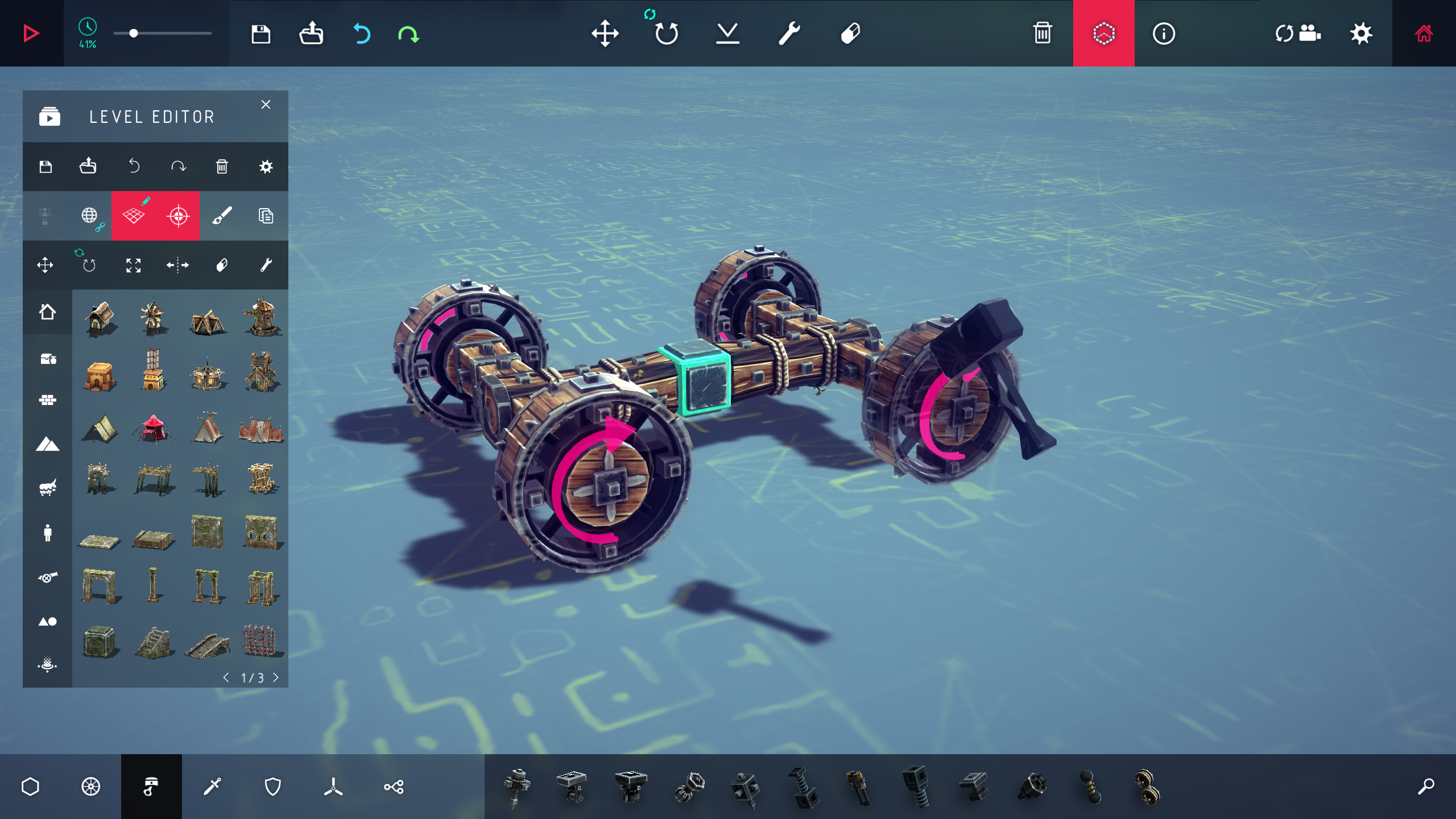}
  \caption{\textbf{The original editor interface of \textit{Besiege}.}}
  \label{fig:game_editor_view}
\end{figure}

\textsc{\envname} is implemented as a set of plug-in modules for the game \textit{Besiege}. These modules expose interfaces for constructing machines from structured descriptions, configuring simulation scenes, controlling powered components, recording block-level state information and specifying task termination conditions. This interface allows candidate machines to be generated automatically, deployed in scripted environments and evaluated using task-specific objectives. The original \textit{Besiege} editor interface, from which \textsc{\envname} inherits its block-based construction paradigm, is shown in Fig.~\ref{fig:game_editor_view}.

The environment records detailed state information during simulation, including block position, orientation, velocity and integrity. Task-specific scripts can additionally record object trajectories, event triggers and termination conditions, such as a component crossing a target line or a projectile reaching its maximum height. \textsc{\envname} also supports multi-process execution, enabling parallel evaluation of candidate machines for agentic search and reinforcement learning.

Because \textit{Besiege} is built on the Unity engine\footnote{\url{https://en.wikipedia.org/wiki/Unity_(game_engine)}} and supports modding, \textsc{\envname} can be extended with additional block types, environmental conditions and task mechanics. In this work, we focus on rigid-body machine construction using the block set (Table~\ref{tab:block_library}) and task families (Sec.~\ref{sec:tasks_rewards}).

\subsection{Construction rules}

Machines in \textsc{\envname} are built by incrementally attaching new blocks to an existing structure, beginning from a special root block. We refer to the newly added block as the \textit{attacher} and the existing block as the \textit{attachee}. In the standard construction rule, each \textit{attacher} uses one connection face, while each \textit{attachee} exposes zero or more attachable faces. Once an attachable face is used, it is treated as occupied and cannot be reused by another standard attachment.

If, after construction, the free end of a block coincides with an attachable face of an existing block, the simulator may automatically create a connection between the two blocks. This allows certain closed-loop or reinforcing structures to emerge from local attachment operations.

A small number of special blocks do not follow the standard single-parent rule. For example, springs have two endpoints and can be attached to two parent blocks. Such blocks may not occupy ordinary attachable faces in the same way as rigid blocks and are represented using two parent attachments in the construction tree.

Although \textit{Besiege} supports post-construction rescaling and rotation of blocks, these operations introduce additional continuous design variables. We therefore fix block scales and exclude post-construction transformations in the main experiments, so that the benchmark focuses on compositional assembly rather than low-level geometric tuning.

\subsection{Simulation and state recording}

After construction, a candidate machine is placed into the scene at a designated initial pose determined by the root block. The initial pose need not be on the ground, although task scripts may impose height or placement constraints. Once simulation begins, the machine is subject to gravity and the rigid-body dynamics implemented by the \textit{Besiege} physics engine.

For each simulation, \textsc{\envname} can record state trajectories for all blocks. The recorded variables include position, orientation, velocity and block integrity, and task-specific scripts may additionally record object trajectories, maximum heights, travel distances, contact events or breakage. These state trajectories are used both for computing task scores and for producing selected feedback summaries for LLM agents during iterative refinement.

\subsection{Block library}
\label{sec:block_library}

\textit{Besiege} contains a larger set of construction blocks. In our experiments, we use a filtered subset of 27 blocks that are most relevant for building classical mechanical structures such as vehicles, levers, trusses, launchers and counterweighted mechanisms. Table~\ref{tab:block_library} summarizes the blocks used in the experiments.

\begin{table*}[h!]
\centering
\begin{tabularx}{\textwidth}{llX}
\toprule
Block & Category & Description \\
\midrule
Starting Block & Root & Root of each machine; initial orientation is along the \(+z\) axis. \\
Small Wooden Block & Structure & Cubic basic construction block. \\
Wooden Block & Structure & Rectangular block shaped like two small wooden blocks. \\
Wooden Rod & Structure & Slender and relatively fragile construction block. \\
Log & Structure & Block shaped like three small wooden blocks arranged in parallel. \\
Brace & Structure & Reinforcement block for increasing structural strength. \\
Steering Hinge & Powered joint & Powered hinge that swings connected substructures left or right. \\
Steering Block & Powered joint & Powered block that rotates attached blocks around its placement axis. \\
Rotating Block & Powered joint & Motor-like block that generates torque around its local axis. \\
Universal Joint & Joint & Unpowered joint with free rotation around its placement axis. \\
Hinge & Joint & Unpowered hinge that swings around an axis perpendicular to placement. \\
Ball Joint & Joint & Unpowered joint allowing free rotation in multiple directions. \\
Axle Connector & Joint & Connector allowing unrestricted \(360^\circ\) rotation. \\
Powered Wheel & Wheel & Powered wheel with radius approximately 1 m. \\
Unpowered Wheel & Wheel & Unpowered wheel requiring external force to rotate. \\
Large Powered Wheel & Wheel & Larger powered wheel with radius approximately 3 m. \\
Large Unpowered Wheel & Wheel & Larger unpowered wheel. \\
Small Wheel & Wheel & Caster-like unpowered wheel. \\
Roller Wheel & Wheel & Shorter caster-like unpowered wheel. \\
Suspension & Support & Buffers forces from multiple directions. \\
Grabber & Interaction & Grabs objects on contact and can release them. \\
Boulder & Payload & Large rock-like object, useful as a projectile or payload. \\
Container & Payload & Holder typically used to carry or launch a boulder. \\
Grip Pad & Contact & High-friction pad. \\
Elastic Pad & Contact & High-elasticity pad. \\
Spring & Elastic connector & Contracting two-end connector with two parent attachments. \\
Ballast & Weight & Heavy cubic block used as a counterweight. \\
\bottomrule
\end{tabularx}
\caption{\textbf{Block library used in the main experiments.} The library includes structural blocks, joints, wheels, powered components, payload-related components and reinforcement or weighting blocks.}
\label{tab:block_library}
\end{table*}

\section{Machine program representation and validity checks}
\label{sec:machine_rep}
This section describes how machine designs are represented for LLM generation and parsed into \textsc{\envname}. The conceptual representation used throughout the paper is a \textit{construction tree}: a rooted assembly in which each new component is attached to an existing component through an explicit parent--child relation. This representation makes component connectivity explicit and is better aligned with the sequential construction process than a representation based only on global coordinates.

Native \textit{Besiege} machine files are XML-based. For exposition, we sometimes describe machine programs as XML-style structured code, because this makes the hierarchical relation between components intuitive. In the reported experiments, however, LLMs generate simplified construction-tree programs serialized as JSON. These JSON programs are parsed, validated and converted to the internal \textsc{\envname} machine format. This representation preserves the information needed for construction: component type, construction order, parent attachment and attachment face, with block-specific parameters assigned by default unless otherwise specified.

To reduce the complexity of the generation problem, we assume that blocks remain at their default scale and are not further transformed after attachment. The attachment operation itself determines the pose of a new block from the parent block, attachment face and block type, and may rotate the newly attached block according to the attachment rule.

\begin{table*}[t]
\scriptsize
  \centering
  \setlength{\tabcolsep}{3pt}
  \renewcommand{\arraystretch}{1.1}
  \newcommand{\cgr}[1]{\textcolor[rgb]{.329, .51, .208}{\textbf{#1}}}
  \newcommand{\cre}[1]{\textcolor[rgb]{1, 0, 0}{\textbf{#1}}}

  \begin{tabularx}{\textwidth}{lCCC>{\centering\arraybackslash}p{1.5cm}>{\centering\arraybackslash}p{1.5cm}}
    \toprule
    \multirow{2}{*}{Models}
    & \multicolumn{3}{c}{Machine Validity (pass/total)}
    & \multicolumn{2}{c}{Blind refinement Score} \\
    \cmidrule(lr){2-4}\cmidrule(lr){5-6}
    & File Valid & 3D Valid & Final Valid & Mean & Max \\
    \midrule
    \multicolumn{6}{l}{\textit{Baseline (construction-tree representation)}} \\
    \midrule
    Gemini 2.5 Pro    & 8/8 & 5/8 & 5/8 & 8.49 & 9.14 \\
    o3            & 3/8 & 3/3 & 3/8 & 0      & 0 \\
    Qwen3-Coder-480B-A35B    & 8/8 & 6/8 & 6/8 & 0.75 & 4.5 \\
    Doubao Seed 1.6    & 7/8 & 3/7 & 3/8 & 4.34 & 4.37 \\
    Claude Opus 4      & 8/8 & 4/8 & 4/8 & 4.17 & 4.36 \\
    DeepSeek-V3   & 7/8 & 7/7 & 7/8 & 0      & 0 \\
    Kimi K2       & 6/8 & 6/6 & 6/8 & 7.18 & 12.02 \\
    Llama 4 Scout 17B 16E        & 8/8 & 8/8 & 8/8 & 3.59      & 11.83 \\
    \midrule
    \multicolumn{6}{l}{\textit{Global position-based representation}} \\
    \midrule
    Gemini 2.5 Pro    & 5/8 & 5/8 & 5/8 & 4.96 & 12.85 \\
    o3            & 0/8 & - & - & - & - \\
    Claude Opus 4      & 0/8 & - & - & - & - \\
    Kimi K2       & 5/8 & 4/5 & 4/8 & 0 & 0 \\
    Llama 4 Scout 17B 16E        & 0/8 & - & - & -      & - \\
    \bottomrule
  \end{tabularx}
  \caption{
      \textbf{Ablation on machine program representation.}
      We compare the construction-tree representation with a global position-based representation. For each representation, we report parse validity, spatial validity, machine validity, and the mean and maximum designer-stage simulation score. Spatial validity is evaluated only for parse-valid outputs, while machine validity requires both parse validity and spatial validity. Mean scores are computed only over machine-valid outputs.
    }
  \label{tab:rep_comparison}
\end{table*}

\subsection{Coordinate-based representation}

The default machine representation used by the game records blocks through global poses. We derive a coordinate-based baseline representation by listing each block independently with its type, position and rotation. For most blocks, this representation contains only the block type and pose. For special blocks with two endpoints, such as springs, the second endpoint must also be specified.

An example coordinate-based representation is shown below:
\Needspace{12\baselineskip}
\noindent\begin{minipage}{\linewidth}
\begin{lstlisting}
[
    {"type": 0, "Position": [0, 0, 0],   "Rotation": [0, 0, 0, 1]},
    {"type": 1, "Position": [0, 0, 0.5], "Rotation": [0, 0, 0, 1]},
    {"type": 2, "Position": [0, 0, 2.5], "Rotation": [0, 0, 0, 1]},
    {"type": 9, "Position": [0, 0.5, 2], "Rotation": [-0.707, 0, 0, 0.707],
                "end-position": [0, 2, 0]}
]
\end{lstlisting}
\end{minipage}

In this representation, each block is recorded independently without explicitly specifying its adjacent blocks or attachment relations. For most block types, only the block type and global pose are recorded. For special two-end blocks, such as springs, an additional \texttt{end-position} field specifies the position of the second endpoint.

This representation is compact but does not explicitly encode adjacency or attachment relations. As a result, an LLM must infer the structure of the machine from global coordinates, which is difficult when machines contain many components or when small spatial differences determine whether an assembly is valid.

\subsection{Construction-tree representation}
\label{sec:tree_rep}

To make assembly structure explicit, we use a construction-tree representation. The machine is represented as an ordered list of construction steps. Each step specifies the component type, its unique identifier and the parent component to which it is attached. For standard blocks, the attachment is specified by a parent identifier and an attachment-face identifier. For special two-end blocks, such as springs, two parent attachments are specified.

The same example machine can be represented as:
\begin{lstlisting}
[
    {"type": 0, "id": 0, "parent": -1, "face_id": -1},
    {"type": 1, "id": 1, "parent":  0, "face_id":  0},
    {"type": 2, "id": 2, "parent":  1, "face_id":  0},
    {"type": 9, "id": 3, "parent_a": 0, "face_id_a": 4,
                         "parent_b": 1, "face_id_b": 6}
]
\end{lstlisting}

Here, \texttt{type} denotes the component type, \texttt{id} denotes the construction-order identifier, \texttt{parent} denotes the parent block and \texttt{face\_id} denotes the parent face used for attachment. The first block is always the unique starting block with \texttt{id=0}. For this root block, the parent and face identifiers are set to sentinel values. Two-end blocks use \texttt{parent\_a}, \texttt{parent\_b}, \texttt{face\_id\_a} and \texttt{face\_id\_b} to specify both attachments.

In the main experiments, most block-specific parameters, such as default scale and default actuation settings, are not exposed as free-form generation variables. This keeps the generated program compact and focuses the task on compositional assembly. When powered or special blocks require additional control settings, these are assigned by the parser using predefined defaults unless the experimental setting explicitly exposes them.

\subsection{Serialization and parsing}

In the reported experiments, LLMs generate construction trees in JSON format. The parser checks whether the generated text can be decoded as a valid structured object, whether all required fields are present and whether parent identifiers refer to previously created blocks. The construction tree is then converted into the internal machine format used by \textsc{\envname} and instantiated in the simulator.

This representation is intentionally minimal. It does not require the LLM to specify global coordinates for every block; instead, spatial placement is determined by the parent attachment relation, face identifier and the block's attachment rule. Because block scales and post-attachment transformations are fixed in the main experiments, the parser can reconstruct the instantiated machine from the construction tree.

\subsection{Validity checks}

We evaluate generated machines at multiple validity levels. \textit{File validity} indicates that the generated JSON program can be parsed into a construction tree with the required fields and valid references. \textit{Spatial validity} indicates that the instantiated assembly satisfies attachment constraints and avoids invalid spatial conflicts such as severe self-intersections. \textit{Machine validity} requires both file validity and spatial validity.

Invalid machines receive zero task reward. For valid machines, \textsc{\envname} executes the simulation and computes task-specific performance scores from the recorded state trajectory. Ablations comparing the construction-tree representation with the coordinate-based representation are reported in Table~\ref{tab:rep_comparison}, and the effect of adding explicit spatial information is reported in Table~\ref{tab:abl:machine-3d-info}.

\section{Task families, parameterization and rewards}
\label{sec:tasks_rewards}

\textsc{\envname} implements eight parameterized task families: Movement, Throwing, Delivery, Jumping, Lifting, Flying, Gearing and Picking. These tasks cover different forms of compositional machine behavior, including locomotion, object transport, projectile launching, vertical motion, aerial movement, transmission-like mechanisms and object retrieval. The main LLM-agent benchmark evaluates seven task families. Picking is implemented as a manipulation-style task but excluded from the main benchmark because it requires coupled mechanism and control-policy design, including navigation to the target, mechanism deployment and object retrieval.

Different experiments use different subsets of the task suite. Reinforcement learning focuses on representative Movement and Throwing tasks, while the user study focuses on Delivery and Throwing. Table~\ref{tab:task_usage} summarizes task usage across the paper.

\begin{table*}[h]
\centering
\begin{tabular}{p{0.15\linewidth}p{0.43\linewidth}ccc}
\toprule
Task family & Objective & Main benchmark & RL & User study \\
\midrule
Movement & Travel along a target direction or track & Yes & Yes & No \\
Throwing & Launch a projectile over distance & Yes & Yes & Yes \\
Delivery & Transport an object across terrain & Yes & No & Yes \\
Jumping & Cross one or more obstacles by jumping & Yes & No & No \\
Lifting & Raise a weighted object & Yes & No & No \\
Flying & Achieve altitude or navigate through marks & Yes & No & No \\
Gearing & Transmit rotation through gear placement & Yes & No & No \\
Picking & Retrieve an object from a constrained location & No & No & No \\
\bottomrule
\end{tabular}
\caption{
\textbf{Overview of \textsc{\envname} task families and their use across experiments.} \textsc{\envname} implements eight parameterized task families; seven are used in the main LLM-agent benchmark. Picking is implemented but excluded from the main benchmark because it requires coupled mechanism and control-policy design.
}
\label{tab:task_usage}
\end{table*}

\subsection{Task definitions}

\begin{figure}[h!]
  \centering
  \includegraphics[width=\linewidth]{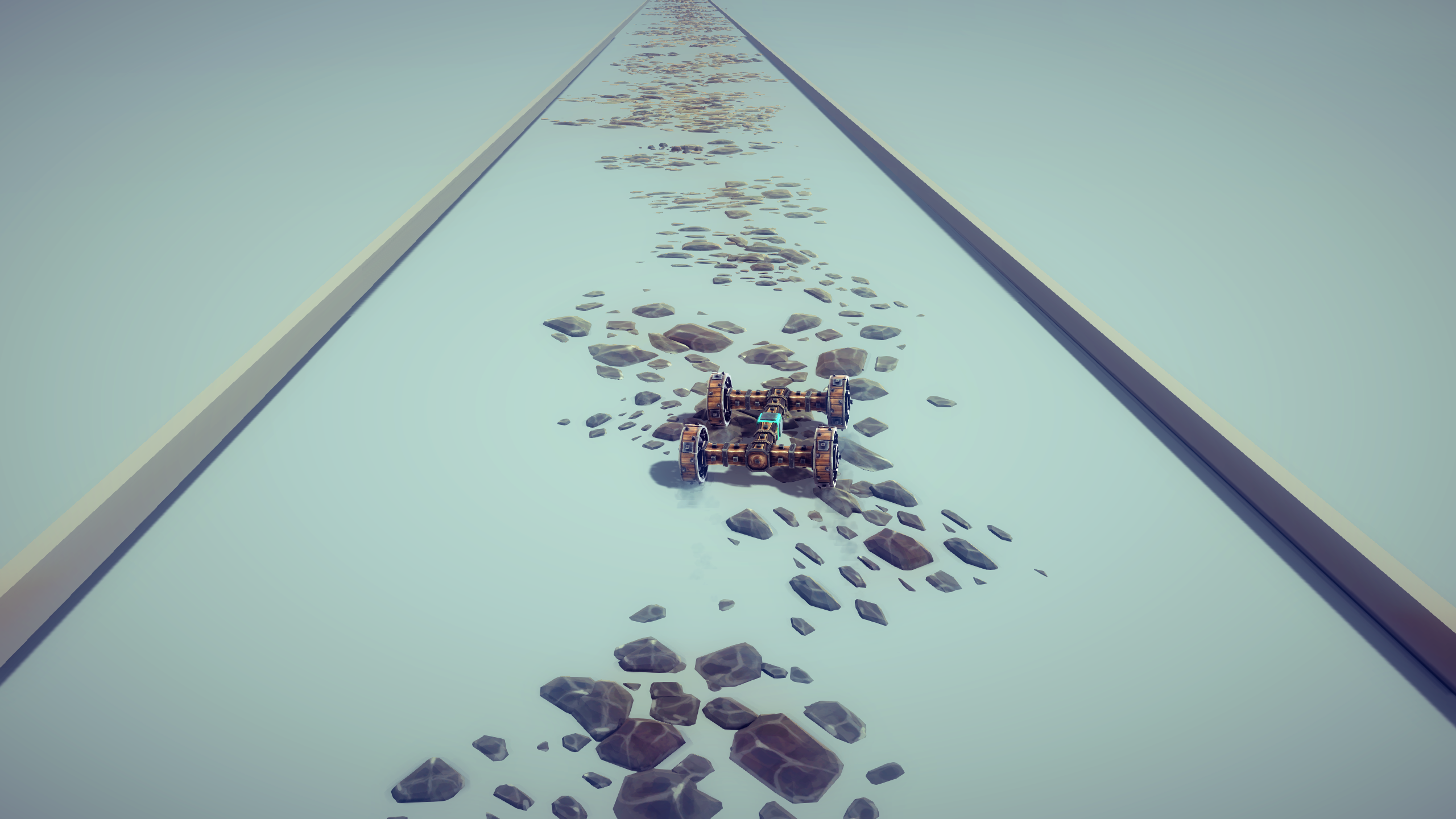}
  \caption{
  \textbf{Illustration of a Movement task variant on rocky terrain}. This setting is more difficult than the controlled Movement environment used in the main experiments.
  }
  \label{fig:task_movement_rocky}
\end{figure}

\begin{figure}[h!]
  \centering
  \includegraphics[width=\linewidth]{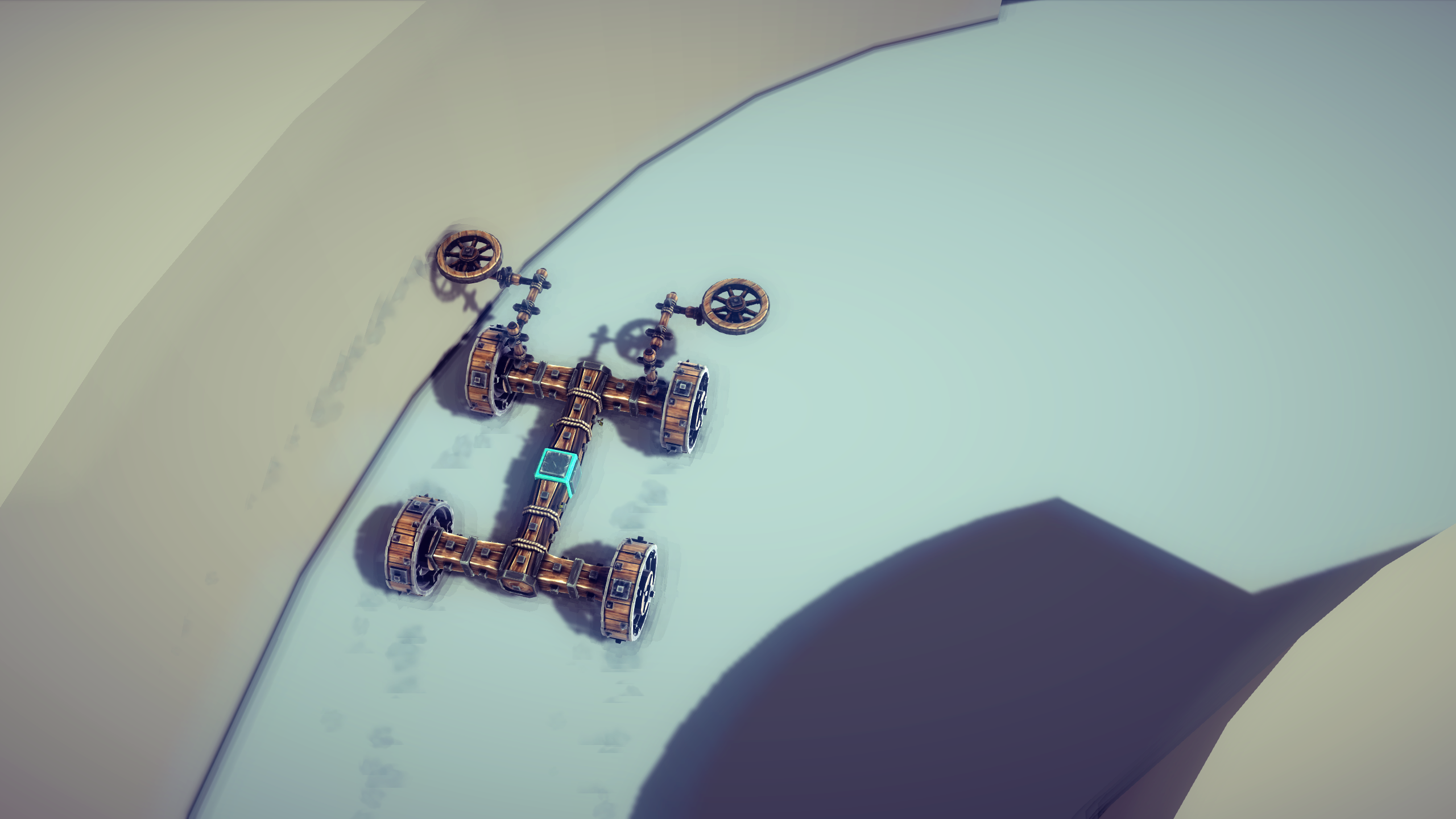}
  \caption{
  \textbf{Illustration of a Movement task variant with a curved track.}
  }
  \label{fig:task_curved_track}
\end{figure}

\paragraph{Movement.}
The Movement task, referred to as the \textit{car} task in some experiments, requires the agent to build a machine that travels along a designated direction or track. The task primarily tests stable assembly, wheel placement, orientation, symmetry and support. Performance is measured by the distance traveled by the machine's root block along the target direction during the simulation window. Examples of more difficult Movement variants are shown in Figs.~\ref{fig:task_movement_rocky} and \ref{fig:task_curved_track}.

\begin{figure}[h!]
  \centering
  \includegraphics[width=\linewidth]{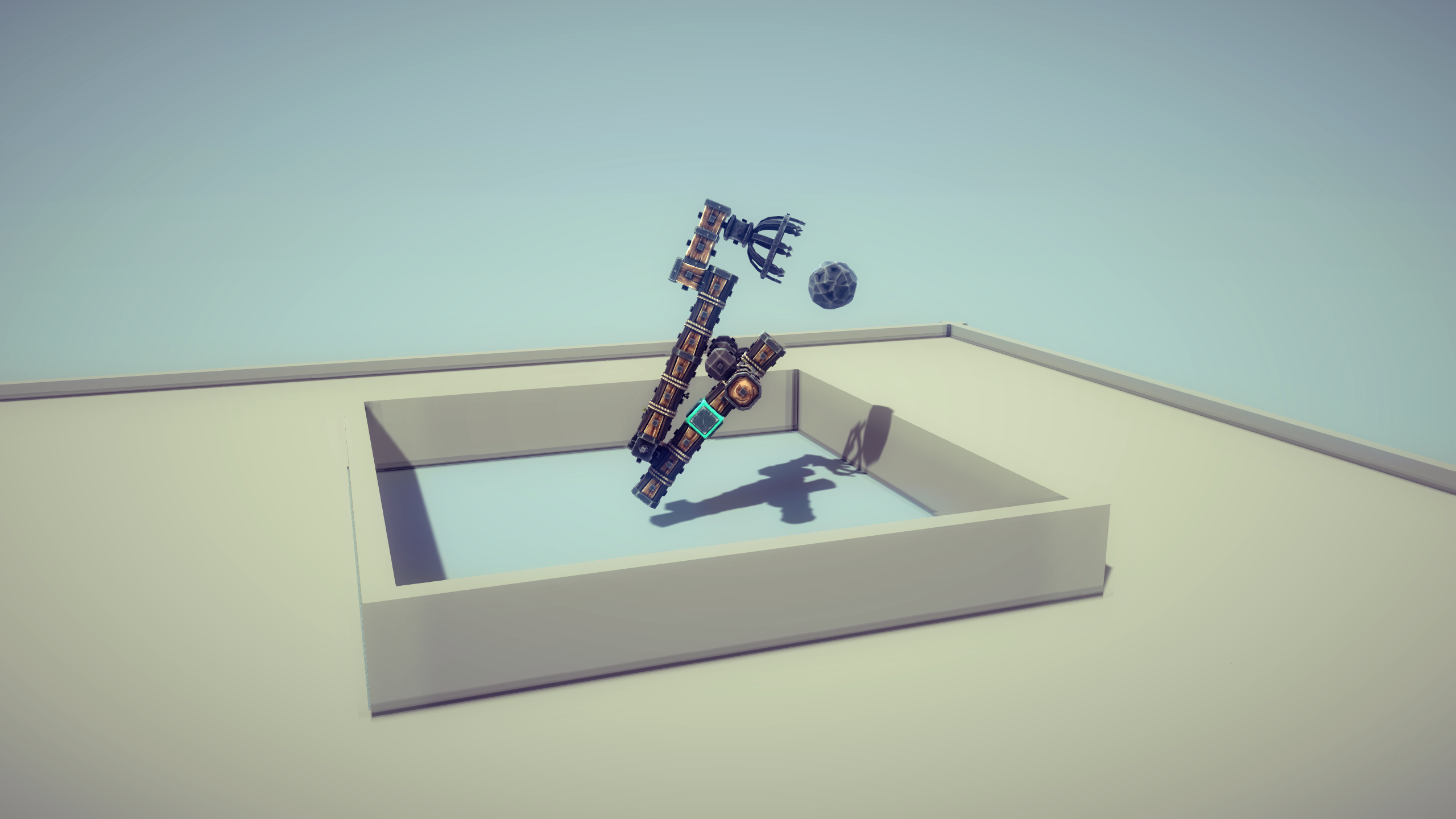}
  \caption{
  \textbf{Illustration of the Throwing task.}
  }
  \label{fig:task_throw}
\end{figure}

\begin{figure}[h!]
  \centering
  \includegraphics[width=\linewidth]{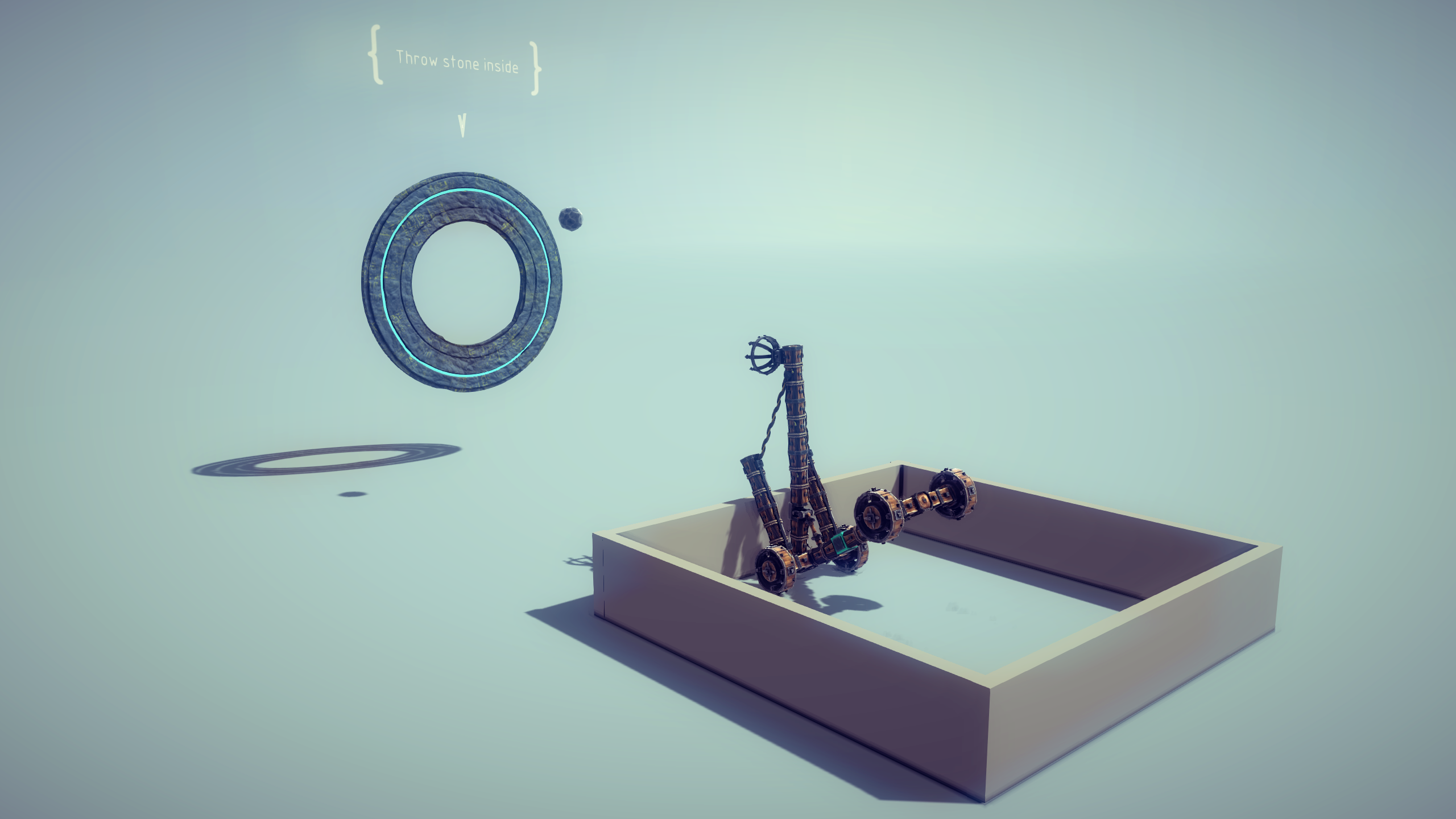}
  \caption{
  \textbf{Illustration of a Throwing task variant.} The objective of the task is to launch the boulder through a target ring.
  }
  \label{fig:task_thru_ring}
\end{figure}

\paragraph{Throwing.}
The Throwing task, referred to as the \textit{catapult} task in some experiments, requires the agent to build a machine that launches a boulder over a long distance. To discourage degenerate strategies such as carrying the boulder forward or allowing it to roll along the ground, the designated building area is enclosed by a medium-height wall. Performance is measured from the projectile trajectory, using horizontal travel distance and, in the reinforcement-learning experiments, maximum height. Example Throwing variants are shown in Figs.~\ref{fig:task_throw} and \ref{fig:task_thru_ring}.

\begin{figure}[h!]
  \centering
  \includegraphics[width=\linewidth]{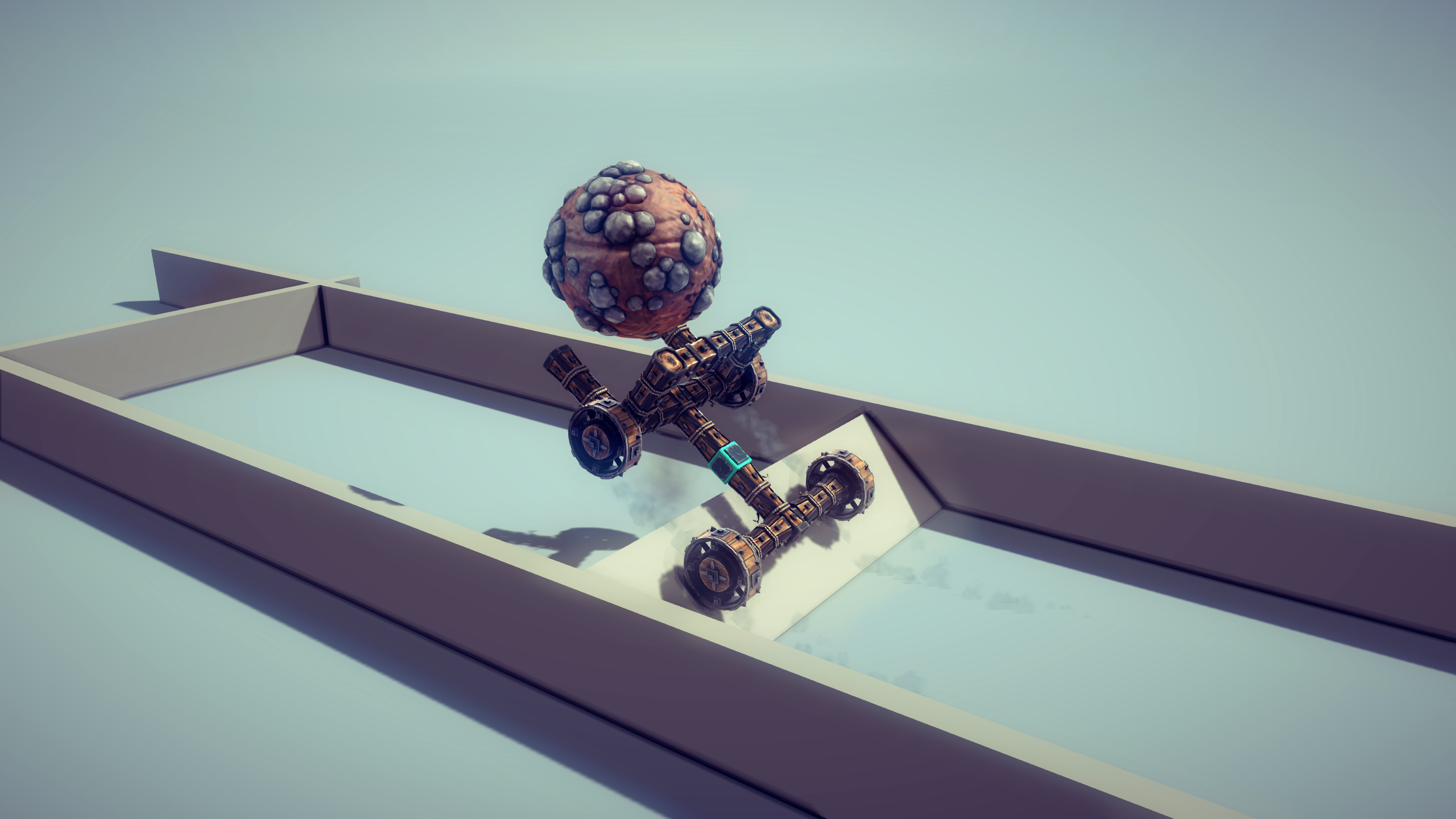}
  \caption{
  \textbf{Illustration of the Delivery task with a bump on the track.}
  }
  \label{fig:task_delivery}
\end{figure}

\paragraph{Delivery.}
The Delivery task requires the agent to build a machine that transports a large stone forward across terrain. This task emphasizes stability, payload support and the ability to maintain function while carrying an object. An example Delivery scene is shown in Fig.~\ref{fig:task_delivery}.

\paragraph{Jumping.}
The Jumping task requires the agent to build a machine that crosses one or more obstacles by jumping. This task tests the ability to generate vertical motion and coordinate structural stability with dynamic actuation.

\paragraph{Lifting.}
The Lifting task requires the agent to design a machine that raises a weighted geometric object. This task emphasizes mechanical advantage, stability under load and control of vertical motion.

\paragraph{Flying.}
The Flying task requires the agent to design a machine that gains altitude or traverses navigation marks while dealing with environmental barriers such as wind. This task tests aerial motion and stability under external forces.

\paragraph{Gearing.}
The Gearing task provides a plate or box containing a powered cog and a distant unpowered target cog. The objective is to transmit rotation from the powered cog to the target cog using unpowered cogs placed on the available surface. This task emphasizes discrete placement and contact-based transmission rather than free-form vehicle construction.

\begin{figure}[h!]
  \centering
  \includegraphics[width=\linewidth]{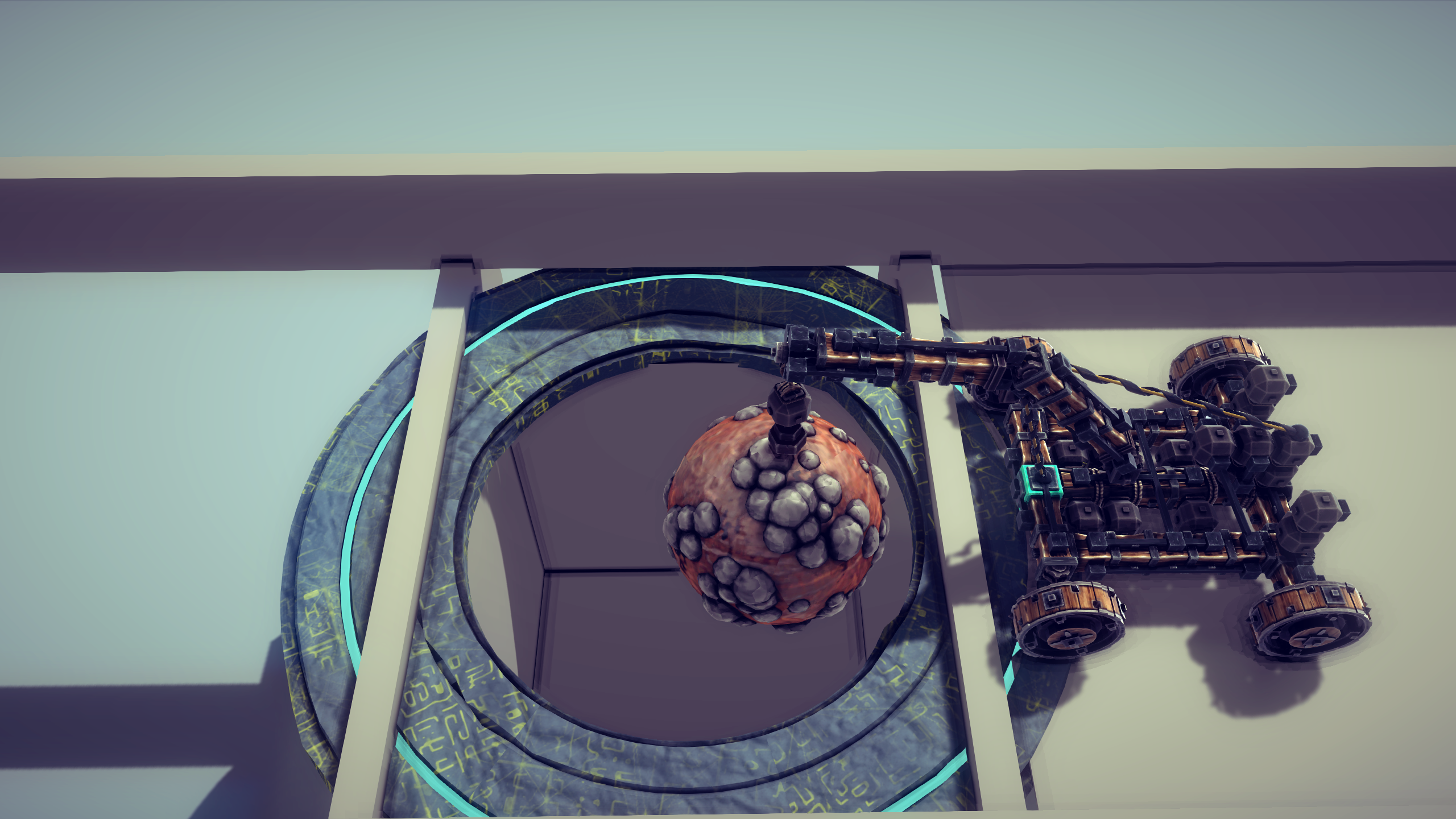}
  \caption{
  \textbf{Illustration of the implemented Picking task.}
  }
  \label{fig:task_pick}
\end{figure}

\paragraph{Picking.}
The Picking task requires the agent to retrieve an object from a constrained location such as the bottom of a well. We implement this task in \textsc{\envname} but exclude it from the main benchmark because it requires co-design of control and mechanism: a successful machine must navigate to the target, deploy a retrieval mechanism and grasp or extract the object. An example Picking scene is shown in Fig.~\ref{fig:task_pick}.

\subsection{Parameterized difficulty}

\begin{figure*}[h!]
  \centering
  \includegraphics[width=\linewidth]{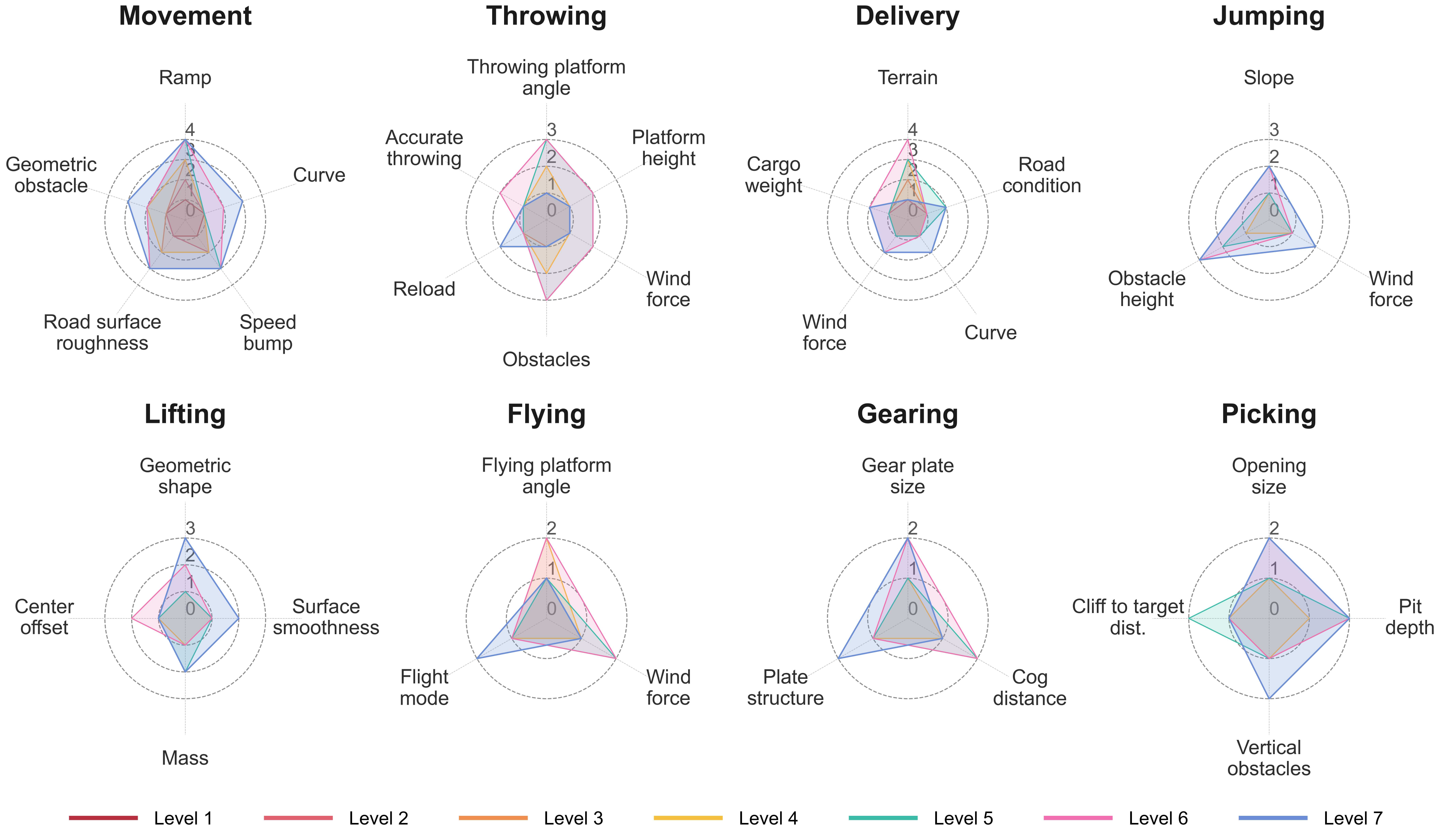}
  \caption{
  \textbf{Parameterized difficulty dimensions across the eight implemented task families.} Each radar chart visualizes controllable factors for a task family, such as terrain roughness, wind force, obstacle height, payload mass, or target geometry. Colored outlines indicate progressively higher scripted difficulty levels. In the main benchmark, we evaluate seven task families and use controlled parameter settings to isolate basic compositional design ability.
  }
  \label{fig:task_parameters}
\end{figure*}

All task families are parameterized, allowing scene difficulty to be varied through scripted parameters. In the main experiments, we use controlled settings to isolate basic compositional design ability, while the environment supports broader difficulty sweeps. Examples of these parameterized difficulty dimensions are shown in Fig.~\ref{fig:task_parameters}. The controllable parameters include:

\begin{itemize}
    \item \textbf{Movement}: ramp, curve, speed bump, road-surface roughness and geometric obstacles.
    \item \textbf{Delivery}: terrain, road condition, curve, wind force and cargo weight.
    \item \textbf{Throwing}: platform angle, platform height, wind force, obstacles, reload requirements and throwing accuracy.
    \item \textbf{Gearing}: gear-plate size, source--target cog distance and plate structure.
    \item \textbf{Flying}: platform angle, wind force and flight mode.
    \item \textbf{Lifting}: object shape, surface smoothness, mass and center offset.
    \item \textbf{Jumping}: slope, wind force and obstacle height.
    \item \textbf{Picking}: opening size, pit depth, vertical obstacles and cliff-to-target distance.
\end{itemize}

\subsection{Additional task examples and benchmark results}

\begin{table*}[t]
\scriptsize
  \centering
  \setlength{\tabcolsep}{3pt}
  \renewcommand{\arraystretch}{1.1}

  \begin{tabularx}{\textwidth}{l*{6}{>{\centering\arraybackslash}X}}
    \toprule
    \multicolumn{7}{l}{\textit{Continuous-score tasks}} \\
    \midrule
    \multirow{2}{*}{Models}
    & \multicolumn{3}{c}{Flying: height}
    & \multicolumn{3}{c}{Delivery: distance} \\
    \cmidrule(lr){2-4}\cmidrule(lr){5-7}
    & Mean & Max & Std & Mean & Max & Std \\
    \midrule
    Gemini 2.5 Pro & 30.51 & 38.50 & 10.68 & 32.57 & 45.80 & 12.07 \\
    OpenAI o3 & 31.09 & 39.20 & 8.95 & 29.33 & 42.70 & 14.24 \\
    Qwen3-Coder-480B-A35B & 3.73 & 6.96 & 3.22 & 0 & 0 & 0 \\
    Doubao Seed 1.6 & 15.61 & 31.19 & 12.56 & 11.42 & 27.30 & 8.83 \\
    Claude Opus 4 & 28.76 & 35.74 & 6.98 & 3.28 & 11.80 & 4.62 \\
    DeepSeek-V3 & 0.31 & 0.51 & 0.20 & 0 & 0 & 0 \\
    Kimi K2 & 1.02 & 1.52 & 0.51 & 0 & 0 & 0 \\
    Llama 4 Scout 17B 16E & 0.26 & 0.51 & 0.25 & 0 & 0 & 0 \\
    Human reference & 256.19 & 568.09 & 181.42 & 88.80 & 214.84 & 49.30 \\
    \bottomrule
  \end{tabularx}

  \begin{tabularx}{\textwidth}{l*{6}{>{\centering\arraybackslash}X}}
    \multicolumn{7}{l}{\textit{Success-count tasks}} \\
    \midrule
    \multirow{2}{*}{Models}
    & \multicolumn{3}{c}{Jumping: successes / 10}
    & \multicolumn{3}{c}{Lifting: successes / 10} \\
    \cmidrule(lr){2-4}\cmidrule(lr){5-7}
    & Mean & Max & Std & Mean & Max & Std \\
    \midrule
    Gemini 2.5 Pro & 5.50 & 10 & 3.54 & 9.80 & 10 & 0.40 \\
    OpenAI o3 & 5.88 & 10 & 4.13 & 5.40 & 10 & 4.45 \\
    Qwen3-Coder-480B-A35B & 1.50 & 2 & 0.50 & 5.50 & 10 & 4.50 \\
    Doubao Seed 1.6 & 1.50 & 3 & 0.76 & 5.67 & 10 & 4.35 \\
    Claude Opus 4 & 2.86 & 9 & 3.91 & 1.50 & 2 & 0.50 \\
    DeepSeek-V3 & 0 & 0 & 0 & 0.75 & 1 & 0.43 \\
    Kimi K2 & 3 & 3 & 0 & 0.33 & 1 & 0.47 \\
    Llama 4 Scout 17B 16E & 1 & 1 & 0 & 0 & 0 & 0 \\
    Human reference & 10 & 10 & 0 & 10 & 10 & 0 \\
    \bottomrule
  \end{tabularx}

    \begin{tabularx}{\textwidth}{l*{6}{>{\centering\arraybackslash}X}}
    \multicolumn{7}{l}{\textit{Gearing}} \\
    \midrule
    Models & & & & & Pass / 50  & \\
    \midrule
    Gemini 2.5 Pro & & & & &45  & \\
    OpenAI o3 & & & & &38  & \\
    Qwen3-Coder-480B-A35B & & & & &0  & \\
    Doubao Seed 1.6 & & & & &36  & \\
    Claude Opus 4 & & & & &4  & \\
    DeepSeek-V3 & & & & &17  & \\
    Kimi K2 & & & & &5  & \\
    Llama 4 Scout 17B 16E & & & & &15  & \\
    Random Guess & & & & &5  & \\
    \bottomrule
  \end{tabularx}
    \caption{
        \textbf{Additional task-family results.}
        We report agent performance across additional evaluated task families beyond the representative Throwing and Movement analyses. Flying is scored by maximum height, Delivery by transported-object distance, Jumping and Lifting by the number of successful trials out of 10, and Gearing by the number of successful attempts out of 50. The Human reference row denotes performance from an experienced \textit{Besiege} player with close-to-expert building skill and is included for scale rather than as part of the short-exposure user study. Random Guess denotes a random-placement baseline for Gearing.
    }
  \label{tab:additional_tasks}
\end{table*}

\begin{figure*}[h!]
  \centering
  \includegraphics[width=\linewidth]{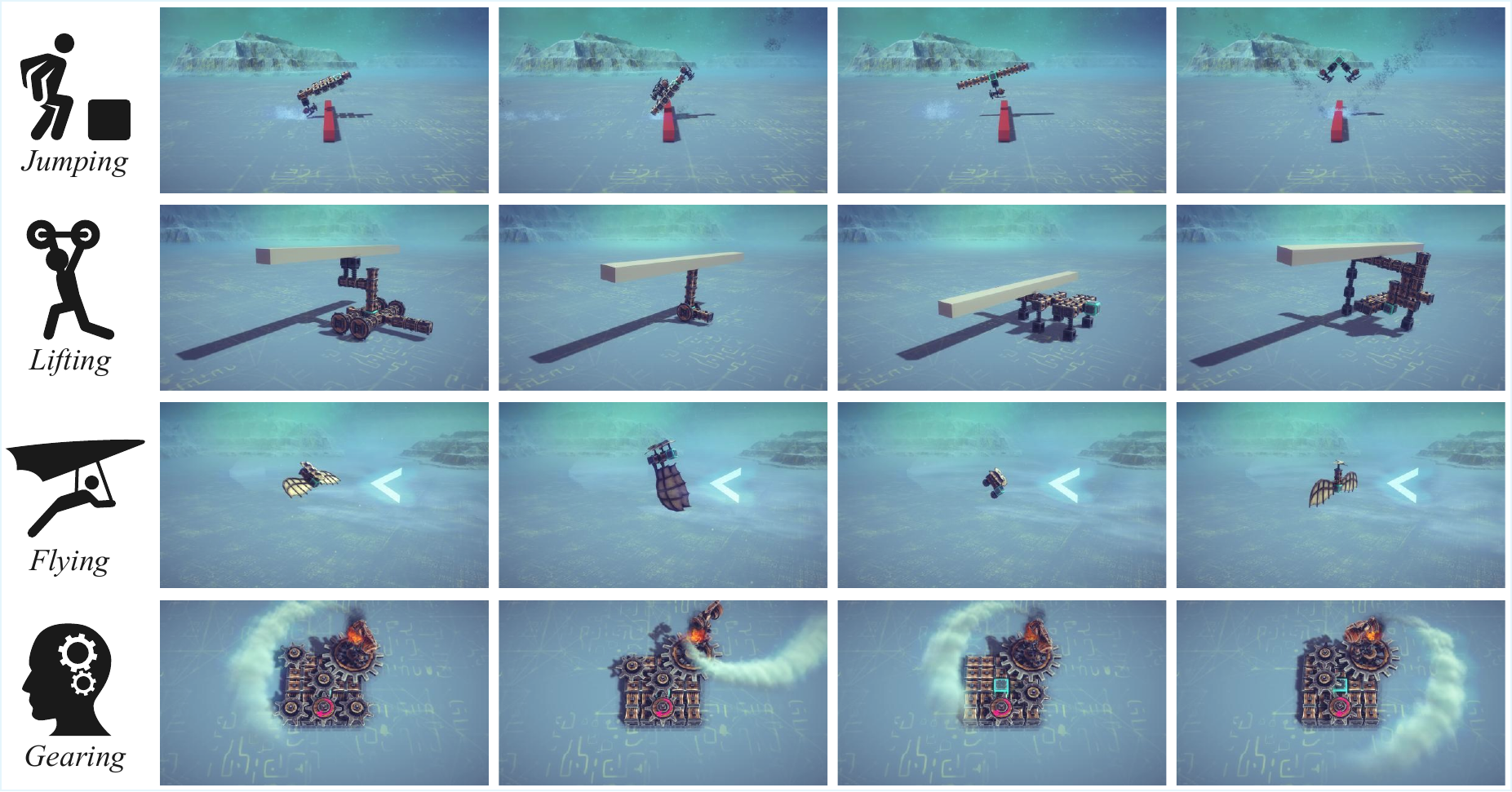}
  \caption{
  \textbf{Additional task-family machine examples.}
  Representative machines generated by the hierarchical agentic workflow on additional evaluated task families. All examples are generated by Gemini 2.5 Pro without task-specific fine-tuning.
  }
  \label{fig:additional_tasks}
\end{figure*}

Representative machines generated by the hierarchical agentic workflow on additional evaluated task families are shown in Figure~\ref{fig:additional_tasks}. All examples are generated by Gemini 2.5 Pro without task-specific fine-tuning.

Quantitative results for Flying, Jumping, Delivery, Lifting and Gearing are reported in Table~\ref{tab:additional_tasks}. Flying is scored by maximum height, Delivery by transported-object distance, Jumping and Lifting by the number of successful trials out of 10, and Gearing by the number of successful placement attempts out of 50. Because Gearing has a constrained placement space and limited refinement actions, we evaluate it using repeated placement attempts rather than the full hierarchical refinement workflow.

\subsection{Additional task illustrations}

\begin{figure*}[h!]
  \centering
  \includegraphics[width=\linewidth]{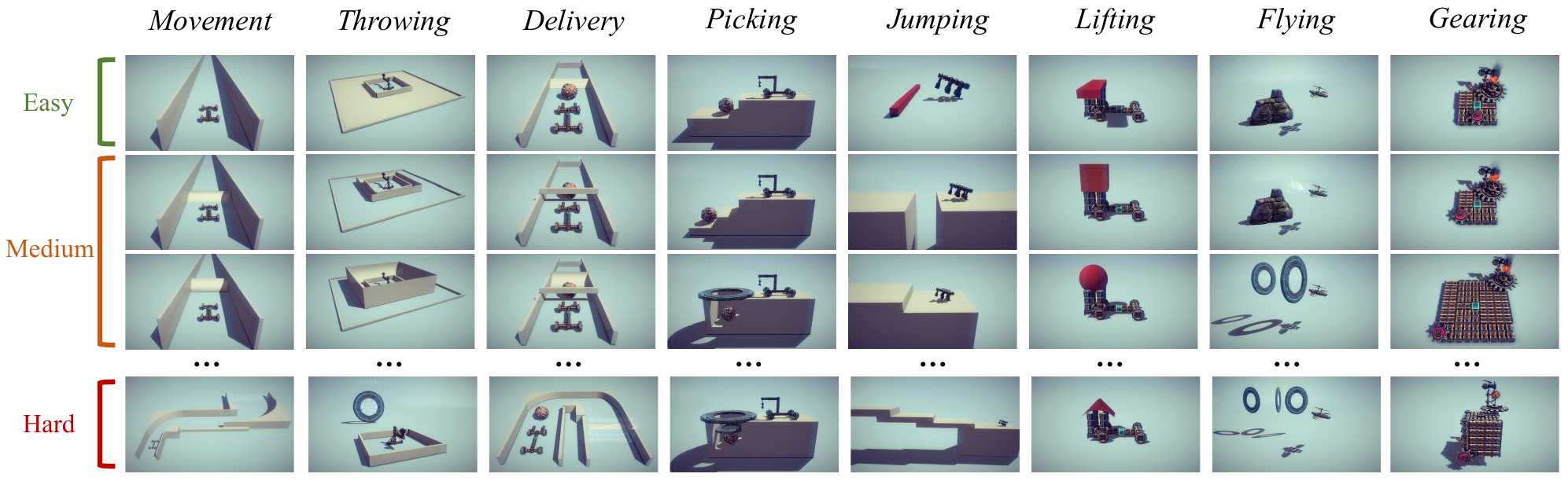}
  \caption{
  \textbf{\textsc{\envname} task illustrations.}
      \textsc{\envname} contains eight task families: \textit{Movement}, \textit{Throwing}, \textit{Delivery}, \textit{Picking}, \textit{Jumping}, \textit{Lifting}, \textit{Flying} and \textit{Gearing}. For each task family, we design multiple instances with different difficulty levels, ranging from \textcolor{mydarkgreen}{easy} to \textcolor{brown}{medium} and \textcolor{red}{hard}. In total, \textsc{\envname} contains 40 task instances.
      }
  \label{fig:BesiegeField_tasks_supp}
\end{figure*}

Fig.~\ref{fig:BesiegeField_tasks_supp} provides additional illustrations of the \textsc{\envname} task families and their difficulty levels.

\subsection{Reward definitions for reinforcement learning}
\label{sec:reward_setting}

For reinforcement learning, we use rewards of the form
\[
R = \mathbbm{1}_{\mathrm{valid}} \cdot P,
\]
where \(\mathbbm{1}_{\mathrm{valid}}\) indicates whether the generated machine satisfies the relevant validity constraints and \(P\) is the task-specific performance score.

\paragraph{Movement.}
For Movement, \(\mathbbm{1}_{\mathrm{valid}}=1\) if the generated construction tree can be parsed and instantiated in \textsc{\envname} without invalid self-collisions; otherwise, \(\mathbbm{1}_{\mathrm{valid}}=0\). The performance score \(P\) is the distance traveled by the root block along the target direction.

\paragraph{Throwing.}
For Throwing, \(\mathbbm{1}_{\mathrm{valid}}\) includes the same parsing and spatial-validity checks as Movement, together with an additional functionality constraint: the boulder must reach a maximum height greater than 3 m during simulation. The performance score \(P\) is the product of the boulder's maximum height and maximum horizontal distance along the target direction. This reward discourages degenerate solutions that optimize only height or only horizontal displacement.

\section{Agentic workflow and search}
\label{sec:agentic_workflow_design}

\begin{figure*}[h]
  \centering
  \includegraphics[width=0.98\linewidth]{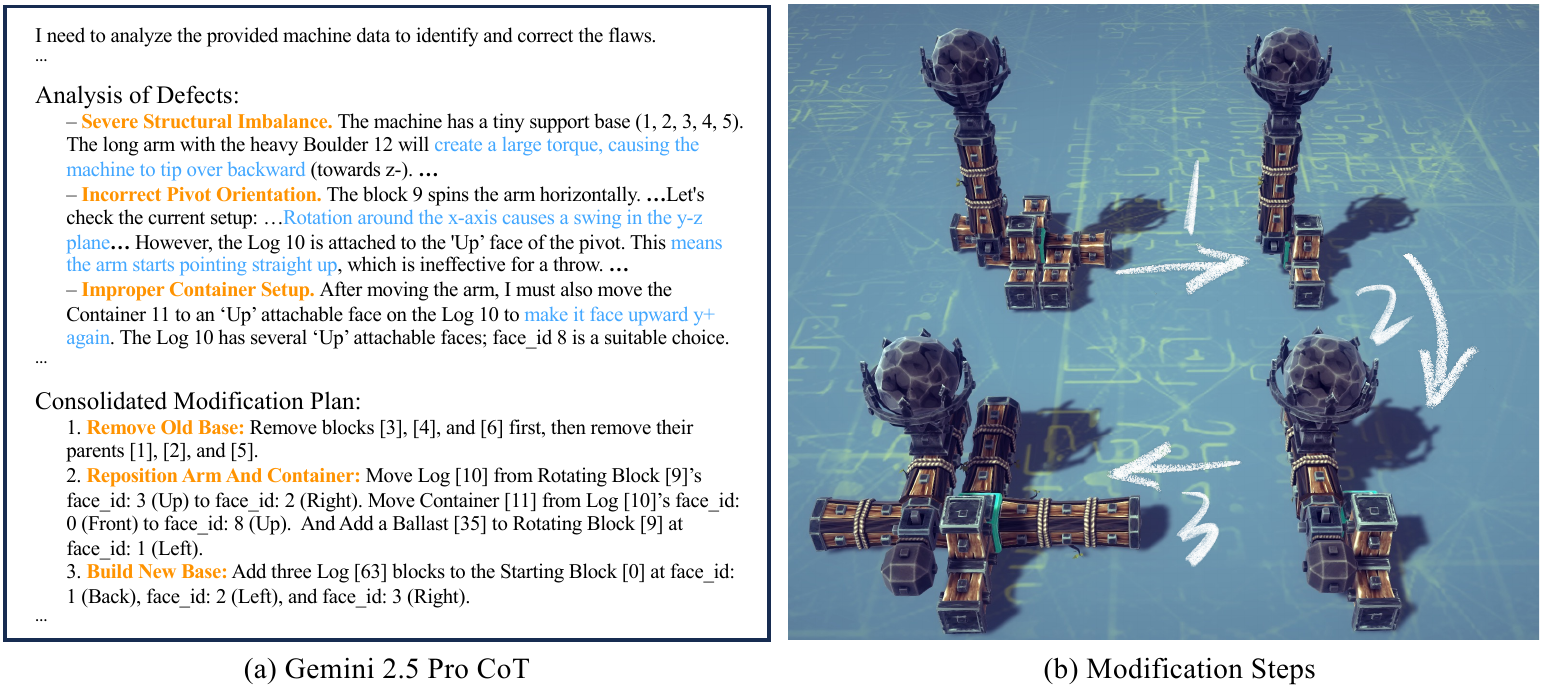}
  \captionsetup{font=footnotesize} %
  \caption{\footnotesize Example CoT of inspector agents (w/ Gemini 2.5 Pro). Blue text highlights the moderate capability of LLMs in spatial reasoning and imagined physical simulation.}
  \label{fig:gemini-cot-example}
\end{figure*}

\begin{figure*}[h]
  \centering
  \includegraphics[width=0.98\linewidth]{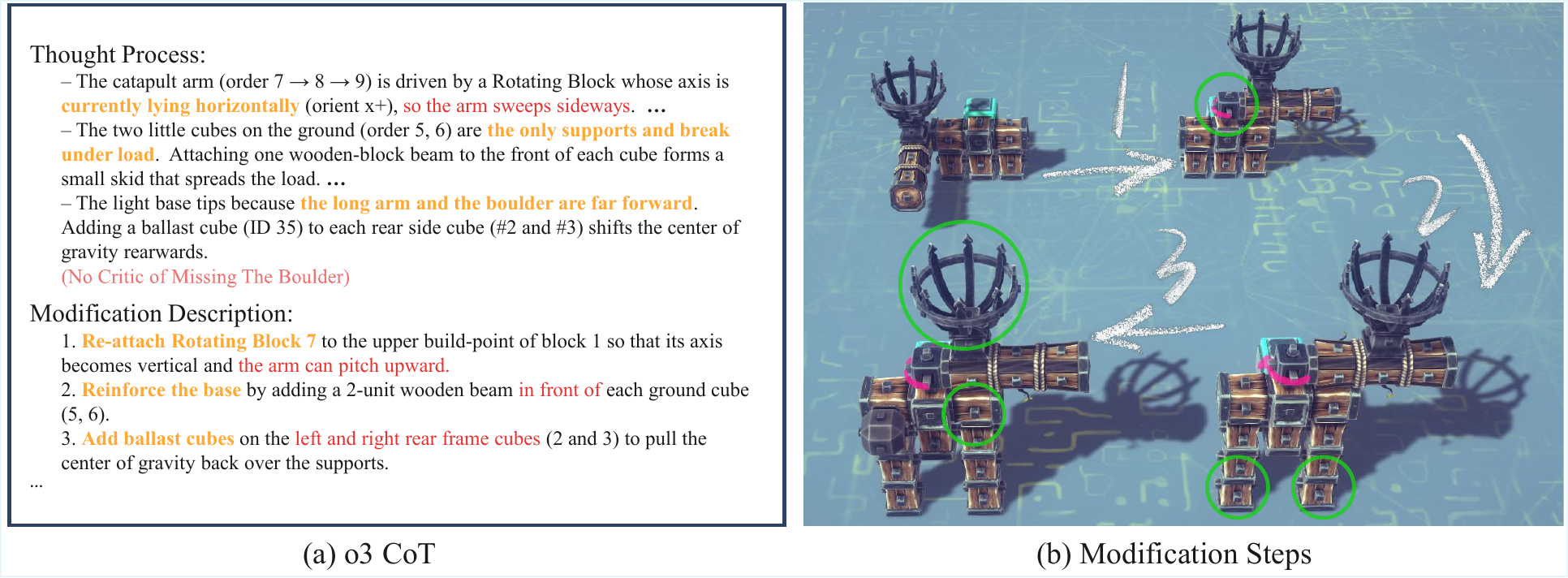}
  \captionsetup{font=footnotesize} %
  \caption{\footnotesize Example CoT of inspector agents (w/ OpenAI o3). Red text highlights reasoning errors.}
  \label{fig:o3-cot-example}
\end{figure*}

This section describes the LLM-agent workflows used to generate and refine machines in \textsc{\envname}. The main paper summarizes the single-agent and hierarchical workflows; here we provide implementation details, including agent roles, hierarchical construction, simulation feedback, active feedback queries and search strategies for iterative refinement.

\subsection{Agent roles}

We use a set of specialized LLM agents to decompose, construct, inspect and refine candidate machines. Each agent receives the relevant task objective, block descriptions, construction syntax and, when applicable, the current machine program and simulation feedback.

\paragraph{Meta-designer.}
The meta-designer receives the task objective and environment constraints, then decomposes the target machine into high-level functional modules. For example, a vehicle may be decomposed into a chassis, wheel assemblies and stabilization components, whereas a catapult may be decomposed into a base, pivoting arm, payload holder and counterweight or actuator. The meta-designer output is used to guide hierarchical construction.

\paragraph{Designer.}
The designer constructs candidate machine programs. In the single-agent setting, the designer directly generates a complete machine program from the task description and block library. In the hierarchical setting, multiple designer agents construct functional modules sequentially, conditioned on the meta-designer blueprint and the partial machine built so far.

\paragraph{Inspector.}
The inspector performs a feedback-free critique after the initial construction stage. It analyzes the constructed machine program and identifies possible high-level structural or conceptual issues, such as missing support, unstable weight distribution, inappropriate actuator placement or inconsistent module connections. The inspector is invoked once as a final pre-refinement check before the machine enters the simulation-guided modification loop, and its report is provided to the refiner.

\paragraph{Refiner.}
The refiner proposes modifications to an existing machine. It receives the current machine program, task objective, edit history, inspector report and selected simulation feedback. The refiner outputs a revised machine program intended to correct failures or improve task performance.

\paragraph{Environment querier.}
The environment querier executes simulations in \textsc{\envname} and converts raw state trajectories into textual feedback for the refiner. Because full block-level trajectories can be too long for the LLM context window, the querier returns a compact summary and may perform targeted follow-up queries for relevant components, time intervals or state variables.

\subsection{Single-agent generation}

In the single-agent setting, a single LLM agent is given the task objective, available block types, construction rules and output format. The model generates a design rationale and a complete construction-tree program. The generated program is parsed, checked for validity and instantiated in \textsc{\envname}. When feedback is enabled, the same agent receives selected simulation feedback and is prompted to revise the machine.

This setting tests whether an LLM can directly map a functional objective to a complete physical assembly without explicit decomposition into modules or specialized agent roles.

\subsection{Iterative editing workflow}

The iterative editing workflow starts from a complete machine generated by a designer agent and then refines it through critique and simulation feedback. Unlike the hierarchical workflow, it does not use a meta-designer or module-wise construction. It therefore tests whether iterative editing alone can improve a directly generated machine.

The workflow has three stages. First, the designer generates an initial design rationale and a complete construction-tree program from the task objective, environment documentation, component descriptions and construction syntax. Second, an inspector performs a feedback-free critique of the generated machine, and the refiner produces a revised machine without observing simulation-derived state feedback; we refer to this stage as blind refinement. Third, the design enters an environment-feedback refinement loop, in which the environment querier runs the machine in \textsc{\envname}, summarizes selected simulation feedback and passes this feedback to the refiner for further modification. The best valid machine discovered during the refinement process is selected as the output. The search procedure used during simulation-guided refinement is described in Sec.~\ref{sec:sim_refinement}.

\subsection{Hierarchical construction}

The hierarchical workflow separates high-level functional decomposition from low-level construction. The meta-designer first decomposes the machine into functional modules and specifies their intended interactions. Designer agents then construct the machine module by module.

For each functional module, we sample multiple candidate constructions in parallel. Invalid candidates are discarded according to the validity checks described in Sec.~\ref{sec:machine_rep}. Valid candidates are retained for the next module. If fewer valid candidates are obtained than the target beam size, valid candidates are duplicated to maintain a fixed number of construction trajectories. This process continues until all functional modules have been constructed or the construction budget is exhausted.

In the main experiments, the hierarchical construction stage maintains eight candidate trajectories. Empirically, the meta-designer typically decomposes a machine into three to four functional modules. The final candidate machines are then passed to the inspector and subsequently to the simulation-guided refinement stage.

\subsection{Simulation feedback and active queries}
\label{sec:env_feedback}

During simulation, \textsc{\envname} can record the full state trajectory of each block, including position, orientation, velocity and block integrity. However, feeding the complete trajectory to an LLM is impractical and often unnecessary. We therefore use a compact feedback interface that returns task-relevant summaries by default, while allowing the environment querier to request additional information when needed.

For Movement, the default feedback includes: (1) machine orientation over time; (2) maximum travel distance along the designated direction; (3) maximum speed; (4) average speed per second; and (5) root-block position sampled every 0.2 s.

For Throwing, the default feedback includes: (1) maximum horizontal distance of the boulder along the designated direction; (2) maximum boulder height; and (3) boulder position sampled every 0.2 s.

Beyond these default summaries, the environment querier may request targeted block-level feedback. Such queries specify a block index, a time interval and one or more state variables of interest. The supported state variables include position, orientation, velocity and, for springs, length. This active-query mechanism allows the agentic workflow to expose detailed physical information only when it is likely to be useful, such as when components collide, break or move in a direction inconsistent with the intended mechanism. When blocks break during simulation, the feedback can include the time and affected component, providing a useful cue for diagnosing structural weakness.

\subsection{Simulation-guided refinement}
\label{sec:sim_refinement}

After initial construction and inspection, each candidate machine enters a simulation-guided refinement loop. The loop alternates between simulation, feedback summarization and machine editing. At each refinement step, the current machine is simulated in \textsc{\envname}, the environment querier extracts relevant feedback, and the refiner generates candidate modifications.

We use an MCTS-style~\cite{mcts_survey} tree-search procedure to organize the refinement process. Each node corresponds to a machine state obtained after one querier--refiner interaction. Expanding a node generates child candidate machines, which are evaluated by simulation. The search then selects promising candidates for further refinement based on their task performance and validity. We use the term ``MCTS-style'' because the procedure follows a tree-based expansion and selection structure, but the environment is not a deterministic board game and each node expansion is performed by an LLM refiner rather than by selecting from a fixed action set.

To reduce invalid generations and prevent child statistics from becoming too sparse, each search node allows up to five retries. In the MCTS-style setting, each expansion generates up to four valid child candidates; if too few valid candidates are found after the retry budget is exhausted, the parent machine is reused as a fallback candidate. The best machine discovered during the search is selected as the final output.

Full algorithmic details are provided in Algorithm~\ref{alg:mcts}. Table~\ref{tab:mcts_ablation} reports how machine performance changes with the number of search rounds, and Fig.~\ref{fig:search_curve} compares the efficiency of different search strategies in the agentic design setting.

\begin{algorithm}
\caption{Monte Carlo Tree Search (MCTS-style refinement)}
\label{alg:mcts}
\begin{algorithmic}[1]
\Require Agentic search node $N$
\Require Root machine $s_0$, maximum iterations $MAX\_ITER$
\Require \textsc{Select}$(node)$: traverse tree using UCB until a leaf node
\Require \textsc{Expand}$(node)$: generate up to four valid child candidates using the LLM refiner
\Require \textsc{Simulate}$(node)$: run \textsc{\envname} simulation and return reward
\Require \textsc{Backpropagate}$(node, reward)$: update visit counts and rewards along the path
\Require \textsc{BestChild}$(node)$: return child with highest simulation score
\Ensure Best machine discovered by the search

\State $root \gets s_0$
\State $max\_retry \gets 5$
\For{$i = 1$ to $MAX\_ITER$}
    \State $node \gets \textsc{Select}(root)$
    \State $retry \gets 0$
    \State $children \gets \emptyset$
    \While{$retry < max\_retry$ and $children$ contains fewer than four valid candidates}
        \State $retry \gets retry + 1$
        \State $candidates \gets \textsc{Expand}(node)$
        \State Keep valid candidates from $candidates$ and add them to $children$
    \EndWhile
    \If{$children = \emptyset$}
        \State $children \gets \{node\}$ \Comment{Fallback to parent machine}
    \EndIf
    \ForAll{$child \in children$}
        \State $reward \gets \textsc{Simulate}(child)$
        \State $\textsc{Backpropagate}(child, reward)$
    \EndFor
\EndFor
\State \Return $\textsc{BestChild}(root)$
\end{algorithmic}
\end{algorithm}

\subsection{Alternative search strategies}

In addition to the MCTS-style refinement procedure used in the main experiments, we evaluate two simpler search strategies.

\paragraph{Best-of-\(N\).}
The best-of-\(N\) strategy samples \(N\) candidate modifications from the refiner and selects the candidate with the highest simulation score. Unlike the MCTS-style procedure, best-of-\(N\) does not maintain a tree of machine states across refinement rounds. Full details are provided in Algorithm~\ref{alg:bestofn}.

\paragraph{Random search.}
The random-search baseline follows the same candidate-generation procedure as best-of-\(N\), but selects a random valid candidate instead of the best-performing one. This baseline controls for the benefit of sampling multiple candidates without using performance-based selection. Full details are provided in Algorithm~\ref{alg:random-search}.

\begin{figure*}[h!]
  \centering
  \includegraphics[width=\linewidth]{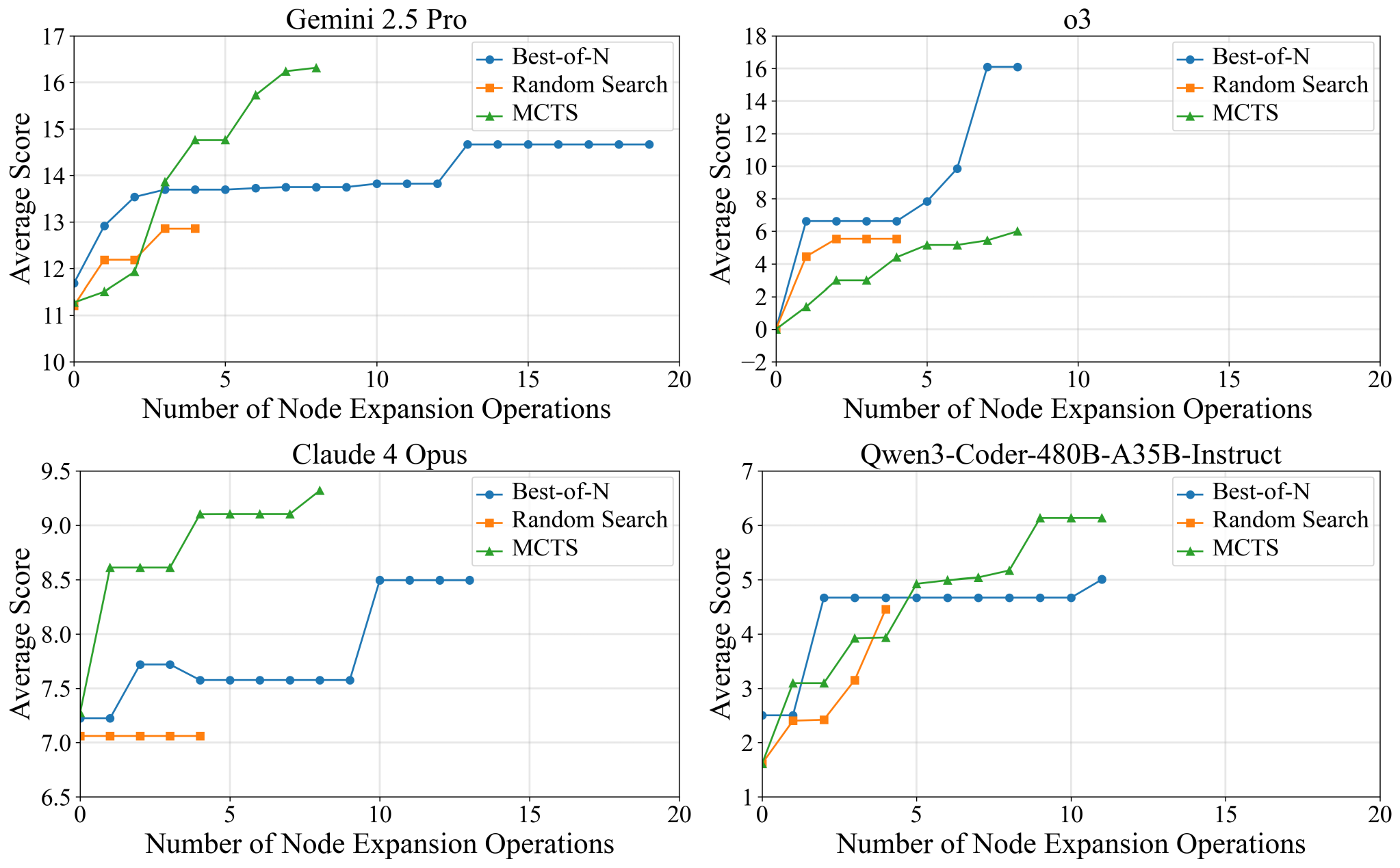}
  \caption{
  Variation in average machine score as the number of LLM node-expansion operations increases under different search strategies.
  }
  \label{fig:search_curve}
\end{figure*}

The full comparison across search strategies is reported in Table~\ref{tab:abl:env-search}, and the corresponding score trends as a function of node expansions are shown in Fig.~\ref{fig:search_curve}. These results help separate the benefit of iterative refinement from the benefit of explicit selection using simulation scores.

\begin{algorithm}
\caption{Random Search}
\label{alg:random-search}
\begin{algorithmic}[1]
\Require Agentic search node $N$
\Require Scoring function $S$, validity check function $F$
\Require Search rounds $R$
\Ensure Final machine and score

\State Input machine $ori\_machine$
\State $max\_retry \gets 5$
\State $machine\_last\_round \gets ori\_machine$

\For{$r=1$ to $R$}
    \State $retry \gets 0$
    \Repeat
        \State $retry \gets retry + 1$
        \State $machine\_next\_round \gets N.generate(machine\_last\_round)$
    \Until{$F(machine\_next\_round)$ or $retry \geq max\_retry$}
    \If{not $F(machine\_next\_round)$}
        \State $machine\_next\_round \gets machine\_last\_round$
    \EndIf
    \State $score \gets S(machine\_next\_round)$
    \State $machine\_last\_round \gets machine\_next\_round$
\EndFor

\State \Return $(machine\_last\_round, score)$
\end{algorithmic}
\end{algorithm}

\begin{algorithm}
\caption{Best-of-\(N\)}
\label{alg:bestofn}
\begin{algorithmic}[1]
\Require Agentic search node $N$
\Require Scoring function $S$, validity check function $F$
\Require Search rounds $R$, number of samples $n$
\Ensure Best machine and score

\State Input machine $ori\_machine$
\State $best\_machine \gets ori\_machine$
\State $best\_score \gets S(ori\_machine)$
\State $max\_retry \gets 5$

\For{$r=1$ to $R$}
    \State $round\_best\_machine \gets best\_machine$
    \State $round\_best\_score \gets best\_score$
    \For{$i=1$ to $n$}
        \State $retry \gets 0$
        \Repeat
            \State $retry \gets retry + 1$
            \State $machine_i \gets N.generate(best\_machine)$
        \Until{$F(machine_i)$ or $retry \geq max\_retry$}
        \If{not $F(machine_i)$}
            \State $machine_i \gets best\_machine$
        \EndIf
        \State $score_i \gets S(machine_i)$
        \If{$score_i > round\_best\_score$}
            \State $round\_best\_score \gets score_i$
            \State $round\_best\_machine \gets machine_i$
        \EndIf
    \EndFor
    \State $best\_machine \gets round\_best\_machine$
    \State $best\_score \gets round\_best\_score$
\EndFor

\State \Return $(best\_machine, best\_score)$
\end{algorithmic}
\end{algorithm}

\begin{table*}[h!]
  \scriptsize
  \centering
  \setlength{\tabcolsep}{6pt}
  \renewcommand{\arraystretch}{1.2}
  \begin{tabularx}{\textwidth}{l*{9}{>{\centering\arraybackslash}X}}
    \toprule
    \multirow{2}{*}{Models} & \multicolumn{3}{c}{Random Search} & \multicolumn{3}{c}{Best-of-\(N\)} & \multicolumn{3}{c}{MCTS} \\
    \cmidrule(lr){2-4} \cmidrule(lr){5-7} \cmidrule(lr){8-10}
     & Avg.I & Mean  & Max & Avg.I & Mean  & Max & Avg.I & Mean  & Max \\
    \midrule
    Gemini 2.5 Pro        & 5 & 15.02 & 20.50 & 20 & 14.67 & 16.66 & 8   & 15.73 & 18.19 \\
    Claude Opus 4         & 5 & 7.67  & 7.88  & 18 & 8.18  & 8.50  & 6   & 9.10  & 9.32  \\
    o3                    & 5 & 7.71  & 11.94 & 8  & 10.60 & 15.07 & 7   & 5.34  & 11.11 \\
    Qwen3-Coder-480B-A35B & 5 & 4.50  & 7.64  & 11 & 5.61  & 9.87  & 8.5 & 5.21  & 6.52  \\
    \bottomrule
  \end{tabularx}
  \caption{
  Ablation study on different search strategies. We compare final scores of the agentic workflow under Random Search, Best-of-\(N\) and MCTS-style refinement. MCTS-style refinement is executed for 5 rounds, with Random Search and Best-of-\(N\) also run for the same number of rounds. Avg.I denotes the average number of node expansions per search round.
  }
  \label{tab:abl:env-search}
\end{table*}

\begin{table*}[h!]
  \scriptsize
  \centering
  \setlength{\tabcolsep}{6pt}
  \renewcommand{\arraystretch}{1.2}
  \begin{tabularx}{\textwidth}{l*{6}{>{\centering\arraybackslash}X}}
    \toprule
    \multirow{2}{*}{Models} & \multicolumn{2}{c}{R2} & \multicolumn{2}{c}{R5} & \multicolumn{2}{c}{R10} \\
    \cmidrule(lr){2-3} \cmidrule(lr){4-5} \cmidrule(lr){6-7}
     & Mean & Max & Mean & Max & Mean & Max \\
    \midrule
    Gemini 2.5 Pro        & 15.04 & 17.31 & 15.73 & 18.19 & 16.44 & 18.19 \\
    Claude Opus 4         & 8.61  & 9.32  & 9.10  & 9.32  & 9.43  & 9.98  \\
    o3                    & 5.33  & 11.11 & 5.34  & 11.11 & 8.46  & 14.52 \\
    Qwen3-Coder-480B-A35B & 5.18  & 6.52  & 5.21  & 6.52  & 5.74  & 6.52  \\
    \bottomrule
  \end{tabularx}
  \caption{
  Ablation study on the effect of search depth in MCTS-style refinement. R2, R5 and R10 denote 2, 5 and 10 rounds of MCTS-style search on the same search tree.
  }
  \label{tab:mcts_ablation}
\end{table*}

\subsection{Edit history}

When edit history is enabled, the refiner receives a compact record of previous modifications and their corresponding outcomes. This history helps prevent repeated invalid edits and repeated reversals between similar machine states. Because full histories can become long, we provide a summarized history rather than raw previous prompts and responses. The effect of edit history is reported in the ablation studies in Sec.~\ref{sec:benchmark_ablations}.

\subsection{Model and decoding settings}

For each LLM evaluated in the agentic benchmark, we use the same task description, block library and construction rules. Unless otherwise stated, models are sampled with the decoding settings reported in the full benchmark tables. Invalid outputs are retried according to the retry budget specified for each stage. The complete prompt templates for the meta-designer, designer, inspector, refiner and environment querier are provided in Sec.~\ref{sec:prompts}.

\subsection{Diagnosing failures in physical machine construction}\label{app:diag-failure-cases}
\begin{figure*}[h!]
  \centering
  \includegraphics[width=\linewidth]{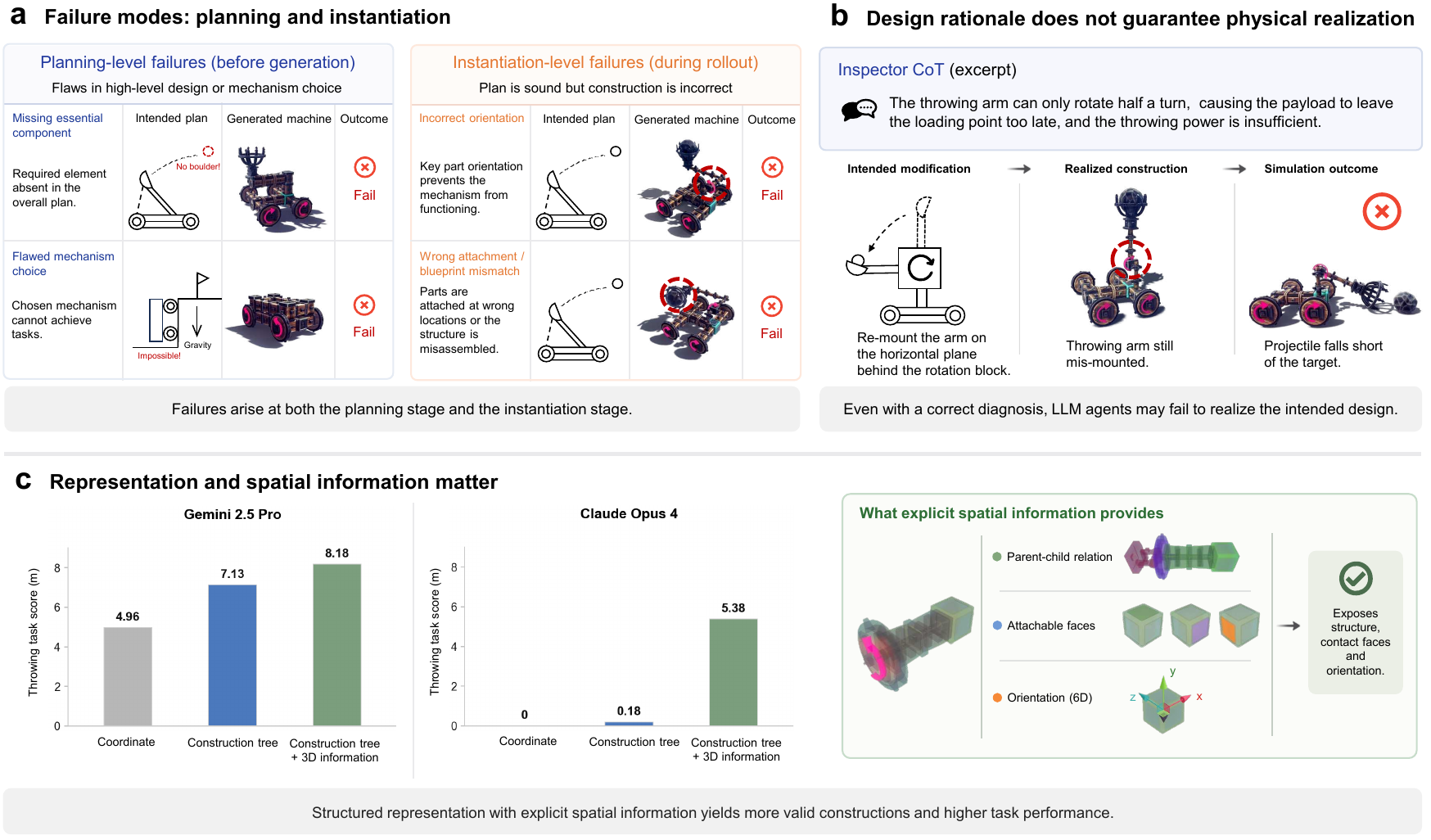}
    \caption{
        \textbf{Failure modes and the grounding gap in LLM machine design.}
        \textbf{a,} LLM-generated machines fail at both the planning and construction stages. Planning-level failures include missing essential components or choosing an unsuitable mechanism. Construction-level failures occur when a plausible plan is turned into incorrect orientations, attachments or spatial configurations.
        \textbf{b,} A correct design rationale does not guarantee physical realization. Even when the inspector identifies the intended modification, the realized construction may still misalign key components, leading to failed simulation outcomes.
        \textbf{c,} Machine representation and explicit spatial information matter. Compared with coordinate-only representations, construction trees improve both normalized task score and valid construction rate, and adding explicit spatial information about parent--child relations, attachable faces and orientation further improves performance.
    }
  \label{fig:failure-mode}
\end{figure*}

We analyze common failure modes in LLM-generated machines. In Figure~\ref{fig:failure-mode}a, failures occur at both the planning and construction levels. Some designs omit essential components or rely on flawed physical mechanisms. Others describe plausible high-level plans but fail during construction through incorrect part orientations, inappropriate parent attachments or incomplete adherence to the intended blueprint. The bottleneck is therefore not only whether a model can describe a plausible mechanism in language, but whether it can turn that mechanism into a spatially and mechanically consistent construction tree.

CoT, here referring to the model's written intermediate reasoning about the design, often contains a reasonable high-level plan, yet the final machine may deviate from the intended structure or build it incorrectly. Figure~\ref{fig:failure-mode}b illustrates this gap. When other LLMs are given Gemini 2.5 Pro-generated CoT and asked to output construction trees directly, the resulting machines become more visually consistent with the intended mechanism, but precise physical assembly remains difficult. This separates high-level planning from low-level spatial construction as distinct sources of error.

Representation and spatial-information ablations reinforce this interpretation. Here, parsed spatial information denotes explicit summaries of parent--child relations, attachable faces and component orientations extracted from the construction tree. Figure~\ref{fig:failure-mode}c shows that construction-tree representations outperform coordinate-only descriptions (Table~\ref{tab:rep_comparison}), and that adding parsed spatial information further improves validity and task score (Table~\ref{tab:abl:machine-3d-info}). These results show that the model interface matters: explicit assembly relations and spatial summaries improve design reasoning, while construction trees alone do not guarantee that the model can infer all relevant physical relationships.

\subsection{Coding-agent harness, visual feedback and no-LLM baselines}
We additionally benchmark a modern coding-agent harness (Codex 5.5~\cite{openai2025codex}) under the same three-stage protocol, with and without an agent-controlled vision variant, and a no-LLM Random-DSL baseline that samples random valid construction trees and applies random legal DSL edits, selecting candidates purely by simulation reward under a matched rollout budget. Codex is strong on Movement but remains much weaker on Throwing, and Random-DSL stays below LLM initialization, indicating that general coding-agent ability and simulator-guided search alone do not solve mechanism-level machine design.

\begin{table*}[h!]
  \scriptsize
  \centering
  \setlength{\tabcolsep}{6pt}
  \renewcommand{\arraystretch}{1.2}
  \begin{tabularx}{\textwidth}{l*{9}{>{\centering\arraybackslash}X}}
    \toprule
    \multirow{2}{*}{Models}
    & \multicolumn{3}{c}{Initial Construction}
    & \multicolumn{3}{c}{Blind Refinement}
    & \multicolumn{3}{c}{Env. Feedback Modification} \\
    \cmidrule(lr){2-4} \cmidrule(lr){5-7} \cmidrule(lr){8-10}
    & Valid & Mean & Max & Valid & Mean & Max & Valid & Mean & Max \\
    \midrule
    \multicolumn{10}{l}{\textit{Throwing}} \\
    Gemini 2.5 Pro        & 5 & 8.49  & 9.14  & 5 & 8.18  & 11.07 & 5 & 15.73 & 18.19 \\
    Codex 5.5 (vision)    & 6\textsuperscript{a} & 1.44  & 2.71 & 3\textsuperscript{a} & 0.99 & 1.77 & 3\textsuperscript{b} & 5.92 & 6.56 \\
    Codex 5.5 (no vision) & 8 & 0.25  & 0.48  & 6 & 0.24  & 0.74  & 6 & 1.40  & 6.39 \\
    Random-DSL            & 8 & 0.90  & 5.69  & 8 & 0.00  & 0.00  & 8 & 0.91  & 4.60 \\
    \midrule
    \multicolumn{10}{l}{\textit{Movement}} \\
    Gemini 2.5 Pro        & 7 & 27.85 & 38.06 & 7 & 17.81 & 39.64 & 7 & 29.96 & 41.52 \\
    Codex 5.5 (vision)    & 7\textsuperscript{a} & 44.21 & 46.60 & 4\textsuperscript{a} & 21.73 & 46.30 & 4\textsuperscript{c} & 47.98 & 48.60 \\
    Codex 5.5 (no vision) & 8 & 43.29 & 46.30 & 2 & 46.90 & 47.20 & 2 & 47.10 & 47.30 \\
    Random-DSL            & 8 & 5.23  & 12.30 & 8 & 4.06  & 11.20 & 8 & 10.40 & 14.80 \\
    \bottomrule
  \end{tabularx}
  \caption{
  \textbf{Stage-wise performance of coding-agent harness and a no-LLM DSL-search baseline.} Each cell reports the number of valid machines, followed by the mean and maximum simulation score among valid machines, after the initial construction stage, blind refinement without environment feedback, and modification with environment feedback. All conditions use the same three-stage protocol and a matched rollout budget of up to eight candidate machines per stage. For the agent-controlled vision variant, \textsuperscript{a} marks stages without vision calls, \textsuperscript{b} 16 vision calls, and \textsuperscript{c} 22 vision calls. Random-DSL samples valid construction trees uniformly at random and applies random legal DSL edits (add/remove/swap a block, reassign a parent face, change orientation, adjust actuation), selecting candidates purely by simulation reward.
  }
  \label{tab:codex_random_dsl}
\end{table*}

\section{Benchmark metrics and additional ablations}
\label{sec:benchmark_ablations}

This section provides metric definitions, benchmark-result references and additional ablations supporting the main results. The main paper reports the central trends; here we provide additional details on validity metrics, agentic-workflow components, feedback settings and search strategies.

\subsection{Evaluation metrics}

We evaluate generated machines using both validity metrics and task-performance metrics.

\paragraph{Parse validity.}
Parse validity is the proportion of generated machine programs that can be parsed into a construction tree with the required fields, valid component identifiers and valid parent references. Outputs that are syntactically malformed, omit required fields or reference nonexistent parent components are counted as parse-invalid.

\paragraph{Spatial validity.}
Spatial validity is the proportion of parsed machines that can be built in \textsc{\envname} without violating spatial constraints. Invalid spatial configurations include severe self-intersections, illegal attachment relations and assemblies that cannot be loaded into the simulator.

\paragraph{Machine validity.}
Machine validity is the proportion of generated machines that satisfy both parse validity and spatial validity. Unless otherwise stated, invalid machines receive zero task reward.

\paragraph{Simulation score.}
For each valid machine, \textsc{\envname} executes the simulation and computes a task-specific score from the resulting state trajectory. We report both the mean score across generated candidates and the maximum score achieved under the evaluation budget. The maximum score is especially relevant for design tasks because practical use often permits sampling or iterative search and then selecting the best candidate.

\paragraph{Best@\(N\).}
For analyses involving inference-time scaling, we report Best@\(N\), the best score obtained among \(N\) generated candidates under the same task and model setting. This metric reflects the performance available when multiple candidate machines can be generated and evaluated before selection.

\subsection{Model versions}
\label{sec:model_versions}

Table~\ref{tab:model_versions} lists the model snapshots used in the experiments. We report exact model identifiers when available.

\begin{table}[h]
\centering
\resizebox{\linewidth}{!}{
\begin{tabular}{ll}
\toprule
Model name in paper & Model snapshot or identifier \\
\midrule
Claude Opus 4 & claude-opus-4-20250514 \\
Gemini 2.5 Pro & gemini-2.5-pro-preview-03-25 \\
OpenAI o3 & o3-2025-04-16 \\
DeepSeek V3 & deepseek-chat-v3.1 \\
Doubao Seed 1.6 & doubao-seed-1-6-250615 \\
Kimi K2 & kimi-k2-0711-preview \\
Qwen3-Coder & Qwen3-Coder-480B-A35B \\
GPT-5 & gpt-5-2025-08-07 \\
Llama 4 Scout & Llama 4 Scout 17B 16E \\
\bottomrule
\end{tabular}
}
\caption{Model snapshots used in the experiments.}
\label{tab:model_versions}
\end{table}

\subsection{Full benchmark results}

The main LLM-agent benchmark evaluates seven task families in \textsc{\envname}: Movement, Throwing, Delivery, Jumping, Lifting, Flying and Gearing. Picking is implemented but excluded from the main benchmark for the reasons described in Sec.~\ref{sec:tasks_rewards}. Full numerical results for the representative Movement and Throwing settings are reported in Table~\ref{tab:test_time_scaling_benchmark}. Additional task-family results for Flying, Jumping, Delivery, Lifting and Gearing are reported in Table~\ref{tab:additional_tasks}.

\begin{table*}[t]
  \scriptsize
  \centering
  \setlength{\tabcolsep}{4pt}
  \renewcommand{\arraystretch}{1.2}
  \begin{tabularx}{\textwidth}{l*{9}{>{\centering\arraybackslash}X}}
    \toprule
    \multirow{2}{*}{Models} & \multicolumn{3}{c}{Meta-Designer \& Designer} & \multicolumn{3}{c}{Blind Refinement} & \multicolumn{3}{c}{Modification w/ Env Feedback} %
    \\
    \cmidrule(lr){2-4} \cmidrule(lr){5-7} \cmidrule(lr){8-10}
     & Valid & Mean & Max & Valid & Mean & Max & Valid & Mean & Max \\
    \midrule
    \multicolumn{10}{l}{\textit{Throwing}} \\
    \midrule
    Gemini 2.5 Pro        &5&8.49&9.14 &5&8.18&11.07&5&15.73&18.19\\
    Claude Opus 4            &4&4.17&4.36&2&5.38&5.8&2&9.10&9.32\\
    o3                       &3&0&0&3&0&0&3&5.34&11.11\\
    Qwen3-Coder-480B-A35B    &6&0.75&4.5&6&1.61&5.0&6&5.21&6.52\\
    Doubao Seed 1.6           &3&4.34&4.37&3&0.31&0.49&3&4.62&4.76 \\
    DeepSeek-V3                &7&0&0&7&0.98&3.18&4&4.82&4.93\\
    Kimi K2                &6&7.18&12.02&2&5.29&7.36&0&-&-\\
    GPT-5                   &8&3.87&4.92&1&5.35&5.35&0&-&-\\
    Llama 4 Scout 17B 16E   &8&3.59&11.83&0&-&-&0&-&-\\
    \midrule
    \multicolumn{10}{l}{\textit{Movement}} \\
    \midrule
    Gemini 2.5 Pro        &7&27.85&38.06&7&17.81&39.64&7&29.96&41.52\\
    Claude Opus 4       &3&2.96&3.41&3&36.18&37.05&3&26.59&38.67  \\
    o3                    &8&29.27&41.43&2&20.03&40.04&2&28.39&36.18\\
    Qwen3-Coder-480B-A35B  &6&5.43&8.72&6&3.90&11.25&6&11.75&34.05\\
    Doubao Seed 1.6       &5&21.80&29.91&5&13.25&26.05&5&18.75&26.02 \\
    DeepSeek-V3           &3&0.27&0.47&3&16.94&29.87&3&17.92&31.94  \\
    Kimi K2               &1&6.74&6.74&1&0.39&0.39&1&14.99&14.99  \\
    GPT-5                 &8&5.67&20.32&8&3.75&9.65&8&8.43&13.72 \\
    Llama 4 Scout 17B 16E &4&1.55&2.00&1&0.47&0.47&1&0.47&0.47 \\
    \bottomrule
  \end{tabularx}
  \caption{
      \textbf{Stage-wise performance of the hierarchical agentic workflow.}
      We report results on the Throwing and Movement tasks after three stages: the initial meta-designer--designer construction stage, inspector--refiner blind refinement without environment feedback, and hierarchical design with environment feedback. Each cell reports the number of valid machines, followed by the mean and maximum simulation score among valid machines. Mean scores are computed only over valid machines. The initial stage generates up to eight candidate machines, and only valid machines are retained for subsequent refinement stages.
    }
  \label{tab:test_time_scaling_benchmark}
\end{table*}

\begin{table*}[t]
  \scriptsize
  \centering
  \setlength{\tabcolsep}{3pt}
  \renewcommand{\arraystretch}{1.1}
  \newcommand{\cgr}[1]{\textcolor[rgb]{.329, .51, .208}{\textbf{#1}}}
  \newcommand{\cre}[1]{\textcolor[rgb]{1, 0, 0}{\textbf{#1}}}

  \begin{tabularx}{\textwidth}{lCCC>{\centering\arraybackslash}p{1.5cm}>{\centering\arraybackslash}p{1.5cm}}
    \toprule
    \multirow{2}{*}{Models}
    & \multicolumn{3}{c}{Machine Validity (pass/total)}
    & \multicolumn{2}{c}{Designer Score} \\
    \cmidrule(lr){2-4}\cmidrule(lr){5-6}
    & File Valid & 3D Valid & Final Valid & Mean & Max \\
    \midrule
    \multicolumn{6}{l}{\textit{Baseline (Meta-Designer \& Designer)}} \\
    \midrule
    Gemini 2.5 Pro    & 8/8 & 5/8 & 5/8 & 8.49 & 9.14 \\
    o3            & 3/8 & 3/3 & 3/8 & 0      & 0 \\
    Qwen3-Coder-480B-A35B    & 8/8 & 6/8 & 6/8 & 0.75 & 4.5 \\
    Doubao Seed 1.6    & 7/8 & 3/7 & 3/8 & 4.34 & 4.37 \\
    Claude Opus 4      & 8/8 & 4/8 & 4/8 & 4.17 & 4.36 \\
    DeepSeek-V3   & 7/8 & 7/7 & 7/8 & 0    & 0 \\
    Kimi K2       & 6/8 & 6/6 & 6/8 & 7.18 & 12.02 \\
    Llama 4 Scout 17B 16E        & 8/8 & 8/8 & 8/8 & 3.59      & 11.83 \\
    \midrule
    \multicolumn{6}{l}{\textit{Single Agent}} \\
    \midrule
    Gemini 2.5 Pro    & 6/8 & 3/6 & 3/8 & 6.13 & 9.00 \\
    o3            & 8/8 & 8/8 & 8/8 & 2.87 & 5.22 \\
    Qwen3-Coder-480B-A35B    & 8/8 & 4/8 & 4/8 & 3.5 & 9.24 \\
    Doubao Seed 1.6    & 7/8 & 6/7 & 6/8 & 4.24 & 8.2 \\
    Claude Opus 4      & 8/8 & 2/6 & 2/8 & 4.76 & 4.91 \\
    DeepSeek-V3   & 7/8 & 6/7 & 6/8 & 4.67 & 4.86 \\
    Kimi K2       & 8/8 & 3/8 & 3/8 & 6.85 & 9.05 \\
    Llama 4 Scout 17B 16E        & 8/8 & 7/8 & 7/8 & 3.63 & 5.64 \\
    \midrule
    \multicolumn{6}{l}{\textit{w/ detailed Meta-Designer \& Designer }} \\
    \midrule
    Gemini 2.5 Pro    & 8/8 & 7/8 & 7/8 & 9.19 &11.94 \\
    o3            & 7/8 & 6/7 & 6/8 & 0.92& 1.18 \\
    Qwen3-Coder-480B-A35B    & 8/8 & 2/8 & 2/8 & 4.87 & 4.87 \\
    Doubao Seed 1.6    & 0/8 & -   & -   & - & - \\
    Claude Opus 4      & 7/8 & 5/7 & 5/8 & 4.13 & 4.79 \\
    DeepSeek-V3   & 8/8 & 8/8 & 8/8 & 6.12 & 9.0 \\
    Kimi K2       & 8/8 & 0/8 & -   & - & - \\
    Llama 4 Scout 17B 16E        & 8/8 & 7/8 & 7/8 & 4.01 & 6.93 \\
    \bottomrule
  \end{tabularx}
  \caption{
      \textbf{Ablation on the meta-designer.}
      We compare the hierarchical baseline with a single-agent setting and a detailed meta-designer setting that provides more explicit step-by-step construction guidance. For each condition, we report parse validity, spatial validity, machine validity, and the mean and maximum simulation score. Spatial validity is evaluated only for parse-valid outputs, while machine validity requires both parse validity and spatial validity. Mean scores are computed only over machine-valid outputs.
    }
  \label{tab:ablation_meta_designer}
\end{table*}

\subsection{Meta-designer ablation}
\begin{figure*}[h!]
  \centering
  \includegraphics[width=\linewidth]{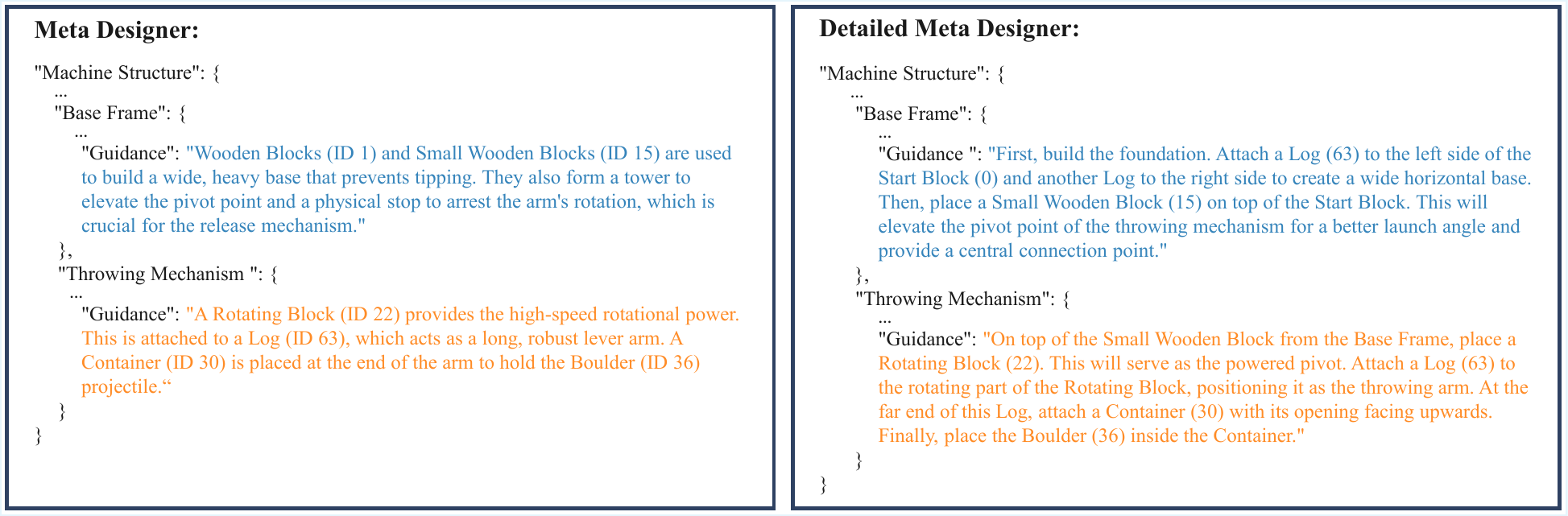}
  \caption{\footnotesize Construction guidance comparison of \textit{Meta-designer} and \textit{Detailed Meta-designer}, sampled with Gemini 2.5 Pro.}
  \label{fig:detailed-meta-designer}
\end{figure*}

The meta-designer is responsible for decomposing a high-level task into functional modules before low-level construction. In Table~\ref{tab:ablation_meta_designer}, we compare hierarchical workflows with and without the meta-designer stage.

The meta-designer can improve performance when the base model is strong enough to produce and follow meaningful module-level plans. For models such as Gemini 2.5 Pro, the decomposition often identifies useful macro-level components, such as a chassis, launcher arm, pivot or support frame, which can then be constructed separately by designer agents. However, introducing an additional planning stage can also introduce compounding errors. This may explain some cases in which hierarchical design reduces performance: module-level plans can be plausible in language while still failing to produce mechanically consistent assemblies.

We also evaluate whether the meta-designer should provide detailed step-by-step construction instructions or only high-level functional-module descriptions. Table~\ref{tab:ablation_meta_designer} reports the quantitative comparison, and Fig.~\ref{fig:detailed-meta-designer} illustrates the difference between the meta-designer and detailed meta-designer prompts. Detailed meta-designer outputs are most useful when the downstream model can reliably interpret and build from them. For less capable models, overly detailed instructions may increase the burden of instruction following and lead to additional errors.

\subsection{Inspector ablation}

The inspector performs a feedback-free critique after initial construction and before simulation-guided refinement. In Table~\ref{tab:ablation_inspector}, we compare workflows with and without this inspection stage.

The inspector can improve performance for stronger models by identifying high-level structural issues before expensive refinement. For example, it may flag missing support, unstable mass distribution, inappropriate actuator placement or an implausible mechanical chain. However, the inspector can also be harmful if its critique is incorrect or overly confident. In such cases, the refiner may modify a partially functional machine according to a flawed diagnosis, reducing performance. This effect is visible for some models whose self-critique quality is weaker than their direct construction ability.

\begin{table*}[h!]
  \scriptsize
  \centering
  \setlength{\tabcolsep}{3pt}
  \renewcommand{\arraystretch}{1.1}
  \newcommand{\cgr}[1]{\textcolor[rgb]{.329, .51, .208}{\textbf{#1}}}
  \newcommand{\cre}[1]{\textcolor[rgb]{1, 0, 0}{\textbf{#1}}}

  \begin{tabularx}{\textwidth}{l*{6}{>{\centering\arraybackslash}X}}
    \toprule
    \multicolumn{1}{c}{\multirow{2.4}{*}{Models}}
    & \multicolumn{6}{c}{Blind Refinement Simulation Scores} \\
    \cmidrule(lr){2-7}
    & \multicolumn{3}{c}{Baseline}
    & \multicolumn{3}{c}{w/o Inspector} \\
    \cmidrule(lr){2-4}\cmidrule(lr){5-7}
    & Valid & Mean & Max & Valid & Mean & Max \\
    \midrule
    Gemini 2.5 Pro
       & 5 & \bf 8.18 & \bf 11.07
       & 5 & 5.67 & 9.37 \\
    o3
       & 3 &0 & 0
       & 3 &\bf 3.08  & \bf 9.24 \\
    Qwen3-Coder-480B-A35B
       & 6 & \bf 1.61 & \bf 5.0
       & 6 & 0.75 & 4.51\\
    Doubao Seed 1.6
       & 3 & 0.31 & \bf 0.49
       & 3 & \bf 0.47 &1.41  \\
    Claude Opus 4
       & 2 &\bf 5.38  &5.8 
       & 4 &5.20  & \bf 8.25  \\
    DeepSeek-V3
       & 7 & \bf 0.98 & \bf 3.18
       & 7 &0.38  &2.16  \\
    Kimi K2
       & 2 & \bf 5.29 &7.36 
       & 6 & 2.31 & \bf 8.91 \\
    Llama 4 Scout 17B 16E
       & 0 & - & -
       & 0 & - & - \\
    \bottomrule
  \end{tabularx}
  \caption{
  Ablation study on the inspector agent. The mean simulation score is calculated only over valid machines after blind refinement. The effect of the inspector is model-dependent: it improves mean performance for some models, such as Gemini 2.5 Pro, but can reduce performance when the feedback-free critique is unreliable.
  }
  \label{tab:ablation_inspector}
\end{table*}

\subsection{Effect of environment feedback}

We compare three feedback settings: active feedback with the environment querier, basic feedback without the environment querier and score-only feedback. In the active-feedback setting, the refiner receives simulation scores, basic task feedback and targeted state information requested by the environment querier. In the basic-feedback setting, the refiner receives simulation scores and default task feedback but no targeted querier-requested information. In the score-only setting, the refiner receives only the scalar simulation score.

As shown in Table~\ref{tab:rl_env_feedback}, active feedback generally improves performance relative to both basic feedback and score-only settings. This suggests that simulation outcomes contain useful information beyond scalar task rewards, but that the information must be distilled into an interpretable form for the LLM. Raw state trajectories are too long and low-level to provide directly; selected summaries, such as object trajectories, machine orientation, breakage events and targeted block states, are more useful for refinement.

\begin{table*}[t]
\scriptsize
  \centering
  \setlength{\tabcolsep}{3pt}
  \renewcommand{\arraystretch}{1.1}
  \newcommand{\cgr}[1]{\textcolor[rgb]{.329, .51, .208}{\textbf{#1}}}
  \newcommand{\cre}[1]{\textcolor[rgb]{1, 0, 0}{\textbf{#1}}}

  \begin{tabularx}{\textwidth}{l*{9}{>{\centering\arraybackslash}X}}
    \toprule
    \multicolumn{1}{c}{\multirow{2.4}{*}{Models}}
    & \multicolumn{9}{c}{Refiner Simulation Scores} \\
    \cmidrule(lr){2-10}
    & \multicolumn{3}{c}{Baseline (Full feedback)}
    & \multicolumn{3}{c}{w/o Env. Querier}
    & \multicolumn{3}{c}{Score Only}\\
    \cmidrule(lr){2-4}\cmidrule(lr){5-7}\cmidrule(lr){8-10}
    & std & Mean & Max & std & Mean & Max& std & Mean & Max \\
    \midrule
    Gemini 2.5 Pro
       & 2.47 &\bf 15.73  &18.19 
       & 4.18 &14.89  &\bf 19.77
       & 2.05 &9.68  &13.18\\
    o3
       & 5.36 &5.34  & \bf 11.11
       & 4.24 &4.06  &8.55
       & 2.76 &\bf 7.05  &10.24\\
    Qwen3-Coder-480B-A35B
       & 0.95 & \bf 5.21 & \bf 6.52
       & 2.32 &4.05 &6.89 
       & 2.53 &2.81 &5.56\\
    Claude Opus 4
       & 0.31 &\bf 9.10 & \bf 9.32
       & 0.82 &8.50  &9.08
       & 0.42 &5.75  &6.05\\
    Doubao Seed 1.6
       & 0.23 &4.62  & 4.76
       & 0.08 &\bf 4.89 &4.94
       & 0.29 &4.79 &\bf 5.05\\
    DeepSeek-V3
       & 0.09 &\bf 4.82  & 4.93
       & 1.96 &4.37  &\bf 6.00
       & 1.91 &2.76  &5.15\\
    \bottomrule
  \end{tabularx}
  \caption{
      \textbf{Ablation on simulation-feedback content.}
      We compare full feedback, feedback without the environment querier and score-only feedback during refinement on the Throwing task. Full feedback includes scalar simulation scores, basic simulation feedback and targeted feedback requested by the environment querier. The no-env.-querier setting removes targeted feedback, while score-only feedback provides only scalar simulation scores. Each cell reports the standard deviation, mean and maximum refiner simulation score.
    }
  \label{tab:rl_env_feedback}
\end{table*}

\subsection{Edit-history ablation}

When edit history is enabled, the refiner receives a compact history of previous modifications and their outcomes. This history helps prevent repeated invalid edits and helps the refiner avoid oscillating between similar machine states. The edit-history ablation is reported in Table~\ref{tab:edit_history}.

\begin{table*}[h!]
  \scriptsize
  \centering
  \setlength{\tabcolsep}{3pt} 
  \renewcommand{\arraystretch}{1.1}
  \newcommand{\cgr}[1]{\textcolor[rgb]{.329, .51, .208}{\textbf{#1}}}
  \newcommand{\cre}[1]{\textcolor[rgb]{1, 0, 0}{\textbf{#1}}}    

  \begin{tabularx}{\textwidth}{l*{4}{>{\centering\arraybackslash}X}}
    \toprule
    \multirow{2}{*}{Models} & \multicolumn{2}{c}{Refiner Avg Retry $\downarrow$} & \multicolumn{2}{c}{Refiner validity rate $\uparrow$} \\
    \cmidrule(lr){2-3} \cmidrule(lr){4-5}
    & Baseline & w/o Modify History & Baseline & w/o Modify History \\
    \midrule
    Gemini 2.5 Pro        & 1.42  & \cgr{1.33}         & 100\% & 100\% \\
    o3                    & 1.94  & \cre{2.37}                 & 97.87\% & \cre{88.89\%} \\
    Qwen3-Coder-480B-A35B & 2.50    & \cre{2.75}          & 82.65\% & \cgr{91.94\%} \\
    Doubao Seed 1.6       & 2.74  & \cre{2.93}            & 85.18\% & 85.18\% \\
    Claude Opus 4       & 3.24  &  \cre{3.69}                 & 94.12\% & \cre{53.85\%}              \\
    DeepSeek-V3           & 1.54  &  \cre{1.68}                 & 100\% &  \cre{98.31\%}            \\
    \bottomrule
  \end{tabularx}
  \caption{
    Ablation on edit history as input to the refiner. We compare the baseline setting, where the refiner receives a compact summary of previous modifications and outcomes, with a setting that removes this history. We report the average number of refiner retries, where lower is better, and the refiner validity rate, where higher is better.
  }
  \label{tab:edit_history}
\end{table*}

\subsection{Stage-wise ablations of agentic workflows}

We analyze how performance changes as different stages are added to the agentic workflows. Table~\ref{tab:test_time_scaling_benchmark} reports stage-wise results for the hierarchical workflow, in which machines are first constructed through meta-designer-guided module decomposition, then refined after feedback-free inspection, and finally refined with selected environment feedback. Table~\ref{tab:test_time_scaling_designer_benchmark} reports the corresponding stage-wise results for the iterative editing workflow, which starts from direct designer generation without meta-designer-based module decomposition.

These comparisons separate the effect of hierarchical decomposition from the effect of iterative editing and simulation-guided refinement. Hierarchical design can stabilize search when the meta-designer produces reliable module plans, but flawed high-level decompositions can propagate errors to downstream construction and refinement. Iterative editing instead starts from a complete designer-generated machine and improves it through critique and feedback, providing a baseline for evaluating whether hierarchical decomposition itself is beneficial.

\begin{table*}[t]
  \scriptsize
  \centering
  \setlength{\tabcolsep}{4pt}
  \renewcommand{\arraystretch}{1.2}
  \begin{tabularx}{\textwidth}{l*{9}{>{\centering\arraybackslash}X}}
    \toprule
    \multirow{2}{*}{Models} & \multicolumn{3}{c}{Designer} & \multicolumn{3}{c}{Blind Refinement} & \multicolumn{3}{c}{Modification w/ Env. Feedback} 
    \\
    \cmidrule(lr){2-4} \cmidrule(lr){5-7} \cmidrule(lr){8-10}
     & Valid & Mean & Max & Valid & Mean & Max & Valid & Mean & Max \\
    \midrule
    \multicolumn{10}{l}{\textit{Throwing}} \\
    \midrule
    Gemini 2.5 Pro      &3&6.13&9.0 &3&8.10&12.09 &3&11.08&21.95 \\
    Claude Opus 4       &2&4.76&4.91 &0&-&-&0&-&-    \\
    o3                 &8&2.87&5.22 &8&2.98&9.17&8&9.14&14.01      \\
    Qwen3-Coder-480B-A35B   &4&3.5&9.24 &4&6.39&10.78&4&10.2&12.02 \\
    Doubao Seed 1.6        &6&4.24&8.2 &6&4.61&8.75&6&6.43&9.10   \\
    DeepSeek-V3            &6&4.67&4.86&5&4.33&4.78&5&4.91&5.24    \\
    Kimi K2              &3&6.85&9.05&2&8.31&8.97&2&11.28&11.39  \\
    GPT-5  &5&1.50&1.88     &5&5.86&12.77&5&7.53&9.48            \\
    Llama 4 Scout 17B 16E  &7&3.63&5.64 &2&5.88&6.95&2&5.12&5.94 \\
    \bottomrule
  \end{tabularx}
  \caption{
      \textbf{Performance of machines generated after different stages of the iterative editing workflow.}
      We report results on the Throwing task for the workflow without meta-designer-based module decomposition. The stages are direct designer generation, blind refinement after feedback-free inspection, and modification with selected environment feedback. Each cell reports the number of valid machines, the mean simulation score and the maximum simulation score. Mean scores are computed only over valid machines.
    }
  \label{tab:test_time_scaling_designer_benchmark}
\end{table*}

\subsection{Search-strategy ablation}

We compare MCTS-style refinement with best-of-\(N\) and random-search baselines. The algorithms and implementation details are described in Sec.~\ref{sec:agentic_workflow_design}. The quantitative comparison is reported in Table~\ref{tab:abl:env-search}, and the effect of search depth is reported in Table~\ref{tab:mcts_ablation}. Fig.~\ref{fig:search_curve} shows the score trend as a function of node expansions.

\subsection{Representation and spatial-information ablations}

\begin{table*}[t]
\scriptsize
  \centering
  \setlength{\tabcolsep}{3pt}
  \renewcommand{\arraystretch}{1.1}
  \newcommand{\cgr}[1]{\textcolor[rgb]{.329, .51, .208}{\textbf{#1}}}
  \newcommand{\cre}[1]{\textcolor[rgb]{1, 0, 0}{\textbf{#1}}}

  \begin{tabularx}{\textwidth}{l*{6}{>{\centering\arraybackslash}X}}
    \toprule
    \multicolumn{1}{c}{\multirow{2.4}{*}{Models}}
    & \multicolumn{3}{c}{Baseline (w/ Parsed Spatial Info.)}
    & \multicolumn{3}{c}{w/o Parsed Spatial Info.} \\
    \cmidrule(lr){2-4}\cmidrule(lr){5-7}
    & Valid & Mean & Max 
    & Valid & Mean & Max \\
    \midrule
    Gemini 2.5 Pro
       & 5 & 8.18 & 11.07
       & 5 & 7.13 & 11.36 \\
    o3
       & 3 & 0 & 0
       & 3 & 0 & 0 \\
    Qwen3-Coder-480B-A35B
       & 5 & 1.61 & 5.0
       & 0 & - & - \\
    Claude Opus 4
       & 2 & 5.38 & 5.8
       & 2 & 0.18 & 0.26 \\
    \bottomrule
  \end{tabularx}
  \caption{
      \textbf{Ablation on parsed spatial information.}
      We compare refinement with and without parsed spatial information from the machine representation. Parsed spatial information summarizes parent--child relations, attachable faces and orientation information. We report the number of valid machines, mean refiner simulation score and maximum refiner simulation score. Mean scores are computed only over valid machines.
    }
  \label{tab:abl:machine-3d-info}
\end{table*}

We also conduct a detailed ablation study in \autoref{tab:abl:machine-3d-info}. The results show clear improvements, especially for Gemini 2.5 Pro and Qwen3-Coder-480B-A35B.

\section{Additional failure analysis and qualitative examples}
\label{sec:failure_examples}

This section provides additional qualitative examples supporting the failure analysis in the main text. These examples illustrate that LLM-generated machines can fail at both the planning level and the instantiation level: a model may propose an unsuitable mechanism, omit an essential component, or produce a plausible high-level design that is incorrectly grounded into spatial attachments, orientations or local geometry.

\subsection{Failure patterns}

\begin{figure*}[h!]
  \centering
  \includegraphics[width=\linewidth]{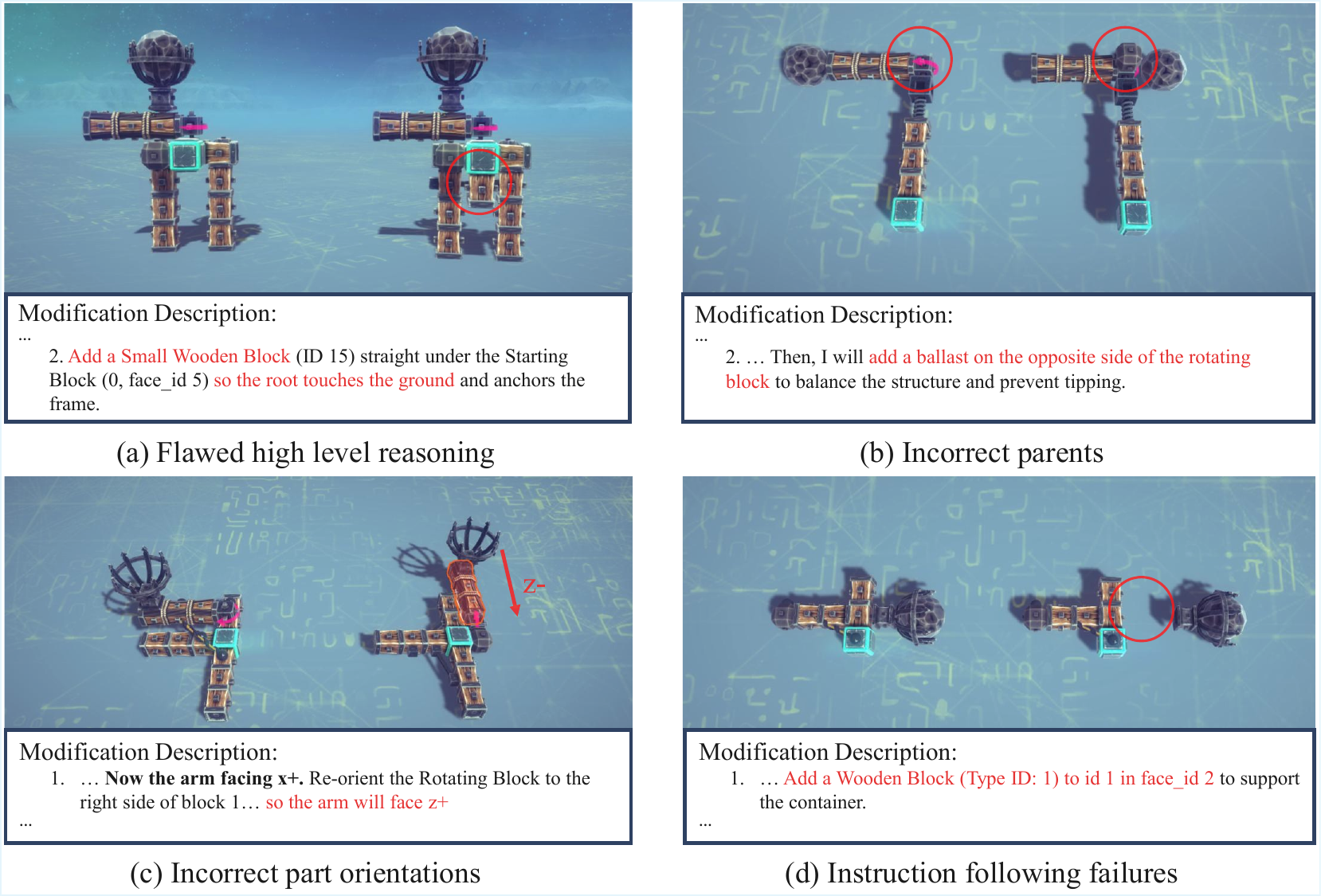}
  \caption{
  \textbf{Recurring failure patterns in LLM-generated machines.}
  In each example, the original machine is shown on the left and the modified or generated machine is shown on the right. The examples are sampled from Qwen3-Coder-480B-A35B-Instruct generations.
  }
  \label{fig:supp-failure-patterns}
\end{figure*}

Generated machines often fail in systematic ways. We observe several recurring categories of errors, including flawed high-level physical reasoning, missing functional components, incorrect attachment relations, incorrect part orientations and failures to follow the intended blueprint. Examples are shown in Fig.~\ref{fig:supp-failure-patterns}. In each case, the generated machine may appear locally plausible, but the realized assembly fails to produce the intended physical behavior.

\subsection{Sensitivity to local spatial modifications}

\begin{figure*}[h!]
  \centering
  \includegraphics[width=\linewidth]{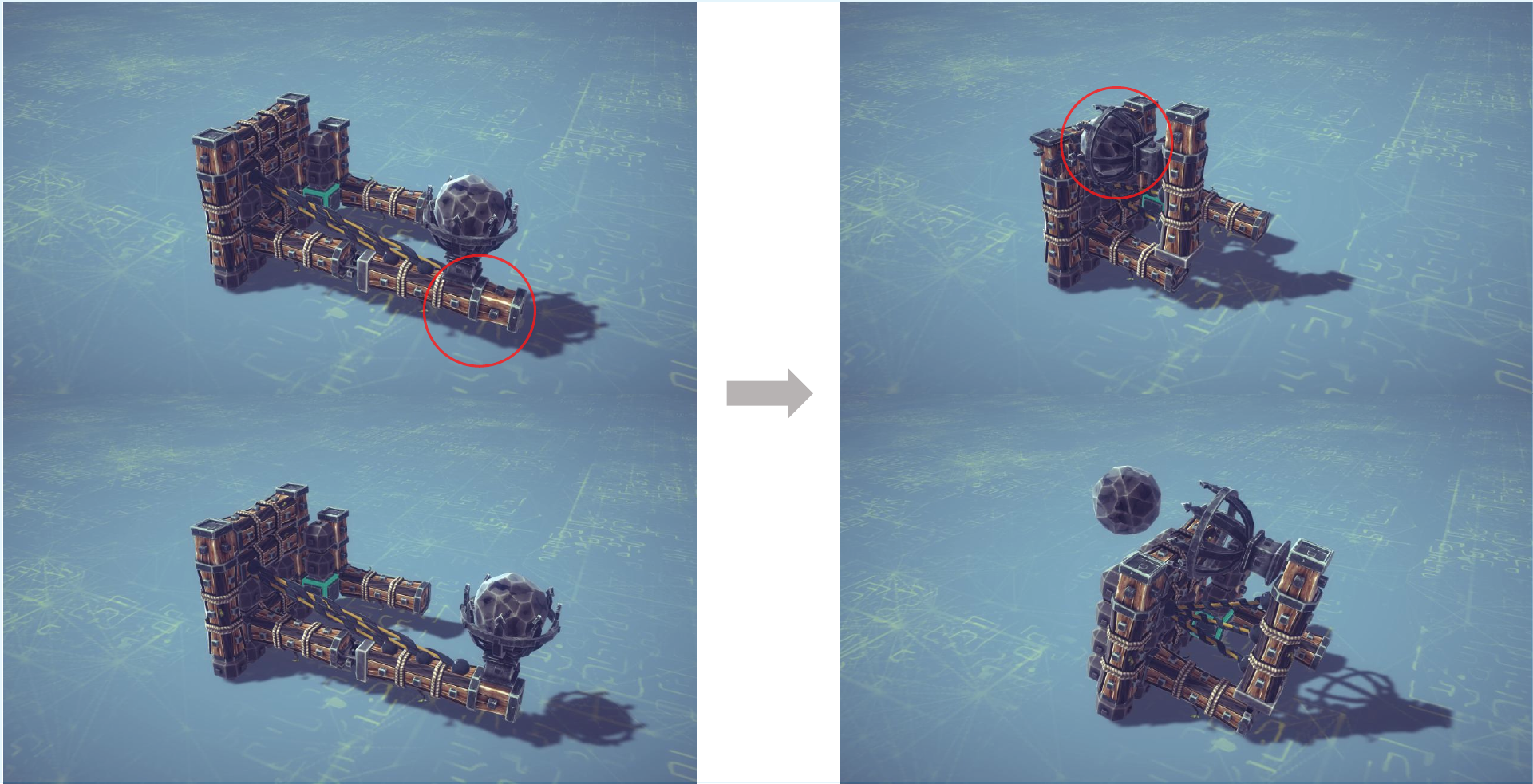}
  \caption{
  \textbf{Sensitivity to local spatial placement.}
  The upper row shows machines sampled from Gemini 2.5 Pro; the lower row shows manually modified versions. Left, designed machines. Right, simulation results. Although the high-level design is feasible, small differences in component placement determine whether the machine successfully launches the boulder.
  }
  \label{fig:tiny-modify}
\end{figure*}

Compositional machine design is sensitive to local spatial precision. A high-level mechanism can be conceptually feasible but fail because a key component is placed or oriented incorrectly. This is especially common in dynamic mechanisms such as catapults, where small changes in pivot placement, container position or actuator orientation can determine whether a projectile is released successfully.

Fig.~\ref{fig:tiny-modify} shows an example in which the high-level design is plausible, but the original machine fails to throw the boulder because the container is placed incorrectly. A small manual modification produces a functioning design. This illustrates why compositional machine design requires more than semantic knowledge of machine parts: the model must also build the design with sufficient geometric and mechanical precision.

\subsection{Human and LLM qualitative examples}

\begin{figure}[h!]
  \centering
  \includegraphics[width=\linewidth]{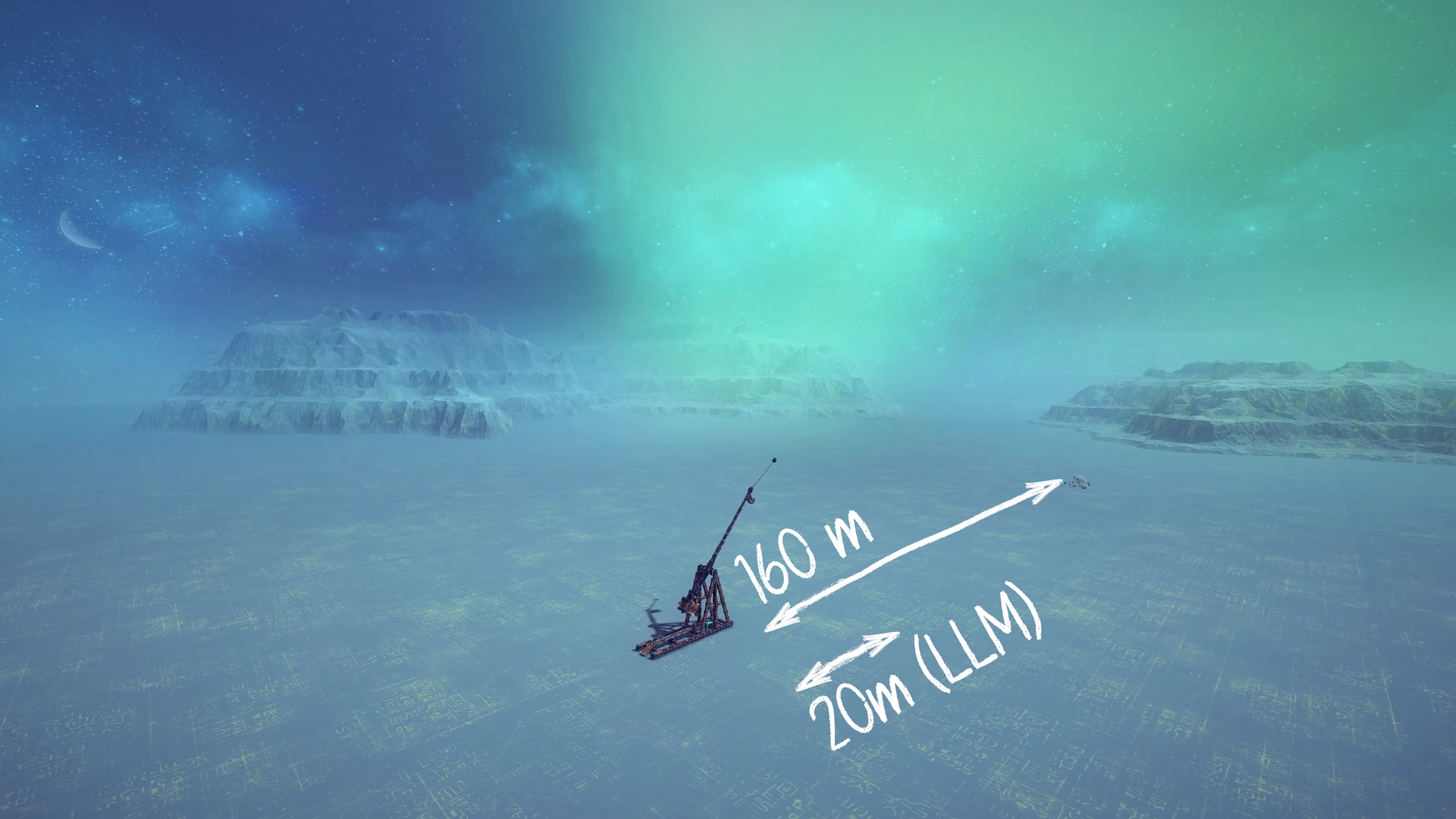}
  \caption{
  \textbf{Qualitative comparison between a human-constructed catapult and an LLM-designed catapult.} The example illustrates differences in spatial coordination, mechanism layout and use of functional parts.
  }
  \label{fig:human-catapult}
\end{figure}

Human-generated machines provide useful qualitative references for understanding the gap between LLM-generated designs and mechanically coherent human-built designs. Fig.~\ref{fig:human-catapult} compares the performance of a human-constructed catapult with the performance of an LLM-designed catapult. The comparison is qualitative rather than a controlled expert benchmark; it illustrates differences in spatial coordination, mechanism layout and use of functional parts.

\subsection{Appearance versus performance}

\begin{figure*}[h!]
  \centering
  \includegraphics[width=\linewidth]{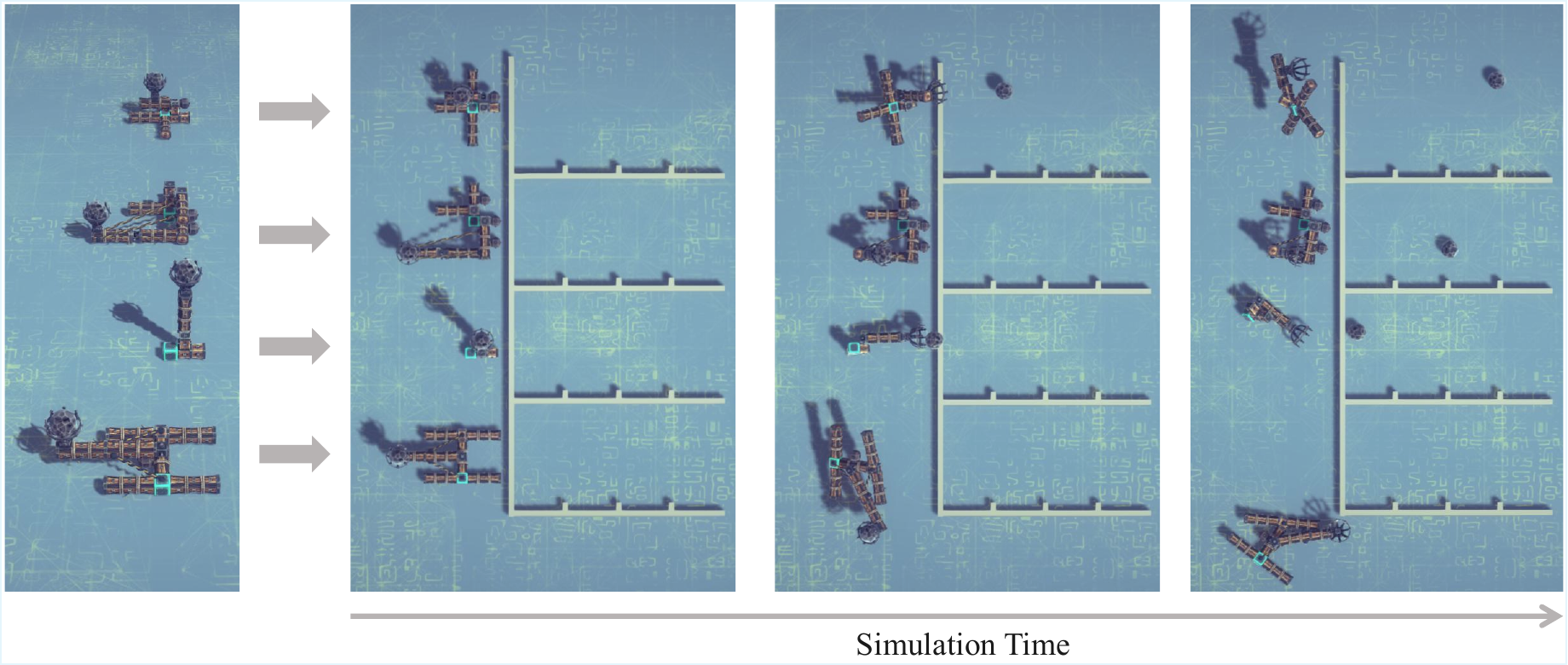}
  \caption{
  \textbf{Boulder-throwing trajectories for machine designs generated by Gemini 2.5 Pro.} From left to right, each row first shows the machine design, followed by a time-lapsed bird's-eye view of the throw.
  }
  \label{fig:throw-traj}
\end{figure*}

Visual plausibility is not always a reliable proxy for physical performance. Some machines appear well aligned with human intuition but fail during simulation because of subtle geometric or mechanical issues. Conversely, some visually unusual machines can achieve high task scores by exploiting the physical dynamics of the environment. Fig.~\ref{fig:throw-traj} illustrates this gap between appearance and simulated behavior.

This observation motivates the use of simulation-derived rewards and validity checks rather than visual plausibility alone. It also suggests that future design objectives may need to balance task performance with additional criteria such as stability, robustness and interpretability if the goal is to produce machines that are both effective and human-legible.

\subsection{Spatial-overlap reasoning diagnostic}
\label{sec:spatial_reasoning}

The benchmark results reveal large differences in machine-construction ability across LLMs. To test whether these differences are related to spatial grounding rather than only familiarity with \textsc{\envname}, we construct a diagnostic spatial-reasoning task. In this task, models are given construction trees for invalid machine designs and asked to identify pairs of blocks that spatially overlap.

Each question contains the block sizes and a construction-tree JSON program. The model must return all overlapping block-id pairs. A correctly identified overlapping pair receives one point, and the final score is computed as the fraction of ground-truth overlapping pairs identified by the model. An example instance is shown below.

\begin{lstlisting}
[System prompt with block sizes]
Q: Give overlap id pairs of this machine in format [(...),(...)]. 
Return an empty list if there are no overlaps.

[
     {"type": 0, "id": 0, "parent": -1, "face_id": -1},
     {"type": 63, "id": 1, "parent": 0, "face_id": 0},
     {"type": 63, "id": 2, "parent": 0, "face_id": 1},
     {"type": 22, "id": 3, "parent": 1, "face_id": 4},
     {"type": 22, "id": 4, "parent": 1, "face_id": 2},
     {"type": 22, "id": 5, "parent": 2, "face_id": 5},
     {"type": 22, "id": 6, "parent": 2, "face_id": 5},
     {"type": 22, "id": 7, "parent": 1, "face_id": 2}
]
A: [(4,7), (5,6)]
\end{lstlisting}

Results are reported in Table~\ref{tab:spatial-reasoning}. Gemini 2.5 Pro achieves the strongest performance among the evaluated models, consistent with its stronger machine-construction performance in the main benchmark. This diagnostic suggests that spatial grounding from construction-tree descriptions is an important bottleneck for compositional machine design.

\begin{table}[h!]
  \centering
  \begin{tabular}{lc}
    \toprule
    Models & Accuracy \\
    \midrule
    Gemini 2.5 Pro        & \textbf{0.416} \\
    OpenAI o3             & 0.143          \\
    Qwen3-Coder-480B-A35B & 0.208          \\
    Doubao Seed 1.6       & 0.208          \\
    Claude Opus 4       & 0.039          \\
    DeepSeek-V3           & 0.065          \\
    Kimi K2               & 0.026          \\
    Llama 4 Scout 17B 16E & 0.026          \\
    \bottomrule
  \end{tabular}
  \caption{
  \textbf{Quantitative results of LLMs on the spatial-overlap QA task.} Gemini 2.5 Pro achieves the highest performance.
  }
  \label{tab:spatial-reasoning}
\end{table}

\subsection{Conditioning on Gemini-generated design rationales}
\label{sec:gemini_cot_conditioning}

To separate high-level design planning from low-level machine instantiation, we evaluate whether other LLMs benefit from design rationales generated by Gemini 2.5 Pro. In this experiment, models are asked to generate machine details conditioned on Gemini-generated chain-of-thought (CoT) rationales rather than on rationales produced by themselves.

\begin{figure}[h!]
  \centering
  \includegraphics[width=\linewidth]{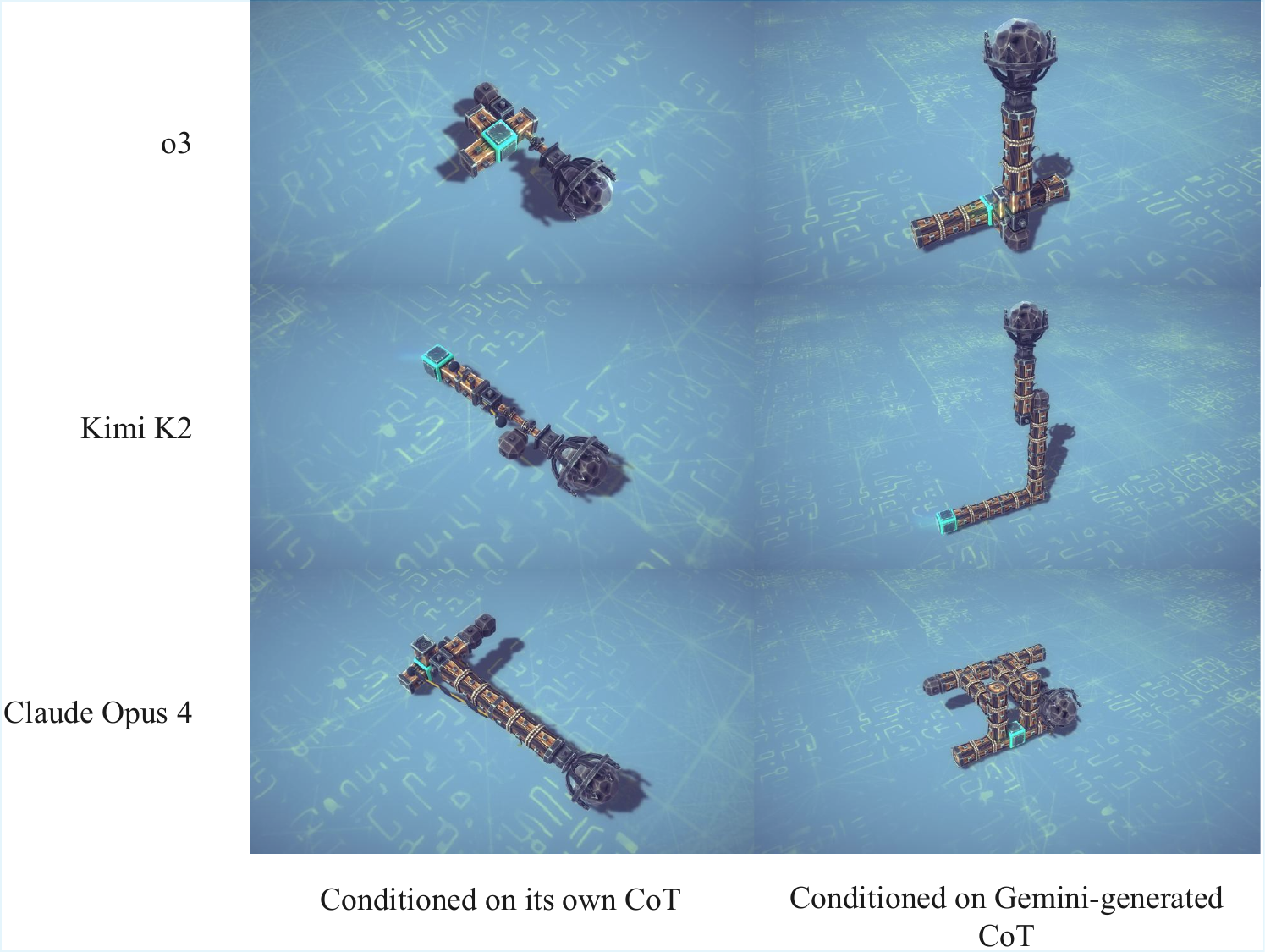}
  \caption{
  \textbf{Machine generations conditioned on Gemini 2.5 Pro design rationales.} Providing stronger high-level rationales makes outputs more visually similar to catapult-like mechanisms, but does not eliminate low-level instantiation errors.
  }
  \label{fig:LLM-feed-gemini-cot}
\end{figure}

Fig.~\ref{fig:LLM-feed-gemini-cot} shows that conditioning on Gemini-generated rationales generally makes generated machines more visually similar to catapult-like mechanisms. This suggests that high-level semantic and physical planning is useful for the task. However, the resulting machines do not fully close the performance gap, indicating that translating a plausible design rationale into precise spatial attachments, orientations and actuation settings remains a separate bottleneck.

\section{Additional generated machine examples}
\label{sec:generated_samples}

This section provides qualitative examples of machines generated by reinforcement-learning fine-tuned models and by the agentic workflow. These examples are intended to complement the quantitative results by illustrating the range of machine structures discovered under different generation settings.

\subsection{Samples from RL-fine-tuned models}

\begin{figure*}[h!]
  \centering
  \includegraphics[width=\linewidth]{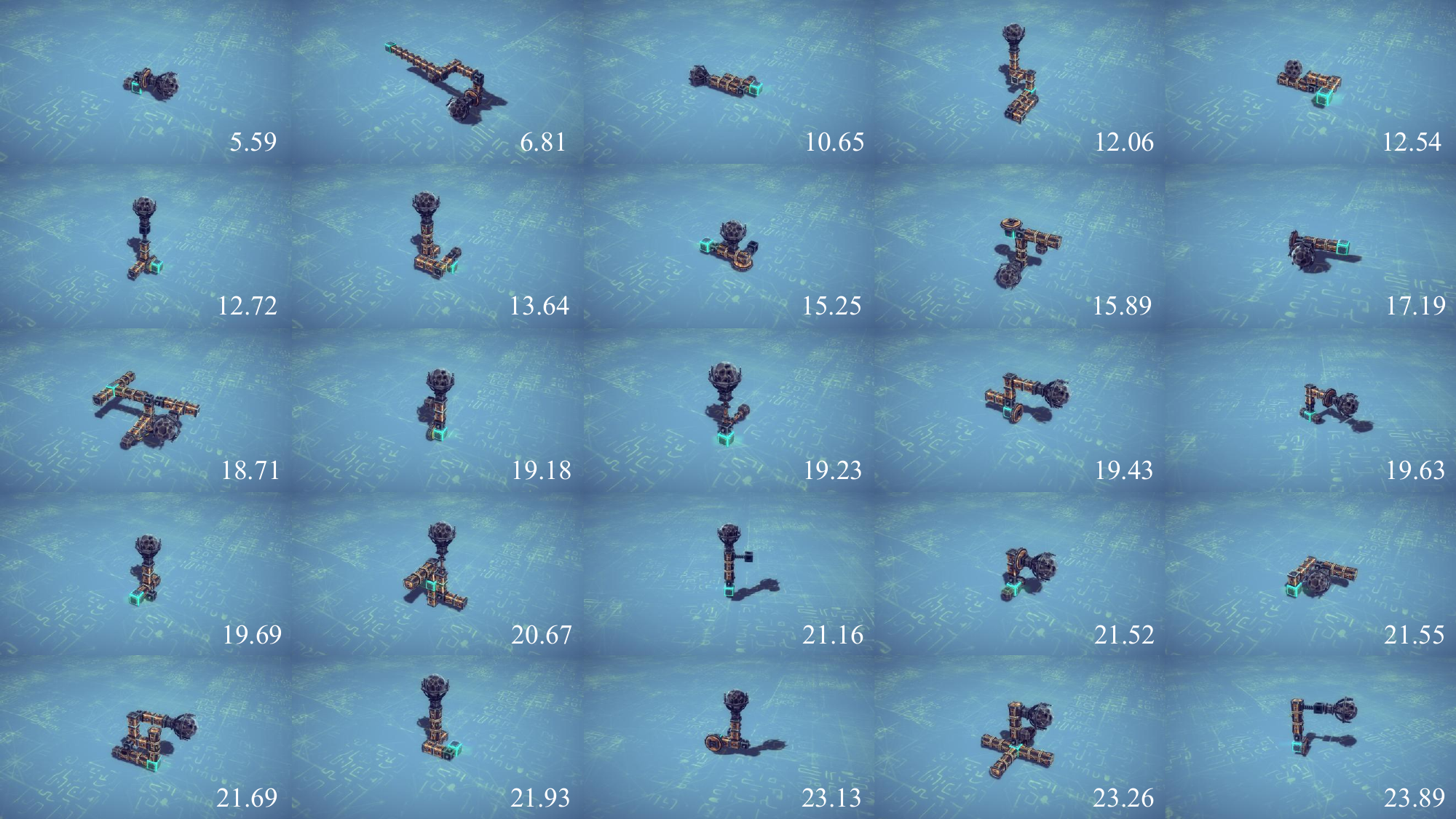}
  \caption{
  \textbf{Representative high-performing samples from the RL-fine-tuned Qwen2.5-14B-Instruct model on the Throwing task.} Throwing distances are shown in the bottom-right corner of each image.
  }
  \label{fig:RL_best_sample}
\end{figure*}

Fig.~\ref{fig:RL_best_sample} shows representative high-performing Throwing machines sampled from rollouts of the Qwen2.5-14B-Instruct model after cold-start fine-tuning and reinforcement learning. The throwing distance for each machine is shown in the bottom-right corner of the corresponding image.

\subsection{Samples from agentic workflows}

\begin{figure*}[h!]
  \centering
  \includegraphics[width=\linewidth]{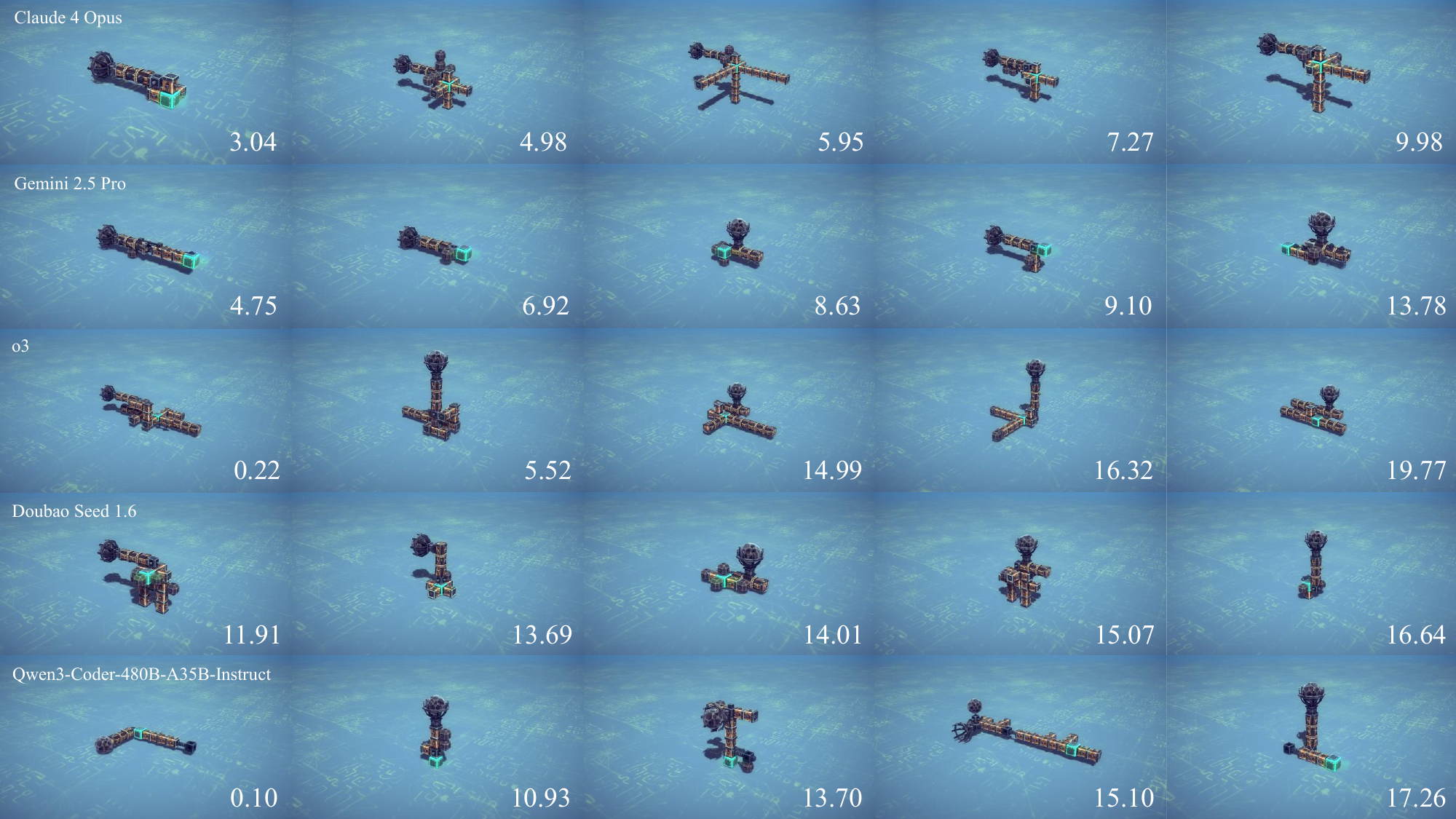}
  \caption{
  \textbf{Gallery of machine samples generated by the agentic workflow.} Rows correspond, from top to bottom, to Claude Opus 4, Gemini 2.5 Pro, OpenAI o3, Doubao Seed 1.6 and Qwen3-Coder-480B-A35B-Instruct. Throwing distances are shown in the bottom-right corner of each image.
  }
  \label{fig:test_time_scaling_gallery}
\end{figure*}

Fig.~\ref{fig:generated_machine_gallery} shows representative machines generated by the agentic workflow across several LLMs, and Fig.~\ref{fig:test_time_scaling_gallery} further provides a detailed illustration with performance comparison on the Throwing task. The examples illustrate both the ability of strong models to generate recognizable mechanical structures and the large variation in task performance across models and samples.

\section{Exploratory diversity analysis}
\label{sec:diversity_analysis}

The main evaluation focuses on validity and task performance. However, diversity is an important property for open-ended compositional design, because many distinct structures can realize the same function. We therefore include an exploratory diversity analysis to characterize the range of valid catapult designs generated by LLMs, by human creators and during reinforcement learning. This analysis is not used as a primary benchmark metric in the main paper, because current models still struggle to reliably generate high-performing single outputs.

\subsection{Data sources}

We analyze valid catapult designs from three sources. First, we sample 20 valid catapults from each of three high-performing LLMs: Gemini 2.5 Pro, OpenAI o3 and Kimi K2. Second, we sample 20 human-authored catapults from the Steam Workshop\footnote{\url{https://steamcommunity.com/workshop/}}. Third, during reinforcement learning, we sample 20 machines every 40 training steps from the RL-from-base-Qwen training run. These samples allow us to compare diversity across human designs, LLM-generated designs and RL-generated designs over training.

\subsection{Shape diversity}

We measure shape diversity using Chamfer Distance (CD)~\citep{yuan2018pcn}, Hausdorff Distance (HD)~\citep{huttenlocher1992multi} and Intersection-over-Union (IoU)~\citep{qi2017pointnet}. Each machine is converted into a point cloud or voxelized occupancy representation before computing pairwise distances.

Chamfer Distance measures the bidirectional average nearest-neighbor distance between two point clouds and provides a smooth global measure of geometric dissimilarity. Hausdorff Distance measures the maximum nearest-neighbor deviation and captures worst-case geometric differences. IoU measures the overlap between voxelized machine volumes, with lower IoU indicating greater geometric diversity. For each source, we compute pairwise distances among machines generated by that source and report the average value for each metric.

\subsection{Topology diversity}

We also measure topology diversity from machine construction trees. Human-authored designs are first parsed into tree structures when possible. We then compare trees using Graph Embedding Similarity and Tree Edit Distance (TED)~\citep{zhang1989simple}.

For Graph Embedding Similarity, each construction tree is treated as an unweighted graph and embedded using Node2Vec~\citep{grover2016node2vec}. We compute cosine similarity between graph embeddings and average over all pairs within each source. Lower similarity indicates greater topology diversity. TED measures the minimum-cost sequence of node insertions, deletions and substitutions required to transform one tree into another, providing a complementary measure of structural dissimilarity that is sensitive to node-label variations.

\subsection{Diversity during reinforcement learning}

\begin{figure}[h!]
  \centering
  \includegraphics[width=\linewidth]{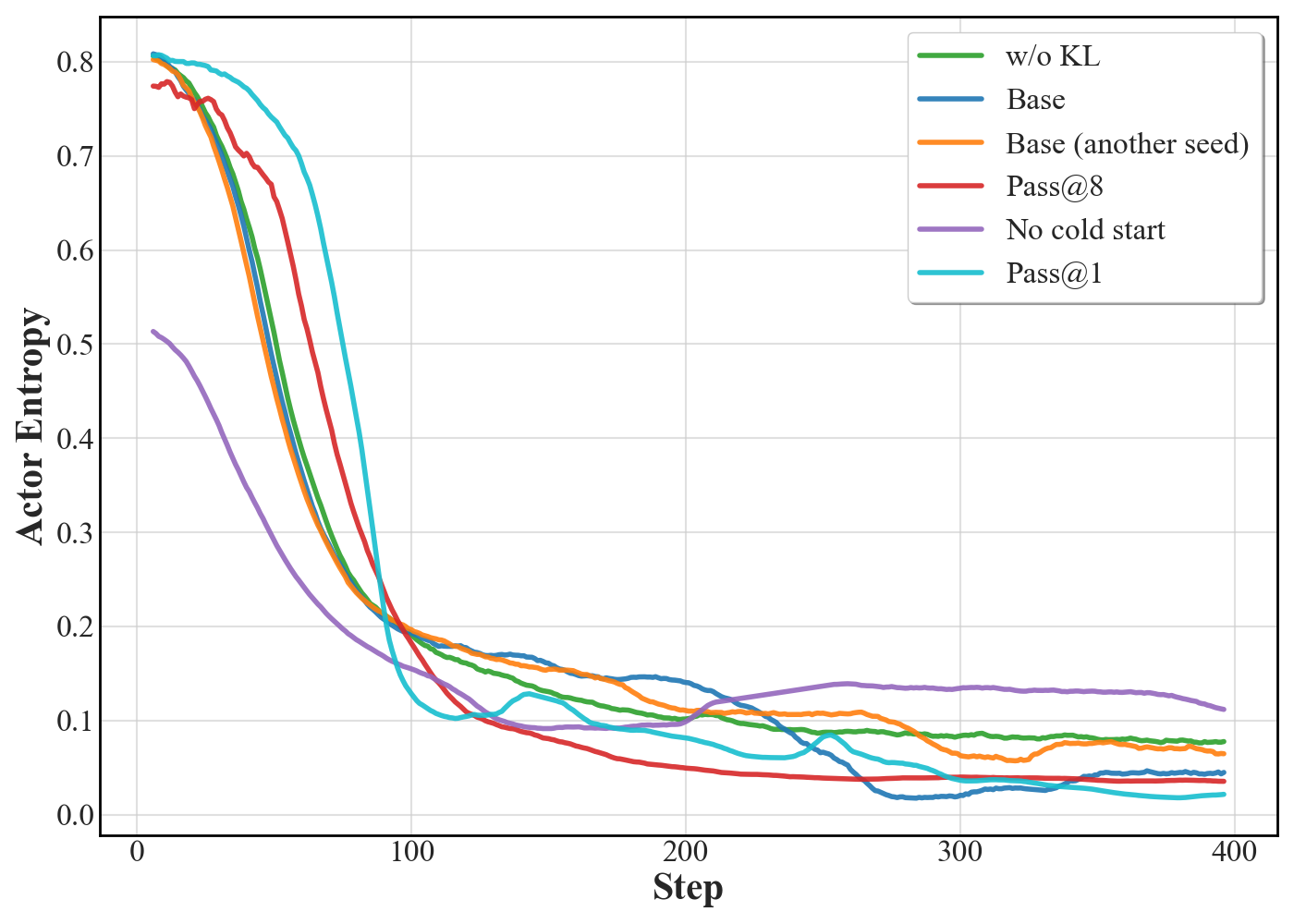}
  \caption{
  \textbf{Entropy during reinforcement learning on the Throwing task.} Entropy decreases as RL training progresses, indicating reduced output diversity.
  }
  \label{fig:RL-catapult-entropy}
\end{figure}

The diversity results are reported in Table~\ref{tab:diversity}. LLM-generated valid machines can exhibit diversity comparable to human-authored machines under some metrics. However, during reinforcement learning, both entropy and structural diversity tend to decrease over training. The entropy trend is shown in Fig.~\ref{fig:RL-catapult-entropy}. 
This suggests that reward optimization can improve high-performing-sample discovery while narrowing the distribution of generated mechanisms.

This exploratory analysis supports the broader observation that diversity-preserving optimization may be important for future machine-design agents. In this work, we focus primarily on validity and task performance, leaving systematic diversity optimization to future work.

\begin{table*}[t!]
  \scriptsize
  \centering
  \setlength{\abovecaptionskip}{5pt}
  \setlength{\belowcaptionskip}{-1pt}
  \setlength{\tabcolsep}{8pt}
  \renewcommand{\arraystretch}{1.2}
  \newcommand{\cgr}[1]{\textcolor[rgb]{.329,.51,.208}{\textbf{#1}}}
  \newcommand{\cre}[1]{\textcolor[rgb]{1, 0, 0}{\textbf{#1}}}
  \renewcommand{\ul}{\underline}
  \renewcommand{\pm}{\mathbin{\text{\pm}}}
  
    \begin{tabular}{lccccc}
      \toprule
      \textbf{Model Name} & \makecell{Chamfer Dist\\($\uparrow$)} & \makecell{Hausdorff Dist\\($\uparrow$)} & \makecell{IoU\\($\downarrow$)} & \makecell{Graph Emb\\Similarity($\downarrow$)} & \makecell{Tree Edit\\dist($\uparrow$)} \\
      \midrule
      Gemini2.5Pro & {\ul 0.287} & {\ul 0.604} & 0.247 & 0.520 & 9.534 \\
      o3           & \textbf{0.311} & \textbf{0.612} & {\ul 0.235} & 0.542 & 9.796 \\
      kimik2       & 0.230 & 0.480 & 0.350 & \textbf{0.392} & {\ul 13.513} \\
      Human        & 0.275 & 0.538 & \textbf{0.206} & {\ul 0.505} & \textbf{15.712} \\
      \bottomrule
      \noalign{\smallskip}
      \textbf{RL Steps} & \makecell{Chamfer Dist\\($\uparrow$)} & \makecell{Hausdorff Dist\\($\uparrow$)} & \makecell{IoU\\($\downarrow$)} & \makecell{Graph Emb\\Similarity($\downarrow$)} & \makecell{Tree Edit\\dist($\uparrow$)} \\
      \midrule
      40   & {\ul 0.370} & {\ul 0.646} & 0.264 & 0.660 & {\ul 10.876} \\
      80   & 0.320 & \textbf{0.661} & 0.292 & 0.725 & 6.142 \\
      120  & 0.360 & 0.577 & 0.307 & {\ul 0.646} & 4.858 \\
      160  & 0.196 & 0.484 & 0.387 & 0.722 & 4.130 \\
      200  & 0.326 & 0.586 & 0.322 & 0.726 & 4.923 \\
      240  & \textbf{0.377} & 0.641 & {\ul 0.214} & 0.672 & 5.701 \\
      280  & 0.110 & 0.401 & 0.621 & 0.760 & 2.884 \\
      320  & 0.095 & 0.338 & 0.742 & 0.767 & 1.630 \\
      360  & 0.276 & 0.525 & 0.390 & 0.729 & 2.381 \\
      400  & 0.226 & 0.460 & 0.473 & 0.737 & 2.356 \\
      \midrule
      Human & 0.275 & 0.538 & \textbf{0.206} & \textbf{0.505} & \textbf{15.712} \\
      \bottomrule
    \end{tabular}
    \caption{
    \textbf{Diversity comparison among agentic LLM systems, RL-trained Qwen2.5-14B-Instruct models and human-authored catapults.} Shape diversity is measured by Chamfer Distance, Hausdorff Distance and IoU; topology diversity is measured by Graph Embedding Similarity and Tree Edit Distance. Bold entries indicate the best value in each column, and underlined entries indicate the second best.
    }
    \label{tab:diversity}
\end{table*}

\section{Cold-start data and reinforcement learning}
\label{sec:rl_details}

\begin{figure*}[t]
  \centering
  \includegraphics[width=\linewidth]{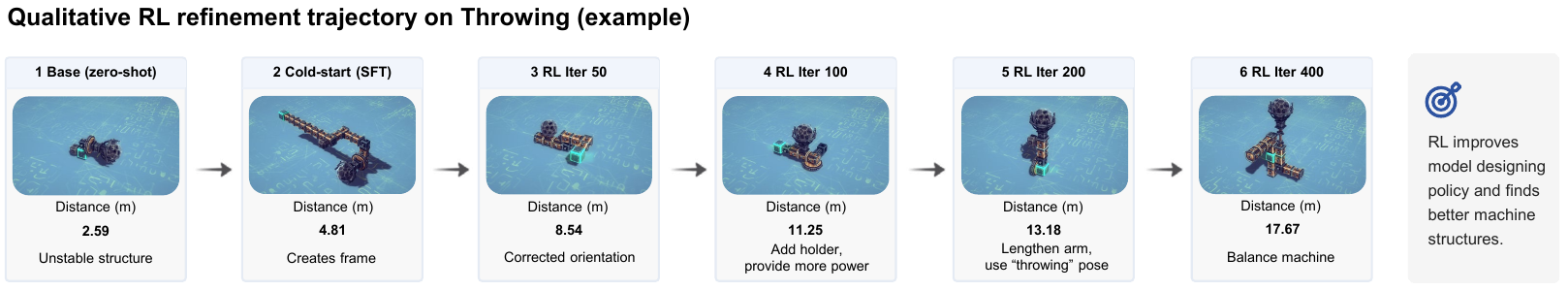}
    \caption{
        \textbf{Qualitative reinforcement learning refinement trajectory on a Throwing task.} During training, the model corrects basic orientation errors and discovers more effective machine structures.
    }
  \label{fig:nmi-rlvr-d}
\end{figure*}

This section provides details of the cold-start dataset, supervised fine-tuning and reinforcement-learning experiments used in the main paper. We use \textit{cold-start} to refer to a supervised fine-tuning stage that gives the base model an initial ability to produce syntactically valid and mechanically plausible machine programs before reinforcement learning begins. This differs from reward optimization: the cold-start model is trained on filtered machine--CoT examples, whereas reinforcement learning updates the model using simulation-derived rewards from \textsc{\envname}. The goal of cold-start fine-tuning is therefore not to solve the benchmark tasks directly, but to initialize the policy in a region of the output space where valid machine construction is sufficiently likely for RL to obtain informative rewards.

\subsection{Cold-start dataset curation}
\label{sec:cold_start_data}

We construct a cold-start dataset using Gemini 2.5 Pro, which produced the strongest machine designs and the most useful design rationales in our benchmark. We use a single-agent generation setting to sample machine programs together with their chain-of-thought (CoT) rationales.

The prompt set contains 100 design objectives. Among them, 75 are captions or descriptions of machines created by \textit{Besiege} players from online communities, and 25 are authored by us. For the authored prompts, we first write simple design objectives that are realizable in \textsc{\envname} and can be evaluated by simple physical rewards, and then add environment constraints or machine-specific requirements.

\begin{figure*}[h!]
  \centering
  \includegraphics[width=\linewidth]{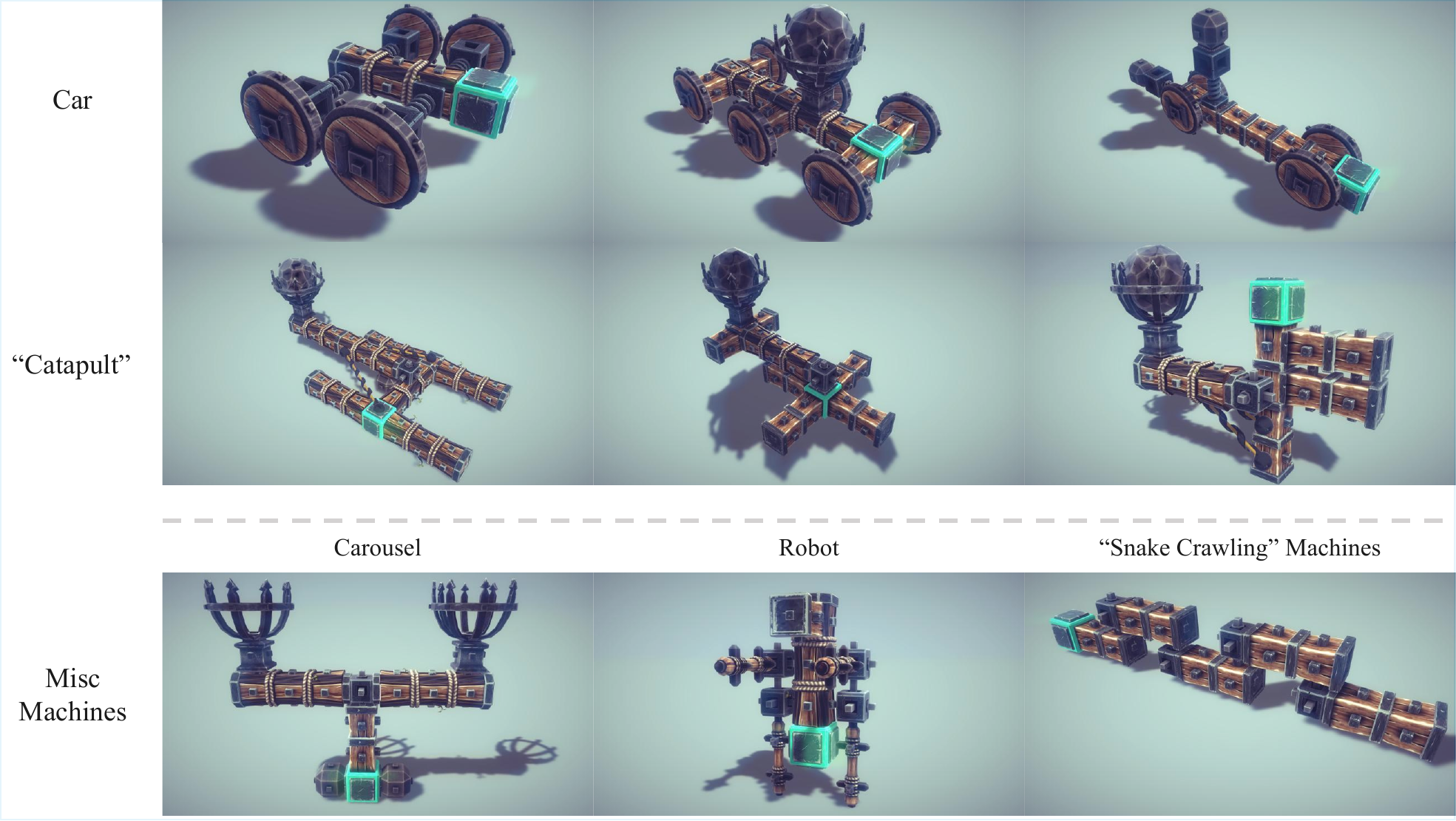}
  \caption{
  \textbf{Examples of Gemini-synthesized machines used in the cold-start dataset.}
  }
  \label{fig:synthesized_dataset}
\end{figure*}

Using this prompt set, we sample 250 candidate machines per prompt. We filter out generations that cannot be parsed, cannot be instantiated in \textsc{\envname}, or do not correspond to a concrete physics-driven functionality. For example, purely decorative structures such as statues are removed. After filtering, we obtain 9,982 valid machine--CoT pairs. We manually compared all objectives against the complete benchmark specification, including prompts, scenes, rewards, and task-specific configurations, and found no exact duplicates, paraphrased benchmark prompts, or reused benchmark-specific settings; the generated programs and reasoning trajectories are likewise distinct from evaluation samples. The objective list is released with the benchmark. A sample gallery from the curated dataset is shown in Fig.~\ref{fig:synthesized_dataset}.

Examples of player-derived or functionality-based prompts include:
\begin{lstlisting}
1. Build a machine that can provide an exciting spinning amusement ride experience.
2. Build a machine that can mimic the movements of a humanoid figure for entertainment or functional demonstrations.
3. Build a machine that can glide smoothly over snow or ice.
\end{lstlisting}

Examples of authored prompts with explicit environment constraints or target-machine requirements include:
\begin{lstlisting}
- Additional Environment Constraints -
1. On an uneven, bumpy straight road, build a small car that must travel in a straight line to the finish.
2. On a straight road stands a stone wall; build a battering ram that must accelerate straight ahead, smash the wall, and finish with minimal damage to the machine.

- Modified Demands for Target Machines -
1. Build a tall tower that must keep a heavy block (id 36) at 15 m height for 5 s without collapse.
2. On a straight road stands a 10 m high wall; build a siege ladder that must advance, extend its top above the wall, and remain upright throughout.
\end{lstlisting}

\subsection{Supervised fine-tuning}
\label{sec:cold_start_details}

We use Qwen2.5-14B-Instruct as the base model for cold-start fine-tuning. The model is trained on the Gemini-synthesized machine--CoT dataset described above. To reduce GPU memory usage, we use parameter-efficient quantized orthogonal fine-tuning (QOFT)~\citep{qiu2025oftv2,qiu2023controlling,liu2024boft,qiu2025reparameterized}, with OFT block size 64. We use 8-bit training with the 8-bit AdamW optimizer implemented in bitsandbytes~\citep{dettmers20228bit}. The learning rate is \(1\times 10^{-6}\), with a linear warmup schedule over 3\% of the total training steps. Each cold-start SFT run was trained on 8 A100 GPUs for 72 hours.

Because invalid machine programs receive zero reward, cold-start fine-tuning reduces the sparsity of the RL signal by increasing the probability of parseable and mechanically plausible generations. The resulting cold-start model is not expected to solve the benchmark tasks by itself; instead, it provides a better initialization for reinforcement learning.

\subsection{Reward functions for reinforcement learning}

The reinforcement-learning experiments use the reward definitions described in Sec.~\ref{sec:reward_setting}. For both Movement and Throwing, invalid machines receive zero reward. For valid machines, the reward is computed from the task-specific performance score.

For \textit{Movement}, the performance score is the distance traveled by the root block along the target direction. For \textit{Throwing}, \(\mathbbm{1}_{\mathrm{valid}}\) additionally requires the boulder to reach a maximum height greater than 3 m. The performance score is the product of the boulder's maximum height and maximum horizontal distance along the target direction. This reward encourages projectile-launching behavior rather than degenerate solutions such as rolling or carrying the boulder.

\subsection{GRPO training}
\label{sec:rl_exp_details}

We implement reinforcement learning using the \texttt{verl} framework~\cite{sheng2025hybridflow}. We evaluate two initialization conditions: RL from the base Qwen2.5-14B-Instruct model, and RL from the cold-start model described above. During reinforcement learning, we apply LoRA~\cite{hu2022lora} adapters of rank 64 to all linear layers.

We use group relative policy optimization (GRPO) with a learning rate of \(5\times 10^{-6}\) and gradient clipping threshold 0.5. The GRPO advantage estimator uses an advantage clipping ratio of 0.2. We apply a KL penalty with coefficient 0.001 relative to the initial policy for each RL run. We do not add an explicit entropy bonus; therefore, the entropy decrease observed during training occurs despite KL regularization.

For rollouts, we sample with temperature 1.0 and top-\(p=0.95\). The maximum input and output lengths are 3440 and 1168 tokens, respectively. Each model is trained for 400 update steps, which takes approximately 48 hours in our setup.

Table~\ref{tab:rl_general} summarizes the effects of cold-start supervised fine-tuning and reinforcement learning on the representative Throwing and Movement tasks. 
Cold-start fine-tuning provides a useful initialization for reinforcement learning, which further improves best-sample performance.

\begin{table*}[t!]
  \scriptsize
  \centering
  \setlength{\tabcolsep}{2.9pt}
  \renewcommand{\arraystretch}{1.25}
  \newcommand{\cgr}[1]{\textcolor[rgb]{.329, .51, .208}{\textbf{#1}}}
  \newcommand{\cre}[1]{\textcolor[rgb]{1, 0, 0}{\textbf{#1}}}     
  \renewcommand{\pm}{\mathbin{\text{\pm}}}

\begin{tabularx}{\textwidth}{
  l|
  *{3}{>{\centering\arraybackslash}X}| 
  *{3}{>{\centering\arraybackslash}X}
}
\multirow{2}{*}{Models} 
 & \multicolumn{3}{c|}{\textit{Throwing}}
 & \multicolumn{3}{c}{\textit{Movement}} \\
 & Validity Ratio & Mean Score & Max Score
 & Validity Ratio & Mean Score & Max Score \\
\shline
Qwen2.5-14B-Instruct           & 11/50 & 0.06 &  2.41 & 46/50 & 4.97 & 19.10 \\
Qwen2.5-14B-Instruct + Cold-Start     &  9/50 & 0.11 &  5.54 & 40/50 & 4.67 & 20.23 \\
Qwen2.5-14B-Instruct + RL      & \textbf{12}/50 & 0.13 &  5.92 &   41/50 &  3.72  &   24.08  \\
Qwen2.5-14B-Instruct + Cold-Start + RL& 11/50 & \textbf{0.14} &  \textbf{7.14} &   \textbf{42}/50  &  \textbf{5.05}  &   \textbf{45.72}  \\
\end{tabularx}
\caption{
    \textbf{Effect of cold-start supervised fine-tuning and RLVR post-training in \textsc{\envname}.} All variants use Qwen2.5-14B-Instruct as the backbone model. We report the number of valid machines out of 50 generations, together with the mean and maximum simulation scores on the \textit{Throwing} and \textit{Movement} tasks.
}
\label{tab:rl_general}
\end{table*}

\subsection{Pass@\(k\) advantage estimation}

Because the design objective emphasizes discovering high-performing machines, we also evaluate a Pass@\(k\)-style advantage estimator. In this setting, multiple rollouts are generated for the same prompt, and the best-performing sample among the group provides the primary learning signal. This objective better matches the practical use case in which multiple candidate machines can be sampled, simulated and compared before selecting the final design.

Due to the implementation structure of \texttt{verl}, the number of rollouts is set equal to \(k\) for the Pass@\(k\) variant~\cite{tang2025optimizing}. The corresponding maximum-score curves, validity curves and Best@\(N\) evaluations are reported in Sec.~\ref{sec:additional_rl_results}.

\subsection{Additional RL results}
\label{sec:additional_rl_results}
\begin{figure}[tp]
  \centering
  \includegraphics[width=\linewidth]{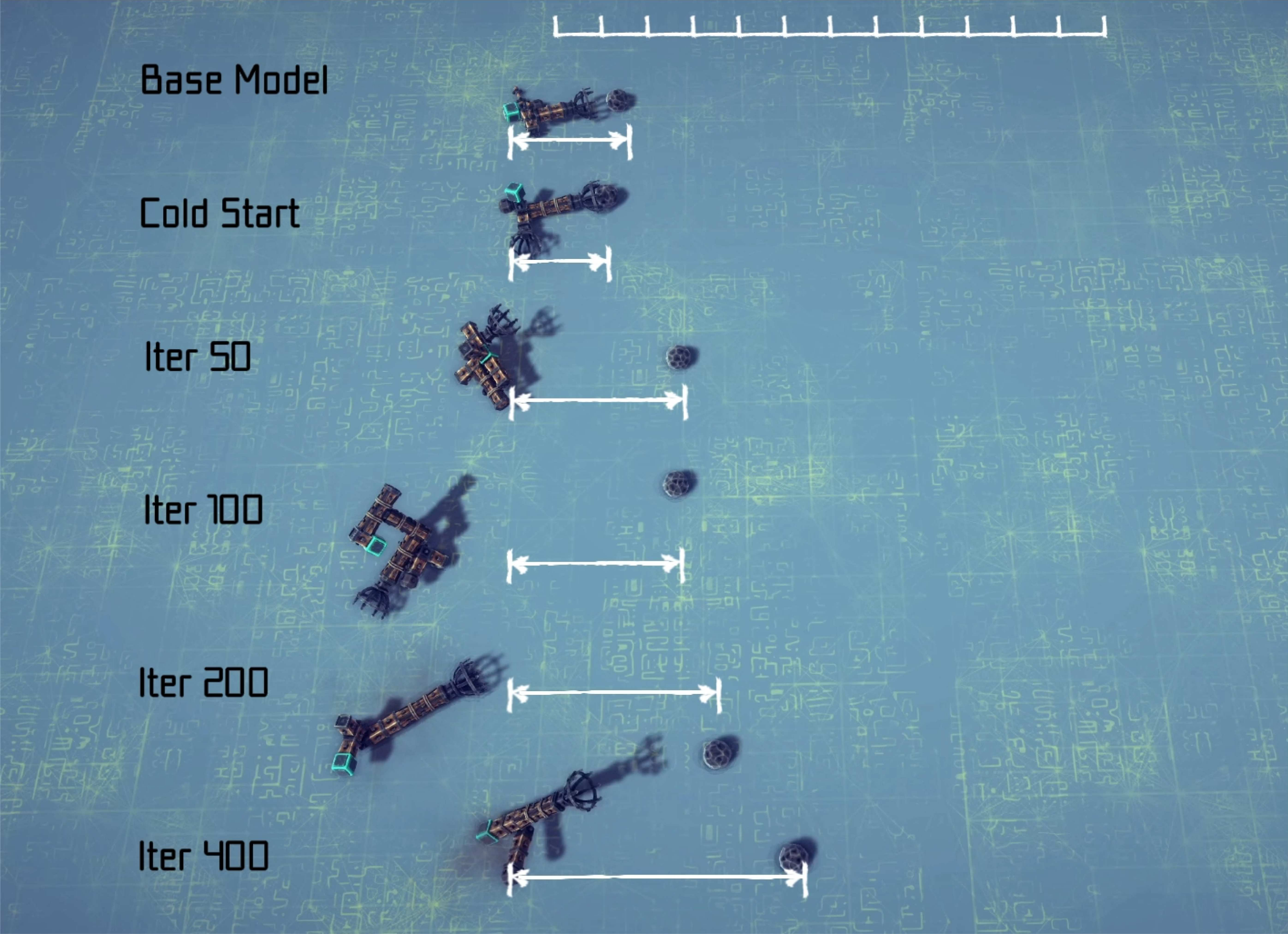}
    \caption{
        \textbf{Qualitative progression of Throwing machines during reinforcement learning.}
        Representative machines sampled from the Qwen2.5-14B-Instruct policy at different training stages: base model, cold-start fine-tuning and RL iterations 50, 100, 200 and 400. The white bars indicate the throwing distance achieved by each machine in simulation.
    }
  \label{fig:nmi-supp-rlresults}
\end{figure}

\begin{figure}[h!]
  \centering
  \includegraphics[width=\linewidth]{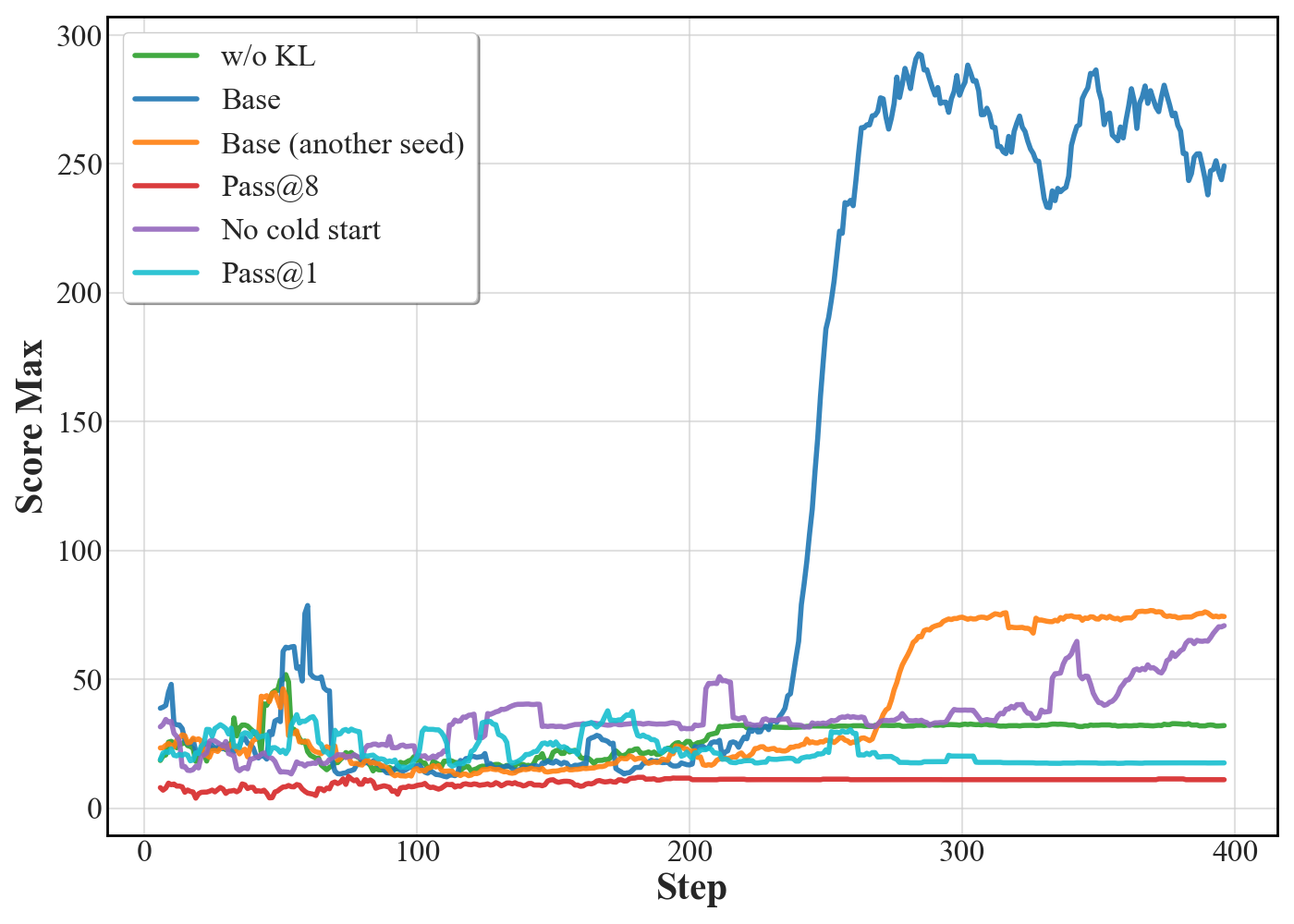}
  \caption{
  \textbf{Maximum machine score on the \textit{Catapult} task across RL training steps.} KL regularization improves the discovery of effective machine structures, while cold-start initialization provides useful prior design knowledge. Pass@64 achieves the strongest maximum score, and Pass@8 improves sample efficiency relative to Pass@1 by discovering high-performing machines with fewer rollouts.
  }
  \label{fig:RL-catapult-max-score}
\end{figure}

\begin{figure*}[h!]
  \centering
  \includegraphics[width=\linewidth]{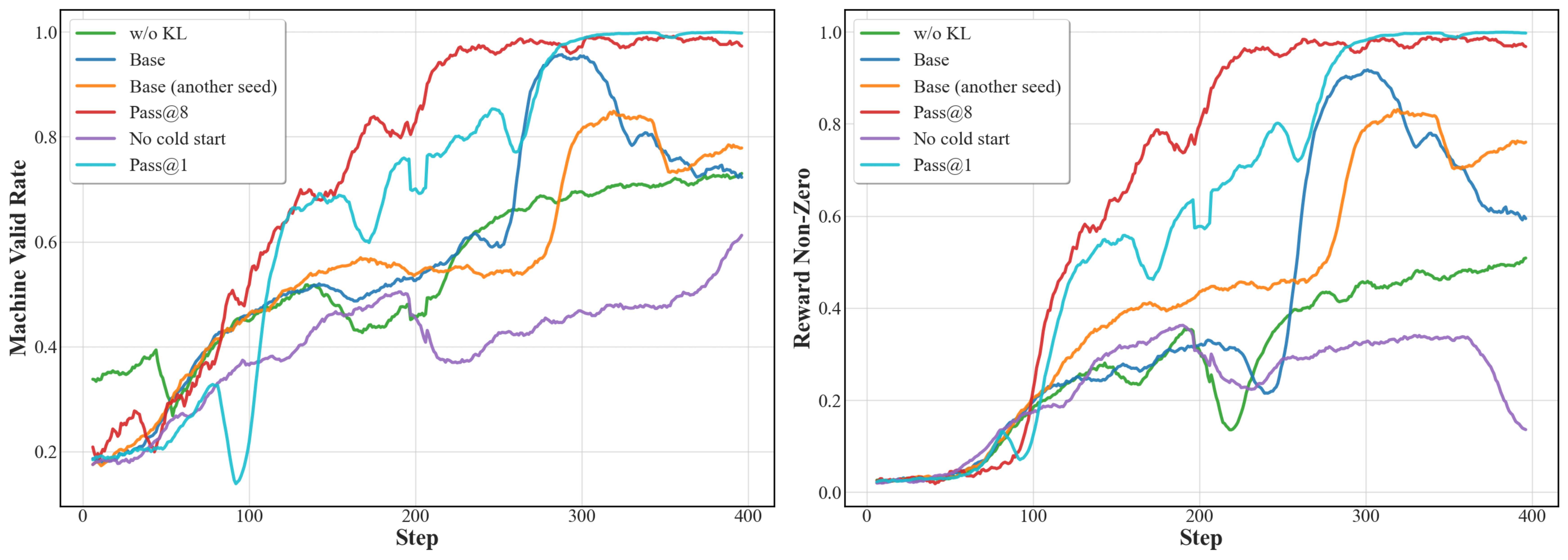}
  \caption{
  \textbf{Machine-validity rate and non-zero-reward rate on the \textit{Catapult} task across RL training steps.} Machine validity measures the fraction of generated machines that can be instantiated and simulated successfully, while non-zero-reward rate measures the fraction of machines that receive positive reward. Validity generally improves during training. Pass@1 and Pass@8 converge early, while removing KL regularization or cold-start initialization leads to more failed rollouts and slower improvement.
  }
  \label{fig:RL-catapult-valid-rate}
\end{figure*}

\begin{figure*}[h!]
  \centering
  \includegraphics[width=\linewidth]{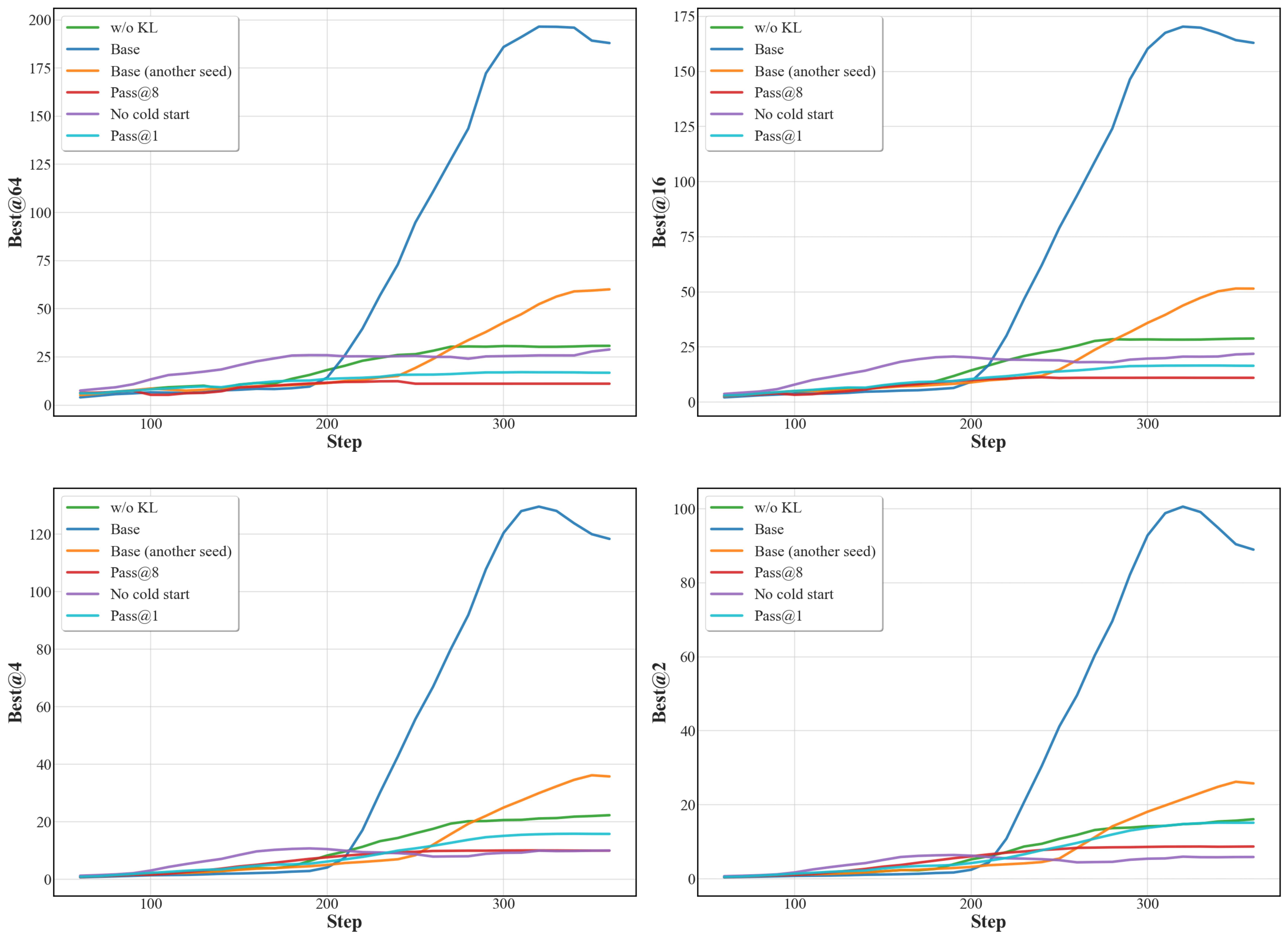}
  \caption{
  \textbf{Best@\(N\) performance on the \textit{Catapult} task.} At each evaluation step, the model generates 64 samples; for each \(N\), we record the best score among \(N\) sampled candidates and average this procedure over 1,000 trials. The base setting achieves the strongest Best@\(N\) performance across both random seeds. Without KL regularization, performance remains competitive among the ablated settings, while Pass@1 and Pass@8 produce fewer high-performing outliers. Without cold-start initialization, the model generates more moderate-quality machines but discovers fewer top-performing designs.
  }
  \label{fig:RL-catapult-best@n}
\end{figure*}

For Throwing, Fig.~\ref{fig:nmi-supp-rlresults} shows a qualitative progression of representative machines sampled at different training stages, from the base model and cold-start model to later RL iterations. Fig.~\ref{fig:RL-catapult-max-score} shows the maximum score achieved during training under different RL settings, and Fig.~\ref{fig:RL-catapult-valid-rate} shows the corresponding machine-validity rate. We report maximum score because, when inference-time scaling is available, the best machine discovered among multiple sampled candidates is often the most relevant outcome for design. Fig.~\ref{fig:RL-catapult-best@n} further reports Best@\(N\) performance for the same settings.

\begin{figure}[h!]
  \centering
  \includegraphics[width=\linewidth]{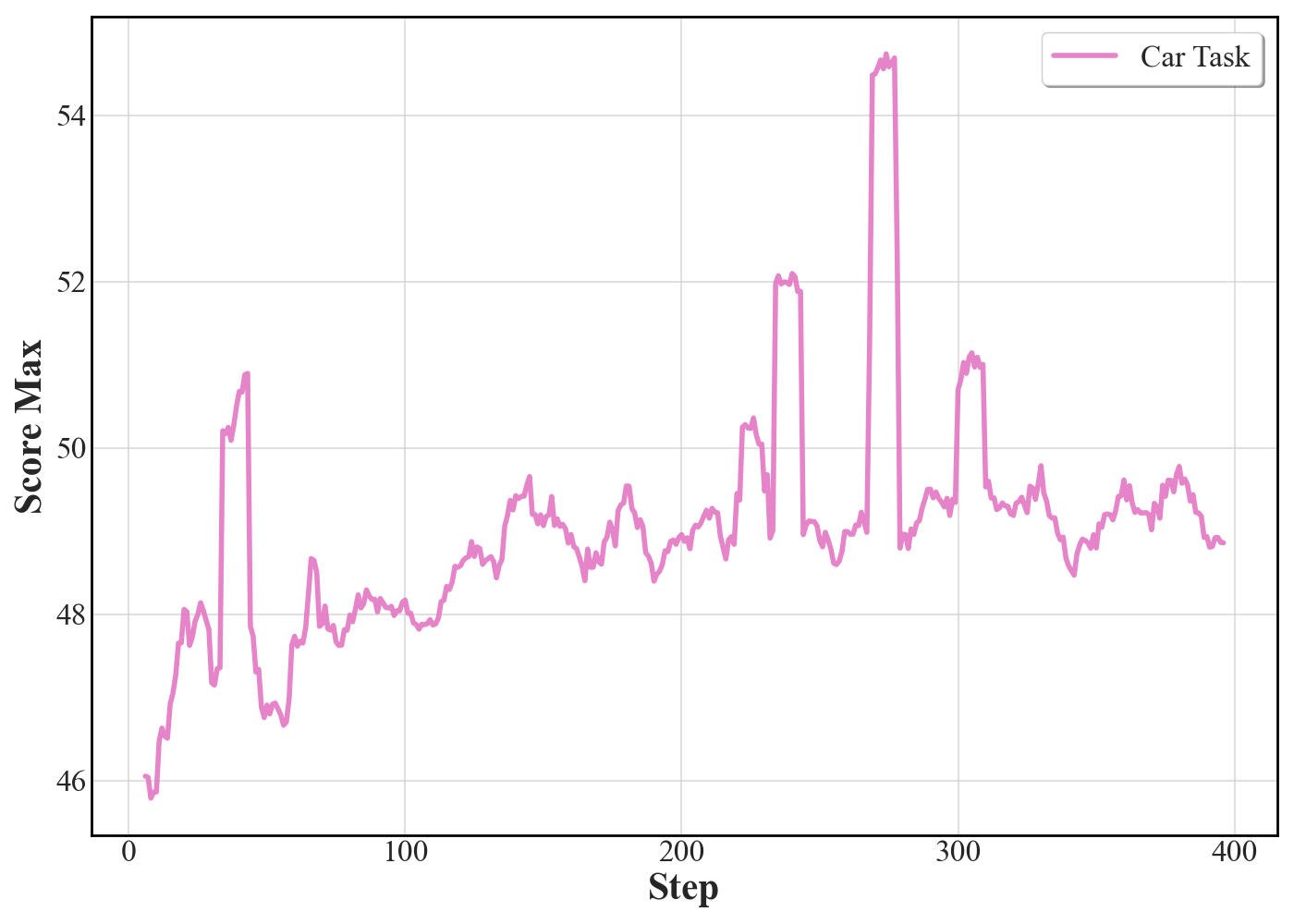}
  \caption{
  \textbf{Maximum machine score on the \textit{Car} task across RL training steps.} We use the same RL hyperparameter setting as in the base \textit{Catapult} experiment. Machine performance increases moderately during training.
  }
  \label{fig:RL-car-max-score}
\end{figure}

\begin{figure*}[h!]
  \centering
  \includegraphics[width=\linewidth]{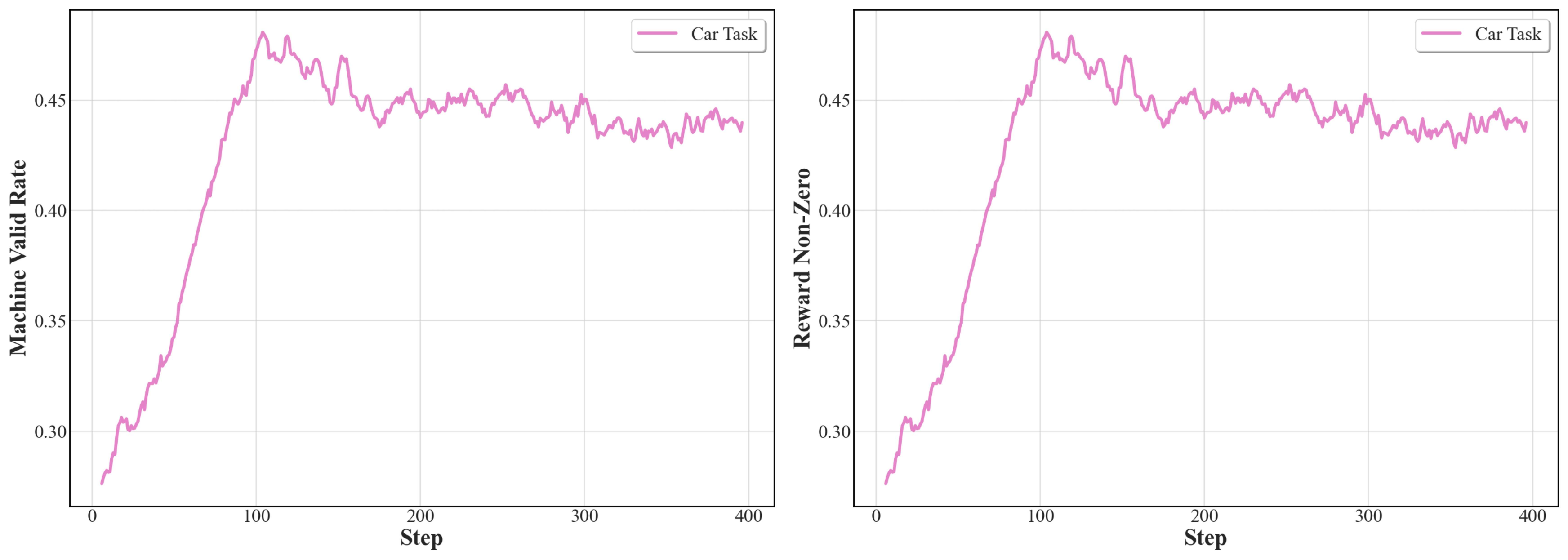}
  \caption{
  \textbf{Machine-validity rate and non-zero-reward rate on the \textit{Car} task across RL training steps.} Machine validity converges early and remains stable throughout the rest of training.
  }
  \label{fig:RL-car-valid-rate}
\end{figure*}

\begin{figure*}[h!]
  \centering
  \includegraphics[width=\linewidth]{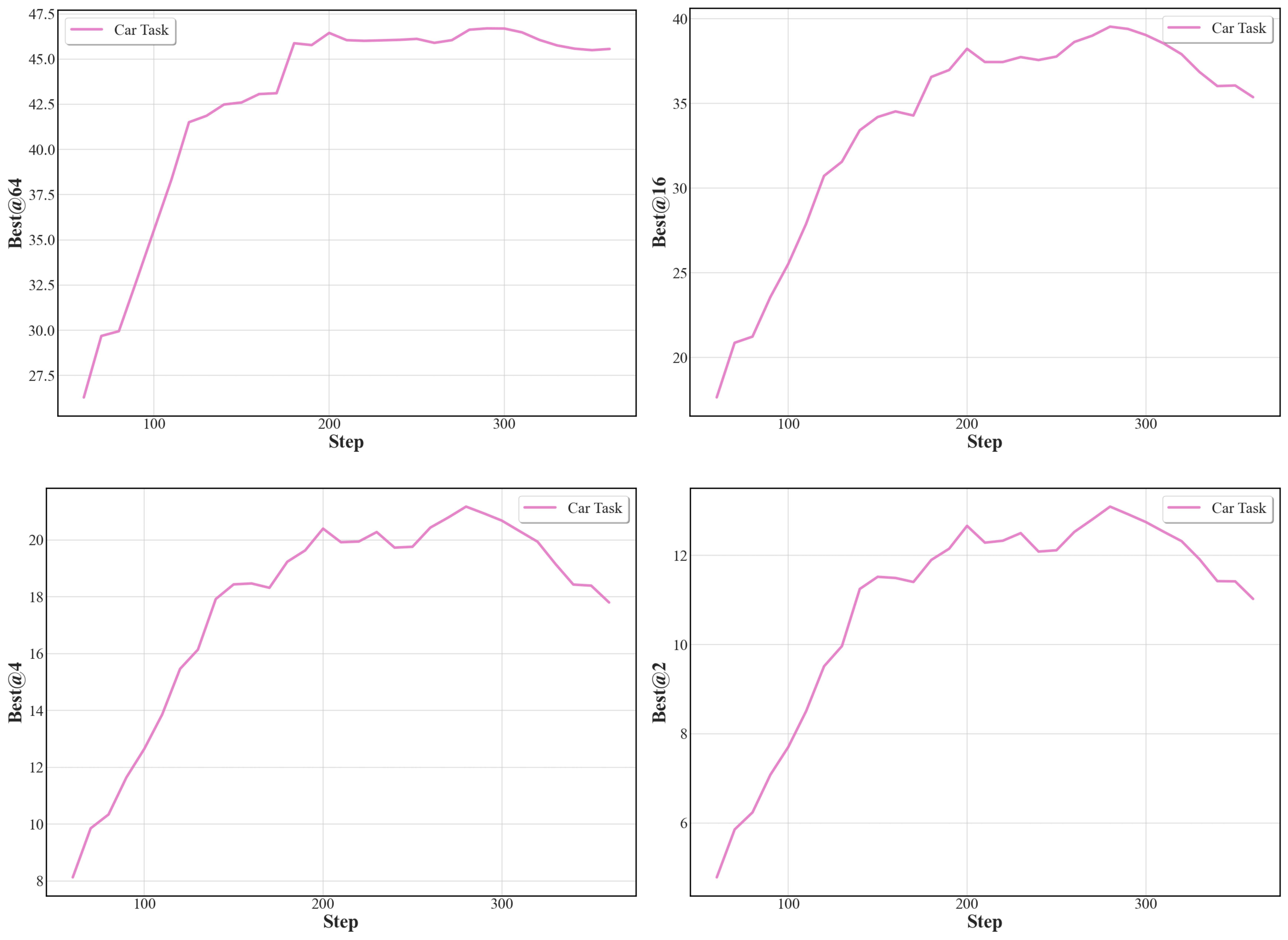}
  \caption{
  \textbf{Best@\(N\) performance on the \textit{Car} task.} Similar to the machine-validity trend, Best@\(N\) performance improves early in training and remains relatively stable afterward.
  }
  \label{fig:RL-car-best@n}
\end{figure*}

Across two random seeds, the Pass@64 setting achieves the strongest maximum score on Throwing. For completeness, we also report the corresponding Movement results in Fig.~\ref{fig:RL-car-max-score}, Fig.~\ref{fig:RL-car-valid-rate} and Fig.~\ref{fig:RL-car-best@n}.

\subsection{Uncertainty estimates and cross-task transfer}

Rewards in \envname are sparse and heavy-tailed, so we additionally report standard deviations and 95\% confidence intervals (bootstrap for means; Wilson for validity) over the 50 generations per setting, together with cross-task transfer without adaptation: the Throwing-trained model is evaluated on Movement and vice versa. 
The results suggest that the learned high-performance design priors do not arise purely from task-specific memorization.
\begin{table*}[h!]
  \scriptsize
  \centering
  \setlength{\tabcolsep}{6pt}
  \renewcommand{\arraystretch}{1.2}
  \begin{tabularx}{\textwidth}{l*{6}{>{\centering\arraybackslash}X}}
    \toprule
    \multirow{2}{*}{Task / Model} & Valid & Mean & Max & SD & \makecell{Mean CI\\(95\%)} & \makecell{Validity CI\\(95\%)} \\
    \midrule
    \multicolumn{7}{l}{\textit{Movement}} \\
    Movement---Base & 46/50 & 4.97 & 19.10 & 5.04 & [3.53, 6.40] & [81.2\%, 96.8\%] \\
    Movement---Cold-start & 40/50 & 4.67 & 20.23 & 4.92 & [3.26, 6.06] & [67.0\%, 88.8\%] \\
    Movement---RL from Base & 41/50 & 3.72 & 24.08 & 4.77 & [2.36, 5.09] & [69.2\%, 90.2\%] \\
    Movement---Cold-start + RL & 42/50 & 5.05 & 45.72 & 7.61 & [2.89, 7.21] & [71.5\%, 91.7\%] \\
    Movement---Throwing$\to$Movement & 28/50 & 4.57 & 43.70 & 10.66 & [1.95, 7.80] & [42.3\%, 68.8\%] \\
    \midrule
    \multicolumn{7}{l}{\textit{Throwing}} \\
    Throwing---Base & 11/50 & 0.06 & 2.41 & 0.34 & [0, 0.16] & [12.8\%, 35.2\%] \\
    Throwing---Cold-start & 9/50 & 0.11 & 5.54 & 0.78 & [0, 0.33] & [9.8\%, 30.8\%] \\
    Throwing---RL from Base & 12/50 & 0.13 & 5.92 & 0.83 & [0, 0.37] & [14.3\%, 37.4\%] \\
    Throwing---Cold-start + RL & 11/50 & 0.14 & 7.14 & 1.02 & [0, 0.43] & [12.8\%, 35.2\%] \\
    Throwing---Movement$\to$Throwing & 11/50 & 0.13 & 5.36 & 0.77 & [0, 0.37] & [12.8\%, 35.2\%] \\
    \bottomrule
  \end{tabularx}
  \caption{
  \textbf{Standard deviations and 95\% confidence intervals for the RL post-training evaluation.} We report the number of valid machines out of 50 generations, together with mean, maximum and standard deviation of simulation scores. Mean confidence intervals are bootstrap estimates and validity confidence intervals are Wilson estimates. Transfer rows evaluate each fine-tuned model on the other task family without adaptation.
  }
  \label{tab:rl_ci}
\end{table*}

\section{User study protocol}
\label{app:user_study}

This section describes the human-participant study used to compare LLM design behavior with short-exposure human design behavior in \textsc{\envname}. The study was designed as a non-expert behavioral reference rather than an expert human upper bound: participants received only brief instruction and practice before attempting the task. We accounted for participants' prior familiarity with \textit{Besiege} and related construction-oriented games when assigning them to experimental conditions, so that game experience was balanced across groups. After completing the study, each participant received 50 RMB as compensation, which was above the typical compensation level for comparable student-participant studies at the institution.

\subsection{Tasks and experimental conditions}

The user study used two tasks: \textit{Throwing} and \textit{Delivery}. These tasks were selected because they have clear quantitative objectives while eliciting different forms of mechanical reasoning. Throwing requires coordinated launcher-like mechanisms, whereas Delivery emphasizes stable transport of a payload across terrain.

Participants were assigned to one of two experimental conditions. In the \textit{Free Building} condition, participants received only the task objective and were allowed to construct freely. In the \textit{Instructed Building} condition, participants additionally received a high-level design rationale generated by Gemini 2.5 Pro before construction. Participants in the Instructed Building condition were not told that the rationale was generated by an AI model. Each participant completed exactly one task in exactly one condition, with no repeated task trials across conditions.

A total of 57 participants were recruited. The Free Building condition included 28 participants, with 14 assigned to Delivery and 14 assigned to Throwing. The Instructed Building condition included 29 participants, with 13 assigned to Delivery and 16 assigned to Throwing.

\subsection{Procedure}

Before the experiment, participants completed a pre-study questionnaire assessing their familiarity with games, AI tools and 3D spatial reasoning. Filler questions were included to reduce the likelihood that participants would infer the study hypotheses.

Participants then watched a 10-minute introductory video explaining the \textsc{\envname} interface, game controls and available building blocks. After the video, they completed a 20-minute practice phase using official \textit{Besiege} levels to gain hands-on familiarity with block placement and basic controls. Only after the practice phase were participants informed of their assigned task objective and experimental rules.

Each main task session was capped at 1.5 hours. Sessions reaching the time limit were terminated by the researcher. Participants followed a think-aloud~\cite{protocol_analysis} protocol and were asked to verbalize their reasoning during construction, testing and revision. They were allowed to pause and ask clarifying questions about the environment, controls or task rules. Participants were free to withdraw from the study at any time; no participant chose to withdraw.

All sessions were recorded through screen capture and audio. After completing the task, participants filled out a post-study questionnaire about their design process, construction decisions, perceived task difficulty and response to any provided design rationale.

\subsection{Ethics and consent}

All participants provided written informed consent before enrollment. The study protocol was approved by the relevant institutional ethics review board. Participants were informed that their screen and audio would be recorded for research purposes, and that they could stop participation at any time without consequence.

\begin{figure*}[tp]
  \centering
  \includegraphics[width=\linewidth]{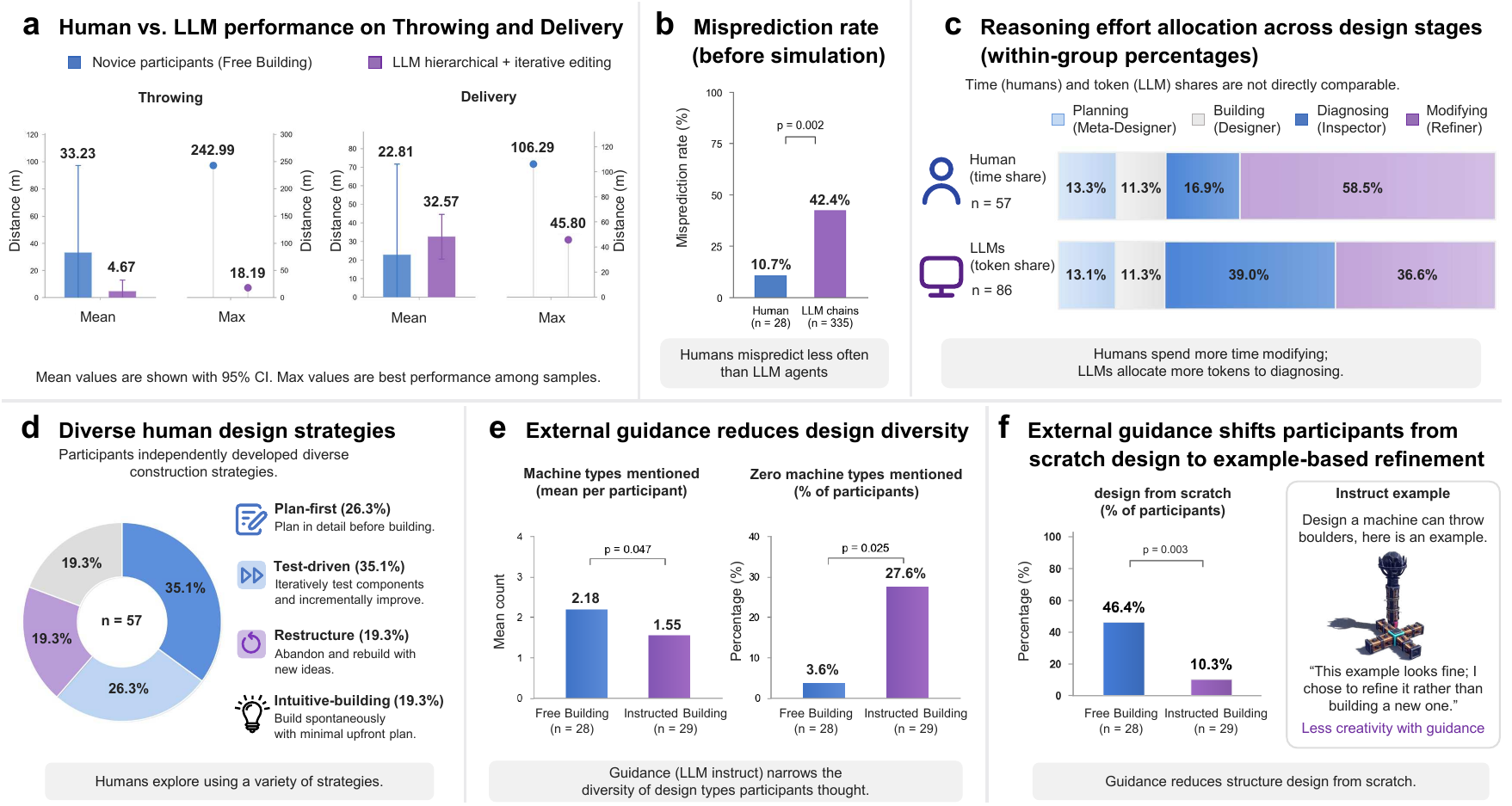}
    \caption{
        \textbf{Human and LLM design strategies in \textsc{\envname}.}
        \textbf{a,} Short-exposure human participants outperform LLM agents on Throwing and achieve comparable mean performance on Delivery. Bars show mean performance with 95\% confidence intervals, and points show best performance.
        \textbf{b,} Humans mispredict machine behavior less often than LLM agents before simulation.
        \textbf{c,} Reasoning-effort allocation differs between humans and LLM agents. Humans spend a larger share of time on simulation-based checking and refinement, whereas LLM agents allocate more tokens to diagnosis. Percentages are within-group proportions and are not unit-equivalent across humans and LLM agents.
        \textbf{d,} Human participants use diverse search strategies, including plan-first, test-driven, restructure/restart and intuitive-building strategies.
        \textbf{e,} External guidance from an LLM-generated rationale reduces the diversity of independently mentioned design types and increases the fraction of participants who mention no alternative design type. \(P\)-values indicate the statistical significance of between-condition differences.
        \textbf{f,} External guidance shifts participants from scratch design toward example-based refinement, suggesting that provided examples can narrow independent principle-first exploration.
    }
  \label{fig:nmi-human-study}
\end{figure*}

\subsection{Human Subject Study and Compensation}

Our user study was approved by the relevant institutional ethics review board. Participants were recruited as adult volunteers from the local university/community population through official recruitment channels of the institution. Before the study, participants were informed of the procedure, approximate duration, compensation, potential risks, data usage, and their right to take breaks or withdraw at any time without penalty. The study required participants to solve tasks in a physics-based construction game under different conditions, including with or without example guidance and LLM assistance, while their behaviors and think-aloud protocols were recorded. The study took approximately 1.5 hours. Each participant received 50 RMB for participation, including participants who withdrew before completion, corresponding to about 33 RMB/hour for a full session. All collected data were anonymized and encoded before analysis, and no personally identifiable information was included in LLM interactions.

\section{User study annotation and behavioral metrics}
\label{sec:userstudy_annotation}

This section describes how raw user-study recordings were transformed into structured behavioral episode sequences and how the behavioral metrics reported in the main text were computed. The annotation pipeline combines transcript segmentation, visual context from screen recordings and constrained LLM-assisted labeling.

\subsection{Annotation pipeline}

Each session recording was accompanied by a polished subtitle file produced using GPT-4o Transcribe. We parsed these subtitle files to extract per-utterance timestamps, transcribed text and speaker identity. Consecutive utterances were then merged into approximately 30-second microsegments, which served as the atomic units for subsequent annotation.

For each microsegment, a representative keyframe was extracted from the corresponding screen recording to provide visual context. Gemini-3.1-Flash was then applied to assign one of six coarse-grained labels to each microsegment, using the transcript text, extracted keyframe and a sliding-window summary of surrounding context. The six coarse labels were: \textit{Clarify constraints}, \textit{Plan or hypothesize}, \textit{Build or adjust}, \textit{Test or observe}, \textit{Diagnose or revise} and \textit{Reflect}.

\begin{figure*}[h!]
  \centering
  \includegraphics[width=\linewidth]{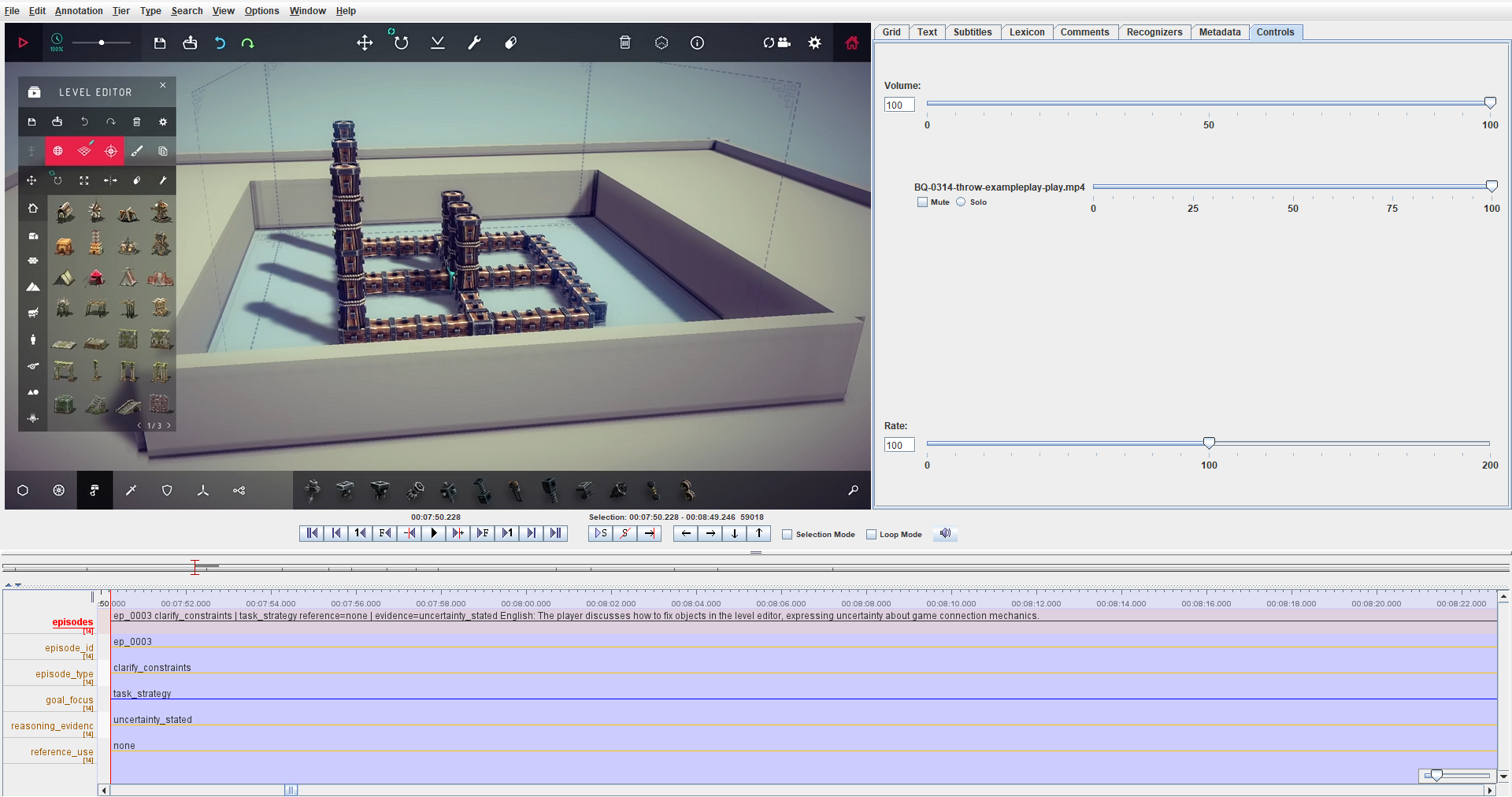}
  \caption{
  \textbf{Screenshot of the ELAN annotation interface used for manual inspection of user-study sessions.} Each row represents an annotation tier, with time-aligned segments displayed on the timeline. The upper panel shows the corresponding \textsc{\envname} screen recording.
  }
  \label{fig:userstudy-ELAN}
\end{figure*}

Adjacent microsegments sharing the same coarse label were merged into episodes using an LLM-assisted sliding-window procedure that determined whether consecutive segments represented a continuation of the previous episode or the beginning of a new one. The resulting episode sequences were exported in ELAN format~\citep{wittenburg2006elan} for video-aligned manual inspection, as shown in Fig.~\ref{fig:userstudy-ELAN}. All sessions were processed in a single batch using unified prompt versions and hyperparameters to ensure cross-session consistency.

\subsection{Fine-grained behavioral labels}

In a second pass, each coarse-labeled episode was refined into one of ten fine-grained behavioral categories by re-invoking Gemini-3.1-Flash with the episode transcript, representative keyframe and a participant profile derived from the post-study questionnaire. Two coarse labels were mapped deterministically: \textit{Clarify constraints} was mapped to \textit{Query}, and \textit{Reflect} was mapped to \textit{Reflect}. The remaining coarse labels were refined using constrained candidate sets, so that the model could only choose labels compatible with the coarse category. Labels outside the candidate set were rejected to preserve codebook consistency.

The ten fine-grained episode types are defined as follows. \textit{Plan} denotes goal-directed episodes in which the participant sets a subgoal or overall direction for the next construction phase. \textit{Hypothesize} denotes tentative causal inference about failure causes or design feasibility, distinguished from \textit{Diagnose} by the absence of confirmatory evidence. \textit{Build} denotes general construction episodes involving adding, removing or repositioning components without a specific failure context. \textit{Test} denotes active triggering of the \textsc{\envname} simulation. \textit{Observe} denotes passive perception of machine behavior following a test, without further active intervention. \textit{Diagnose} denotes episodes in which the participant arrives at an evidence-grounded explanation of why the machine failed or where the problem lies. \textit{Revise} denotes targeted local modifications made in response to prior observation or diagnosis, without changing the overall mechanism. \textit{Restructure} denotes substantial redesign or mechanism replacement that changes the core construction strategy. \textit{Reflect} denotes meta-level commentary on the process, experience or strategy. \textit{Query} denotes episodes in which the participant asks the experimenter about block functions, task rules or environmental constraints.

\subsection{AI-agent comparison protocol}

The AI-side data were obtained from outputs of the agentic pipeline, with task types and sample sizes aligned to the human-participant study where possible. The underlying LLM used by the agentic pipeline was sampled across runs rather than fixed to a single model instance. This comparison is intended to characterize broad behavioral differences between short-exposure human participants and LLM-agent workflows, rather than to equate human time with LLM token use.

\subsection{Construction-style clustering}

To characterize recurring human construction styles, we applied hierarchical clustering~\citep{johnson1967hierarchical} to fine-grained behavior sequences using Levenshtein edit distance~\citep{levenshtein1966binary} and average linkage. The number of clusters was selected by maximizing the silhouette coefficient~\cite{silhouettes}. These clusters were used as descriptive summaries of behavioral diversity rather than as stable psychological traits.

\subsection{Reasoning-effort allocation}

To compare human behavioral allocation with the stages of the agentic pipeline, we mapped fine-grained human episodes to LLM-agent stages. \textit{Plan} and \textit{Hypothesize} were mapped to the meta-designer stage; \textit{Build} was mapped to the designer stage; \textit{Observe}, \textit{Diagnose} and \textit{Reflect} were mapped to the inspector stage; and \textit{Revise} and \textit{Restructure} were mapped to the refiner stage.

For human participants, reasoning-effort allocation was computed from the total duration of episodes assigned to each mapped stage. For LLM agents, the corresponding quantity was computed from token counts assigned to each agent stage. Because minutes and tokens are not directly comparable units, we report within-group proportions: human time shares sum to one across stages, and LLM token shares sum to one across stages.

\subsection{Surface-level diagnosis}

We define surface-level diagnosis as proceeding from a post-simulation episode directly to construction or local revision without an intervening planning or hypothesis episode. Operationally, after an episode of \textit{Test}, \textit{Observe}, \textit{Diagnose} or \textit{Revise}, we examined at most the next five episodes. If the participant proceeded directly to \textit{Build} or \textit{Revise}, the case was labeled as surface-level diagnosis. If the participant first entered \textit{Plan} or \textit{Hypothesize} before later building or revising, the case was labeled as deeper diagnosis.

\subsection{Design-type diversity under guidance}

To quantify whether external guidance narrowed exploration, we measured the diversity of machine structure types mentioned by participants. We used a curated keyword dictionary over interview transcripts and open-ended questionnaire responses. The dictionary retained only terms that unambiguously referred to a specific mechanical configuration, such as \textit{slingshot}, \textit{catapult}, \textit{pulley} and \textit{ramp}.

Using this procedure, we identified an average of 2.18 structure types per participant in the Free Building condition and 1.55 structure types per participant in the Instructed Building condition. The proportion of participants for whom no structure type was identified increased from 3.6\% in the Free Building condition to 27.6\% in the Instructed Building condition. 
These statistics provide evidence that external high-level guidance can reduce independent exploration of alternative machine structures.

\subsection{Foundation-first reasoning}

We coded foundation-first reasoning from interview transcripts and open-ended questionnaire responses. A participant was labeled as foundation-first if their first planning-related response emphasized the base, chassis, support, stability or reinforcement of the machine. The keyword set included terms such as \textit{base}, \textit{chassis}, \textit{support}, \textit{stability} and \textit{reinforcement}, together with manually checked synonymous expressions.

\subsection{Machine-behavior misprediction}

For human participants, machine-behavior misprediction was coded from open-ended questionnaire responses and manual inspection of experiment recordings. A case was labeled as misprediction-then-correction when a participant expressed an expectation that the machine would behave successfully or in a specific way, then revised that belief after observing a failed or unexpected simulation.

For LLM agents, we identified analogous cases using text from the refiner and inspector stages. The refiner stage was checked for optimistic predictions using terms such as \textit{stable}, \textit{succeed}, \textit{success} and \textit{accelerates}. The subsequent inspector stage was checked for failure-related terms such as \textit{fail}, \textit{failed}, \textit{failure} and \textit{crash}. Cases matching this success-before-failure pattern were labeled as machine-behavior misprediction.

\subsection{Statistical analysis}

The choice of statistical analysis was determined by the specific research question and data type. Performance comparisons used task-specific scores from the final or best machine produced under the study budget. Behavioral proportions, such as misprediction rate, foundation-first reasoning and design-type diversity, were computed at the participant level for human data and at the run level for LLM-agent data. Exact statistical tests and sample sizes for each reported comparison are provided alongside the corresponding figures and tables.

\section{Participant profile and additional user-study results}
\label{sec:userstudy_profile}

This section summarizes participant backgrounds and additional questionnaire results from the user study. These results provide context for interpreting the behavioral comparisons in the main text. The study was designed as a short-exposure non-expert reference; participants were not selected for expertise in engineering design or expert-level \textit{Besiege} gameplay.

\subsection{Participant profile}

\begin{figure*}[h!]
  \centering
  \includegraphics[width=\linewidth]{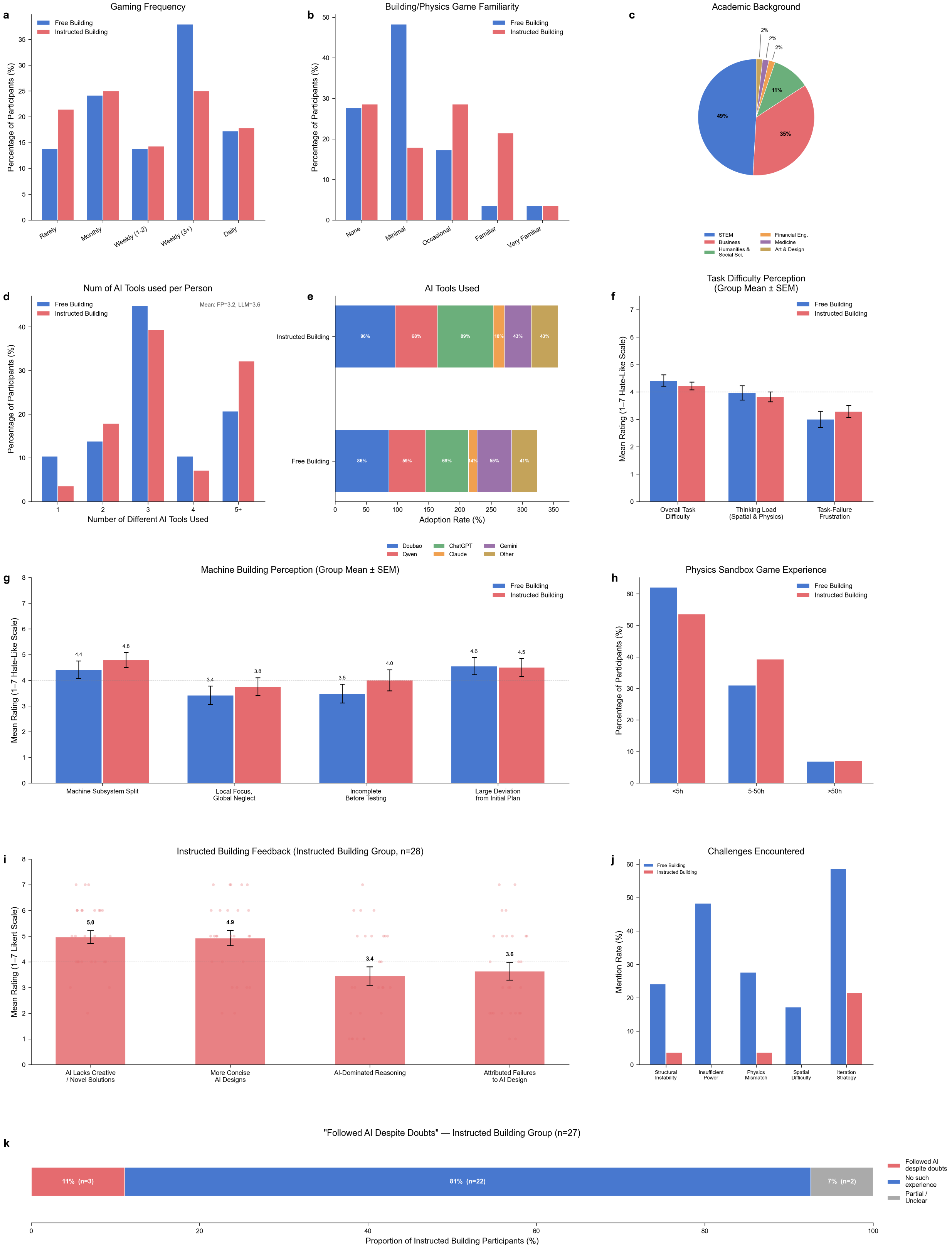}
  \caption{
  \textbf{Participant profile for the user study.} The profile summarizes background variables relevant to construction experience, spatial reasoning, gaming familiarity and AI-tool usage.
  }
  \label{fig:userstudy-participant}
\end{figure*}

A total of 57 participants were recruited for the user study. The Free Building condition included 28 participants, with 14 assigned to Delivery and 14 assigned to Throwing. The Instructed Building condition included 29 participants, with 13 assigned to Delivery and 16 assigned to Throwing. A summary of participant backgrounds is shown in Fig.~\ref{fig:userstudy-participant}.

\subsection{Gaming experience}

Participants reported a broad distribution of general gaming experience. Approximately half of the participants reported prior experience with construction or 3D spatial-manipulation games, such as sandbox builders or block-placement games, while roughly one third reported no such experience. Around two thirds of participants reported having played building or physics-simulation games at least once. This distribution suggests that the cohort was not entirely unfamiliar with game-like spatial interaction, but was not composed of expert builders or expert \textit{Besiege} players.

\subsection{Academic background}

Participants were approximately balanced between STEM and non-STEM academic backgrounds. This balance reduces the likelihood that the study results are dominated by participants with specialized engineering, mechanics or spatial-reasoning training.

\subsection{AI-tool usage}

Participants reported familiarity with an average of 3.4 distinct AI tools. We did not observe a substantial difference in self-reported AI-tool familiarity between the Free Building and Instructed Building conditions, suggesting that prior exposure to AI tools was broadly balanced across conditions.

\subsection{Perceived task difficulty}

Post-study questionnaire responses indicated that participants generally rated the tasks as moderately difficult, with a slight skew toward the harder end of the scale. This suggests that Throwing and Delivery provided the intended level of difficulty: they were challenging enough to elicit non-trivial construction strategies, but not so difficult that participants were unable to make progress within the allotted session time.

\subsection{Construction approach}

Participants varied substantially in how they approached construction. Some participants reported planning the overall mechanism before placing many blocks, whereas others described a more iterative process of building, testing and revising. Questionnaire responses also varied in whether participants reported decomposing the machine into subsystems before construction or modifying the entire machine holistically. This variability is consistent with the strategy diversity reported in the main text.

\subsection{Response to external guidance}

Among participants in the Instructed Building condition, responses to the provided design rationale varied. Some participants reported that the rationale was useful for understanding the task or identifying a starting design, while others found it difficult to interpret or felt that it constrained their own design thinking. This heterogeneity is consistent with the quantitative finding that external guidance reduced design-type diversity on average, while its effect on individual construction behavior varied across participants.

\subsection{Reported challenges}

Participants in the Free Building condition most commonly reported structural instability and balance as major challenges. Typical failure modes included machines breaking apart after activation, tipping over due to an unbalanced center of mass, or failing to maintain stable contact with the ground or payload. These responses suggest that structural stability and balance were salient bottlenecks for participants without external design guidance.

Participants also reported difficulties in predicting the behavior of dynamic mechanisms before simulation, especially in the Throwing task. Even when participants understood the intended mechanism at a high level, they often found it difficult to anticipate how actuator placement, pivot location or payload position would affect the resulting trajectory. This is consistent with the main-text observation that interpreting and acting on physical feedback is intrinsically challenging in compositional machine design, although human participants still mispredicted behavior less frequently than LLM agents.

\section{Data, code and model availability}

The \envname environment, agentic design pipeline, Gemini 2.5 Pro-generated dataset, and RL-fine-tuned models are publicly available at our project page (besiegefield.github.io). Licensing information for individual released artifacts is provided with the corresponding resources.

\section{Prompt templates and implementation details}
\label{sec:prompts}

This section describes the prompt structure used by the LLM agents. All agent prompts share a common context containing the task objective, available block types, construction rules and output-format requirements. Specialized agents receive additional context depending on their role, such as a meta-designer blueprint, partial machine program, inspector report, edit history or simulation feedback.

\subsection{Shared prompt context}

Each agent is provided with a shared description of the \textsc{\envname} construction setting. This shared context includes: (1) the task objective; (2) the available block library; (3) construction rules, including parent--child attachment constraints and special two-parent blocks; (4) the construction-tree JSON format; and (5) validity requirements. The shared context is intended to make the output directly parseable and to reduce ambiguity in component naming, parent references and attachment-face identifiers.

A simplified template for the shared context is shown below. The full prompt used in experiments includes the complete block library and output-format specification.

\begin{lstlisting}
You are designing a machine in BesiegeField.

Task objective:
{TASK_DESCRIPTION}

Available blocks:
{BLOCK_LIBRARY}

Construction rules:
- Machines are built from a root block.
- Each new block must attach to an existing parent block.
- Each attachment must specify the parent block id and attachment face.
- Special blocks such as springs may have two parent attachments.
- The output must be a valid construction-tree JSON program.

Validity requirements:
- All block ids must be unique.
- Parent ids must refer to previously created blocks.
- Required fields must be present for every block.
- The machine should avoid self-collisions and invalid attachments.
\end{lstlisting}

\subsection{Designer prompt}

The designer generates a machine program from the task objective and construction context. In the single-agent setting, the designer generates the full machine. In the hierarchical setting, the designer receives a functional module from the meta-designer and constructs only the corresponding part of the machine, conditioned on the partial construction built so far.

\begin{lstlisting}
You are the Designer agent.

Your goal is to construct a machine or machine module that satisfies the task objective.

Task objective:
{TASK_DESCRIPTION}

High-level plan or module to build:
{MODULE_DESCRIPTION}

Current partial machine:
{PARTIAL_MACHINE_JSON}

Use the available blocks and construction rules to produce a valid construction-tree JSON program.
First reason briefly about the intended mechanism.
Then output the final machine program in the required JSON format.
\end{lstlisting}

In all experiments, the machine program is parsed automatically. Text outside the designated JSON output region is ignored during parsing.

\subsection{Meta-designer prompt}

The meta-designer decomposes the task into functional modules before construction. It does not output a machine program directly. Instead, it identifies the major mechanical subsystems and specifies how they should interact.

\begin{lstlisting}
You are the Meta-Designer agent.

Given the task objective and available blocks, decompose the target machine into functional modules.
For each module, describe:
1. its function;
2. the main blocks likely needed;
3. how it should connect to other modules;
4. the physical role it plays in the final machine.

Task objective:
{TASK_DESCRIPTION}

Available blocks:
{BLOCK_LIBRARY}

Output a concise module-level blueprint.
\end{lstlisting}

The module-level blueprint is then provided to designer agents during hierarchical construction.

\subsection{Inspector prompt}

The inspector critiques a candidate machine before simulation-guided refinement. The inspector receives the task objective, the generated construction-tree program and, when available, the meta-designer blueprint. It identifies likely structural or conceptual issues without using new simulation feedback.

\begin{lstlisting}
You are the Inspector agent.

Inspect the following machine program for the task objective.
Identify likely issues that may prevent the machine from working.

Task objective:
{TASK_DESCRIPTION}

Machine program:
{MACHINE_JSON}

Optional high-level blueprint:
{BLUEPRINT}

Consider:
- missing functional components;
- incorrect use of block affordances;
- unstable structure;
- incorrect attachment or orientation;
- insufficient actuation or control;
- mismatch between the blueprint and the actual construction.

Output a concise issue report and suggested repair directions.
\end{lstlisting}

The inspector is invoked once after the initial construction stage. Its report is provided to the first refiner.

\subsection{Environment-querier prompt}

The environment querier summarizes simulation results and may request targeted block-level state information. The querier receives the task objective, the current machine, scalar task score and default simulation summary. It then selects which additional state variables are useful for diagnosing failure.

\begin{lstlisting}
You are the Environment Querier.

A candidate machine has been simulated in BesiegeField.
Summarize the observed behavior and identify which additional state information is needed.

Task objective:
{TASK_DESCRIPTION}

Machine program:
{MACHINE_JSON}

Default simulation summary:
{DEFAULT_FEEDBACK}

Task score:
{SCORE}

You may request targeted feedback using:
- block id;
- time interval;
- state variable: position, orientation, velocity, or spring length.

Output:
1. a concise behavioral summary;
2. likely failure causes;
3. optional targeted feedback queries.
\end{lstlisting}

The query output is executed by the environment interface, and the returned state summaries are passed to the refiner.

\subsection{Refiner prompt}

The refiner edits an existing machine based on the task objective, current machine program, inspector report, edit history and simulation feedback. The refiner is instructed to preserve useful structure when possible, but to perform larger restructuring when local repair is unlikely to solve the failure.

\begin{lstlisting}
You are the Refiner agent.

Revise the current machine to improve task performance.

Task objective:
{TASK_DESCRIPTION}

Current machine program:
{MACHINE_JSON}

Inspector report:
{INSPECTOR_REPORT}

Simulation feedback:
{SIMULATION_FEEDBACK}

Edit history:
{EDIT_HISTORY}

Your revision should:
- fix invalid or weak structures;
- preserve useful components when possible;
- change the mechanism if local edits are insufficient;
- output a valid construction-tree JSON program.

First explain the intended modification briefly.
Then output the revised machine program in the required JSON format.
\end{lstlisting}

When multiple candidate refinements are needed, the same prompt is sampled multiple times with independent decoding seeds.

\subsection{Output parsing and retry policy}

For each generated output, we extract the designated JSON program and attempt to parse it into a construction tree. Outputs that cannot be parsed or that violate required structural fields are treated as file-invalid. During generation and refinement, invalid outputs may be retried up to the retry limit specified for that stage. In the tree-search refinement loop, each node allows up to five retries to obtain valid child candidates. If the retry budget is exhausted, the parent machine is retained as a fallback candidate.

\subsection{Prompt versioning}

All main experiments use fixed prompt templates for each agent role. Prompt changes are treated as separate experimental conditions and are not mixed within a reported result. For the user-study annotation pipeline, all sessions are similarly processed using a fixed annotation prompt version and fixed model settings.

\section{Gemini 2.5 Pro design-rationale examples}
\label{sec:gemini_cot_samples}

This section provides representative Gemini 2.5 Pro design rationales used in the cold-start dataset and rationale-conditioning experiments. We include these examples to illustrate the type of high-level mechanical planning produced by the strongest model in our benchmark. The examples are not used as human-written expert solutions; they are model-generated rationales paired with machine programs and filtered by validity.

\subsection{Single Agent}
\begin{lstlisting}
system:
<System Prompt>
user:
Design a machine to throw a boulder (type id 36) in a parabolic trajectory. You must use exactly one boulder. Optimize for throw distance, height, and speed. Apply the principle of leverage correctly, balancing the weight of the projectile.

assistant:
<think>
**Analyzing Catapult Components**
I'm now building a catapult with a rigid frame using `Wooden Block`s to ensure stability. I plan to use a `Log` as the throwing arm, attached to a constructed pivot point. The `Container` will hold the boulder at the arm's end.

**Devising Launch Mechanisms**
I'm reconsidering my approach. With the provided constraints, a trebuchet is out. The single boulder and lack of hinges necessitate a catapult. I must leverage the `Rubber Band`. My plan is to design a tension-based catapult for the boulder, prioritizing stability and throwing arm efficiency.

**Optimizing for Distance**
I'm now focused on the tension-powered catapult. My plan is for a solid base, a pivot, a lever arm, a boulder-holding container, and rubber bands. I'm working on the arm's length and the positioning of the rubber bands for maximum launch power and the best trajectory for the boulder.

**Assessing the Lever Arm**
I'm now working on the arm, crucial for force. I must decide on its length and how it attaches to the pivot. I'm considering different attachment methods to create a stable but rotatable lever, exploring options for maximum launch power and height.

**Evaluating Construction Strategies**
I'm stuck! I need to use a catapult given the constraints. But, the lack of a hinge restricts my options. I'm exploring different ways to connect the throwing arm to the frame and leverage the `Rubber Band` for propulsion.

**Designing a Hybrid Catapult**
I've sketched out the structure, pivoting on the `Rotating Block`, with a base and uprights. The throwing arm will comprise a `Log`, with a `Container` at the tip. I am struggling with the rubber band attachment. I will have to ensure the right angles and tension.

**Evaluating Rubber Band Placement**
I'm now focusing on the final adjustments. With the core design finalized, the key is the `Rubber Band` placement. The goal is maximizing launch power by carefully considering the band's attachment points and the angle of force application. I aim to achieve a long throw with this arrangement.

**Evaluating Lever Arm Placement**
I'm now focused on the throwing arm's construction and its link to the rotating pivot. I've built the base and the pivot structure. The lever arm's length and the location for the `Container` are my next priorities. The rubber band attachments also need to be evaluated.

**Engineering the Launch Mechanism**
I'm now focused on a tension-based catapult. It's time to build, starting with a solid base and a stable pivot point using the `Rotating Block`. My priorities are a strong throwing arm, the right `Container` for the boulder, and, critically, the strategic placement of the `Rubber Band`s.

**Refining the Design**
I'm now focused on the core catapult design. The need for a simple build, given the constraints, favors a straightforward tension catapult. I am planning a solid base, a lever arm (most likely a `Log`), a container for the boulder, and the crucial application of `Rubber Band`s to generate the launch force.

**Finalizing Catapult Design**
I've spent time analyzing the limitations and refining the catapult build. I've now decided on the layout. The catapult will be built around a sturdy frame, a pivot using a `Rotating Block`, an arm (lever), and the utilization of the `Rubber Band` for propulsion. I am ready for the final assembly.

**Constructing the Base**
I'm now focused on the basic frame. I've decided to construct it from logs for stability. The goal is a sturdy structure to anchor the catapult's components, primarily the `Rotating Block` and lever arm. I'm focusing on the geometry.

**Analyzing Leverage Application**
I'm now focused on catapult design. Given the one-boulder rule and lack of hinges, I'll focus on a lever-based catapult. I am now determining the lever arm length.

**Building a Lever Catapult**
I'm planning a basic tension-based catapult using the `Rotating Block` as a pivot. I'm focusing on the throwing arm's dimensions, the `Rubber Band` placement to maximize launch power, and the `Container` block to hold the boulder.

**Constructing Launch Support**
I'm now focused on the basic design. With the user's need to launch a single boulder, a tension-based catapult is optimal. I'll build it around the `Rotating Block`. I'm using a `Log` for the arm and planning a base, rubber band connections, and the boulder's container.

**Planning the Build**
I'm streamlining the design. The key will be a stable base, a `Rotating Block` for the pivot, and a lever arm. I'll use a `Container` to hold the boulder, and the `Rubber Band`s will be the power source. Now I am evaluating the dimensions and the best placement of each part.

**Constructing the Launching Mechanism**
I'm designing a simplified catapult, keeping the focus on its functionality and the user's requirement for a single boulder and the available blocks. I plan to use a `Rotating Block` as the pivot. The rest of the design must center around this `Rotating Block` component.

**Constructing the Catapult**
I've established the foundation of the catapult, focusing on a solid base and pivot mechanism. The challenge now is to construct the lever arm and connect it effectively to the `Rotating Block`, while also determining how to hold the boulder. I will prioritize a design maximizing force transfer for launch.

**Constructing a Lever Arm**
I'm solidifying the basic build: a catapult, powered by rubber bands, given the constraints. The next step is constructing the lever arm, deciding its dimensions and how it will attach to the pivot point for optimal force transfer.

**Constructing the Final Catapult**
I've finalized the construction plan. The tension-based catapult, using the `Rotating Block` for the pivot and a lever arm, is my focus. I will use a `Container` for the boulder, and the `Rubber Band` for launching, paying close attention to force application and frame design.

**Outlining Catapult Construction**
I've sketched out the build, focusing on a tension-powered catapult. A base, a pivoting lever arm, and a boulder holder are planned. My design will depend on a stable foundation and the proper application of rubber bands for generating enough force for distance and height.

**Constructing the Catapult**
I've arrived at a straightforward plan for a tension-powered catapult, using the available blocks. I'm focusing on a design to maximize launch potential. The base, pivot point, arm, and rubber bands will be the key components for an efficient, functional build.

**Devising the Final Catapult**
I've finalized the core design. I am now sure of a tension-based catapult; the concept is simple and effective. My focus is now on the construction idea and creating the blocks. I am convinced it will be a simple build and the machine can leverage the `Rotating Block` and `Rubber Band` to meet the user's launch goals.

**Constructing a Launching System**
I'm finalizing the blueprint now. The tension-based catapult will use a base, a pivot with the `Rotating Block`, and a lever arm for the boulder. I'll ensure maximum launch power through strategic rubber band placement and precise measurements. I need a strong yet streamlined design.

**Constructing a Functional Catapult**
I've finalized the plan. The design uses a tension-based catapult, with a "Rotating Block" for a pivot. A sturdy base with two Log arms ensures stability, while two Rubber Bands generate the launch force. The boulder holder is a "Container", and the system prioritizes lever arm efficiency for maximum height and distance.

**Formulating Final Design**
I'm now certain: a tension-based catapult, incorporating the `Rotating Block` for the pivot, will meet the user's needs. The goal is maximum launch power within the given constraints. I'll sketch out the construction, outlining the key blocks and how they fit together. I've a solid plan.

Construction Idea:
I will design a tension-powered catapult to launch the boulder. The design prioritizes a long throwing distance and height by utilizing a long lever arm.

The machine has four main parts:
1.  **Base Frame:** A T-shaped base, constructed from a central 'Wooden Block' and two 'Log's extending sideways, provides a wide, stable footprint to handle the throwing forces and prevent the machine from tipping over.
2.  **Support Tower:** Two 'Log's rise vertically from the ends of the base, creating a tall, sturdy support structure for the pivot axle.
3.  **Lever Arm Assembly:** This is the core of the catapult. A 'Rotating Block' is used as a functional hinge/axle, mounted on one of the support uprights. A 'Small Wooden Block' acts as a hub on this axle, with a long 'Log' attached pointing up (the throwing arm) and a 'Small Wooden Block' attached pointing down (the short lever). This structure creates the necessary leverage. At the end of the throwing arm, a 'Container' is mounted via an adapter block to hold the 'Boulder'.
4.  **Power System:** Two 'Rubber Band' blocks provide the propulsive force. They connect from a forward-projecting 'Log' on the base to the short lever arm. The stretched rubber bands contract rapidly, pulling the short arm down and forward, which in turn whips the long throwing arm upwards and forwards, launching the boulder in a high parabolic trajectory.

This design maximizes the lever ratio to achieve high velocity for the projectile, while the robust frame maintains stability during the launch.

necessary_blocks ```[1, 63, 22, 15, 30, 36, 9]```
</think>

JSON:
```json
[
{"type": 0, "id": 0, "parent": -1, "face_id": -1}
{"type": 1, "id": 1, "parent": 0, "face_id": 1}
{"type": 63, "id": 2, "parent": 1, "face_id": 1}
{"type": 63, "id": 3, "parent": 1, "face_id": 3}
{"type": 63, "id": 4, "parent": 2, "face_id": 6}
{"type": 63, "id": 5, "parent": 3, "face_id": 6}
{"type": 22, "id": 6, "parent": 4, "face_id": 4}
{"type": 15, "id": 7, "parent": 6, "face_id": 3}
{"type": 63, "id": 8, "parent": 7, "face_id": 3}
{"type": 15, "id": 9, "parent": 7, "face_id": 4}
{"type": 15, "id": 10, "parent": 8, "face_id": 0}
{"type": 30, "id": 11, "parent": 10, "face_id": 3}
{"type": 36, "id": 12, "parent": 11, "face_id": 0}
{"type": 63, "id": 13, "parent": 0, "face_id": 0}
{"type": 9, "id": 14, "parent_a": 13, "face_id_a": 1, "parent_b": 9, "face_id_b": 1}
{"type": 9, "id": 15, "parent_a": 13, "face_id_a": 3, "parent_b": 9, "face_id_b": 2}
]
```
\end{lstlisting}

\subsection{Meta Designer}
\begin{lstlisting}
system:
<System Prompt>

user:
Design a machine to throw a boulder (type id 36) in a parabolic
trajectory. You must use exactly one boulder. Optimize for throw distance
, height, and speed. Apply the principle of leverage correctly, balancing
the weight of the projectile.

meta-designer output:
```json
{
    "definition": "Construct a machine that can throw boulders in a parabolic trajectory, designed to maximize throw distance and height by utilizing the principle of leverage.",
    "function_points": [
        "Establish a stable foundation to withstand recoil and prevent tipping.",
        "Implement a powered lever arm for projectile propulsion.",
        "Carry and release a boulder projectile at the apex of the arm's swing."
    ],
    "design_structure": [
        {
            "function_name": "Base and Support Frame",
            "description": "A wide, static frame built from the Start Block. It provides stability for the machine and includes a vertical support to mount the rotating mechanism, as well as a stop block to halt the throwing arm's motion.",
            "related_function_points": [
                "Establish a stable foundation to withstand recoil and prevent tipping."
            ]
        },
        {
            "function_name": "Powered Throwing Arm",
            "description": "A composite structure consisting of a powered pivot, a long lever arm, and a projectile holder. The Rotating Block provides rapid rotational force, the Log acts as the lever to amplify speed, and the Container holds the Boulder, releasing it via inertia.",
            "related_function_points": [
                "Implement a powered lever arm for projectile propulsion.",
                "Carry and release a boulder projectile at the apex of the arm's swing."
            ]
        }
    ],
    "build_order": [
        "Base and Support Frame",
        "Powered Throwing Arm"
    ],
    "machine_structure": {
        "Base and Support Frame": {
            "type": [
                1,
                15
            ],
            "guidance": "Wooden Blocks (ID 1) and Small Wooden Blocks (ID 15) are used to build a wide, heavy base that prevents tipping. They also form a tower to elevate the pivot point and a physical stop to arrest the arm's rotation, which is crucial for the release mechanism."
        },
        "Powered Throwing Arm": {
            "type": [
                22,
                63,
                30,
                36
            ],
            "guidance": "A Rotating Block (ID 22) provides the high-speed rotational power. This is attached to a Log (ID 63), which acts as a long, robust lever arm. A Container (ID 30) is placed at the end of the arm to hold the Boulder (ID 36) projectile."
        }
    }
}
```

\end{lstlisting}

\subsection{Designer}
\begin{lstlisting}
system:
<System Prompt>

designer-layer1 input:
On the basis of the existing structure, generate [Base and Support Frame], [A wide, static frame built from the Start Block. It provides stability for the machine and includes a vertical support to mount the rotating mechanism, as well as a stop block to halt the throwing arm's motion.], as ['Establish a stable foundation to withstand recoil and prevent tipping.'], using block type[1, 15], [Wooden Blocks (ID 1) and Small Wooden Blocks (ID 15) are used to build a wide, heavy base that prevents tipping. They also form a tower to elevate the pivot point and a physical stop to arrest the arm's rotation, which is crucial for the release mechanism.]

designer-output-layer1:
Based on the existing structure, a Base and Support Frame was generated to establish a stable foundation to withstand recoil and prevent tipping. Ultimately, the block types [1, 15] were decided upon. No blocks were removed as they were all suitable for the design.
JSON:
```json
[
{"type": "0", "id": 0, "parent": -1, "face_id": -1}
{"type": "1", "id": 1, "parent": 0, "face_id": 1}
{"type": "1", "id": 2, "parent": 0, "face_id": 2}
{"type": "1", "id": 3, "parent": 0, "face_id": 0}
{"type": "1", "id": 4, "parent": 1, "face_id": 2}
{"type": "1", "id": 5, "parent": 2, "face_id": 4}
{"type": "1", "id": 6, "parent": 0, "face_id": 3}
{"type": "1", "id": 7, "parent": 6, "face_id": 6}
{"type": "1", "id": 8, "parent": 7, "face_id": 6}
{"type": "15", "id": 9, "parent": 8, "face_id": 0}
]
```

designer-layer2:
On the basis of the existing structure, generate [Powered Throwing Arm], [A composite structure consisting of a powered pivot, a long lever arm, and a projectile holder. The Rotating Block provides rapid rotational force, the Log acts as the lever to amplify speed, and the Container holds the Boulder, releasing it via inertia.], as ['Implement a powered lever arm for projectile propulsion.', "Carry and release a boulder projectile at the apex of the arm's swing."], using block type [22, 63, 30, 36], [A Rotating Block (ID 22) provides the high-speed rotational power. This is attached to a Log (ID 63), which acts as a long, robust lever arm. A Container (ID 30) is placed at the end of the arm to hold the Boulder (ID 36) projectile.]
<3D Info>

designer-output-layer2:
Based on the existing structure, a Powered Throwing Arm was generated as Implement a powered lever arm for projectile propulsion.,Carry and release a boulder projectile at the apex of the arm's swing.. 
Ultimately, the block types [22, 63, 30, 36] were decided upon, while [] were removed due to [].
JSON:
```json
[
  {"type": "0", "id": 0, "parent": -1, "face_id": -1},
  {"type": "1", "id": 1, "parent": 0, "face_id": 1},
  {"type": "1", "id": 2, "parent": 0, "face_id": 2},
  {"type": "1", "id": 3, "parent": 0, "face_id": 0},
  {"type": "1", "id": 4, "parent": 1, "face_id": 2},
  {"type": "1", "id": 5, "parent": 2, "face_id": 4},
  {"type": "1", "id": 6, "parent": 0, "face_id": 3},
  {"type": "1", "id": 7, "parent": 6, "face_id": 6},
  {"type": "1", "id": 8, "parent": 7, "face_id": 6},
  {"type": "15", "id": 9, "parent": 8, "face_id": 0},
  {"type": "22", "id": 10, "parent": 8, "face_id": 4},
  {"type": "63", "id": 11, "parent": 10, "face_id": 3},
  {"type": "30", "id": 12, "parent": 11, "face_id": 0},
  {"type": "36", "id": 13, "parent": 12, "face_id": 0}
]
```
\end{lstlisting}

\subsection{Inspector}
\begin{lstlisting}
system:
<System Prompt>

inspector input:
Task Introduction

Construct a machine that can throw boulders in a parabolic trajectory, designed to maximize throw distance and height by utilizing the principle of leverage.

JSON Information
<JSON>

3D Information
<3D Information>

Questions
<Inspector Questions>

inspector-output:
<Three-Dimensional Perception of the World>
{
    "Coordinate System": "Left-Handed System",
    "Up": "y",
    "Right": "x",
    "Forward": "z",
    "Frontmost Block": {"3": [0.0, 0.0, 1.5]},
    "Rearmost Block": {"9": [2.0, 2.0, -3.0]},
    "Leftmost Block": {"5": [-2.0, 0.0, 1.5]},
    "Rightmost Block": {"7": [2.0, 1.5, 0.0]},
    "Topmost Block": {"13": [1.0, 7.4, -2.0]},
    "Lowest Block": {"0": [0, 0, 0]}
}
</Three-Dimensional Perception of the World>

<Question Answer>
1.  **Dynamic Blocks Analysis:**
    The only dynamic block in this machine is the `Rotating Block` [10].
    *   **Position and Orientation**: The `Rotating Block` [10] is located at center coordinates [1.0, 2.0, -2.0] with a global orientation of 'x-'.
    *   **Analysis**:
        a. **Impact on the machine**: The `Rotating Block` [10] serves as the pivot for the throwing arm. It is the core component intended to generate the throwing motion.
        b. **Direction of force**: The block is oriented along the 'x-' axis, meaning its rotation axis is the global x-axis. When activated, it will generate a torque, causing its sub-blocks to rotate in the y-z plane.
        c. **Impact on sub-blocks and machine**: The `Rotating Block` [10] will cause its child blocks ([11], [12], [13]) to swing upwards and forwards in a powerful arc. Simultaneously, it will exert an equal and opposite reaction torque on its parent block [8] and the entire support structure, which will likely cause the machine to violently tip over or break.

2.  **Special Static Blocks Analysis:**
    The special static blocks are the `Container` [12] and the `Boulder` [13].
    *   **Container [12]**: It is located at the end of the throwing arm (`Log` [11]) with its center at [1.0, 6.9, -2.0]. Its orientation is 'y+' (upwards). This placement and orientation are rational, as it correctly positions it to hold the `Boulder` [13] like a cup at the end of the throwing arm.
    *   **Boulder [13]**: It is placed inside the `Container` [12] with its center at [1.0, 7.4, -2.0]. As the projectile, its placement is logical and necessary for the machine's function.

3.  **Balance Check (self-gravity):**
    a. **Center of Gravity**: The calculated center of gravity for the entire machine is approximately at (0.85, 4.22, -1.46). The block closest to this point is the `Log` [11].
    b. **Sinking/Collapse**: Yes, parts of the machine will collapse. The entire throwing mechanism, including the heavy `Boulder` [13] (mass 5), rests on a tall, slender support tower made of single `Wooden Block`s ([6], [7], [8]). This structure is extremely fragile and not braced. The machine is severely top-heavy, and the immense weight concentrated on this single tower will cause it to bend and collapse under gravity alone, even before the `Rotating Block` [10] is activated.

4.  **Comprehensive Analysis:**
    a. **Force Summary**: The primary force is gravity, which makes the top-heavy machine highly unstable due to the high position of the `Boulder` [13]. Upon activation, the `Rotating Block` [10] will generate a massive torque on the throwing arm and an equal reaction torque on the flimsy support tower. The combination of static top-heaviness and dynamic reaction forces will lead to immediate structural failure.
    b. **Logically Unreasonable Blocks**:
        *   **Hierarchical Structure**: `Starting Block` [0] -> `Wooden Block` [6] -> `Wooden Block` [7] -> `Wooden Block` [8].
        *   **Reason for Unreasonableness**: The support tower consisting of blocks [6], [7], and [8] is fundamentally flawed. It is a single, unbraced column tasked with supporting the entire throwing arm and the heavy `Boulder` [13]. This design is far too weak to handle the static weight, let alone the powerful dynamic forces and reaction torque from the `Rotating Block` [10]. The machine is guaranteed to collapse.
</Question Answer>

<Summary of Design Defects>
1. Problem description, involving blocks: [6, 7, 8], belonging to structure: "Base and Support Frame"
   The support tower for the throwing mechanism, composed of blocks [6], [7], and [8], is critically weak. It is a single, unbraced column made of wooden blocks, which is completely inadequate to support the massive weight of the throwing arm and the `Boulder` [13]. The structure will collapse under its own weight or instantly shatter upon activation of the `Rotating Block` [10] due to the immense reaction torque.
</Summary of Design Defects>
\end{lstlisting}

\subsection{Environment Querier}
\begin{lstlisting}
system:
<System Prompt>

environment-querier input:
Task Introduction
Construct a machine that can throw boulders in a parabolic trajectory, designed to maximize throw distance and height by utilizing the principle of leverage.

Mechanical JSON Structure
<JSON>

3D Information
<3D Information>

Environment Feedback
machine damaged:
    machine parts
        DoubleWoodenBlock order_id:8 occurred at 0.4 sec block order_id 13 
boulder throwing distance 5.63
boulder max height 7.3
boulder actual position in first 5 seconds
[1.0, 7.01, -1.5], [0.84, 7.34, -2.44], [0.64, 5.69, -6.2], [0.61, 3.28, -9.97], [0.57, 1.68, -10.14], [0.38, 1.78, -9.45], [0.17, 1.72, -8.76], [-0.03, 1.52, -8.13], [-0.22, 1.22, -7.61], [-0.36, 0.85, -7.23], [-0.47, 0.48, -6.98], [-0.53, 0.16, -6.86], [-0.55, -0.07, -6.83], [-0.55, -0.18, -6.87], [-0.53, -0.17, -6.92], [-0.52, -0.03, -6.94], [-0.52, 0.21, -6.9], [-0.55, 0.51, -6.76], [-0.6, 0.85, -6.52], [-0.69, 1.18, -6.16], [-0.81, 1.45, -5.72], [-0.95, 1.66, -5.2], [-1.11, 1.76, -4.64], [-1.28, 1.76, -4.07], [-1.44, 1.66, -3.53]

environment-querier-output:
<Thought Process>
The user wants me to analyze a boulder-throwing machine that is failing its task.

**1. Analyze the Task and Machine Design:**
*   **Objective:** Build a leverage-based machine to throw a boulder for maximum distance and height. This implies a catapult or trebuchet design.
*   **Machine Structure:**
    *   **Base:** A wide base is constructed using `Wooden Block`s (ID 1, 2, 3, 4, 5, 6) connected to the `Starting Block` (ID 0). This seems designed for stability.
    *   **Tower:** A vertical tower is built on the right side of the base with `Wooden Block`s (ID 6, 7, 8). Block 8 is the apex of this support structure.
    *   **Lever Arm:** A `Rotating Block` (ID 10) is mounted on top of the tower (on block 8), serving as the fulcrum. A `Log` (ID 11) acts as the throwing arm, with a `Container` (ID 12) and `Boulder` (ID 13) at the top. A `Ballast` (ID 14) is attached to the other side of the `Rotating Block` as a counterweight.
    *   **Power System:** The machine uses a combination of a powered `Rotating Block` (ID 10) and two `Rubber Band`s (ID 15, 16). The rubber bands connect the top of the tower (block 8) to points on the base (blocks 3 and 5).

**2. Analyze the Environmental Feedback:**
*   **`machine damaged`**: This is the most critical piece of feedback. The machine is breaking during operation.
*   **`machine parts: DoubleWoodenBlock order_id:8`**: The failure point is identified as block 8. This is the `Wooden Block` that supports the entire rotating lever arm assembly.
*   **`occurred at 0.4sec`**: The failure is almost instantaneous, indicating a massive, immediate overload on the structure.
*   **Boulder Performance:** The boulder travels a mere 5.63 units and ends up behind its starting Z coordinate. This confirms a catastrophic failure of the launch mechanism, not just a weak throw. The arm is likely collapsing or swinging backward.

**3. Synthesize and Hypothesize the Root Cause:**
*   The point of failure (block 8) is subjected to several immense forces:
    1.  **Static Load:** The weight of the entire lever arm assembly (rotating block, log, container, boulder, ballast) rests on it.
    2.  **Dynamic Load:** The torque from the `Rotating Block` (ID 10) creates a powerful reaction force on its mount point (block 8).
    3.  **Tension Load:** The two `Rubber Band`s (ID 15, 16) are stretched from block 8 down to the base. When the simulation starts, they will contract with immense force, pulling down and forward on block 8.
*   **Hypothesis:** The combination of these forces, particularly the sudden, immense tension from the rubber bands, is causing block 8 to shear off from the rest of the tower (block 7). The tower itself, being a single column of wooden blocks, is structurally weak and cannot withstand this combination of bending and shear stress. The rubber bands are incorrectly placed; instead of powering the lever arm, they are destroying the support structure. The machine is literally tearing itself apart.

**4. Formulate a Plan for Further Investigation:**
To confirm this hypothesis, I need to observe the behavior of the key components involved in the structural failure around the time of the collapse (0.4s).

*   **Block 8 (Failure Point):** I need to see its position and rotation to confirm it's being ripped from its mounting.
*   **Block 10 (Rotating Block/Fulcrum):** Its movement will show the forces being applied by the lever arm onto block 8.
*   **Block 11 (Lever Arm):** Its motion will confirm the direction of the attempted throw and the overall collapse dynamics.
*   **Rubber Bands (15 & 16):** I need to check their `length` property. A rapid decrease in length will prove they are contracting and applying the destructive force I suspect.

Based on this analysis, I will request detailed feedback on these specific components to verify the cause of the failure.
</Thought Process>
<Required Feedback>
[
    {
        "id": 8,
        "duration": [0.0, 0.8],
        "properties": ["position", "rotation"]
    },
    {
        "id": 10,
        "duration": [0.0, 0.8],
        "properties": ["position", "rotation"]
    },
    {
        "id": 11,
        "duration": [0.0, 0.8],
        "properties": ["position", "rotation"]
    },
    {
        "id": 15,
        "duration": [0.0, 0.8],
        "properties": ["length"]
    },
    {
        "id": 16,
        "duration": [0.0, 0.8],
        "properties": ["length"]
    }
]
</Required Feedback>
\end{lstlisting}

\subsection{Refiner}
\begin{lstlisting}
system:
<System Prompt>

refiner input:
Task Introduction
Construct a machine that can throw boulders in a parabolic trajectory, designed to maximize throw distance and height by utilizing the principle of leverage.

Mechanical JSON Structure
<JSON>

3D Information
<3D Information>

Environment Feedback
machine damaged:
    machine parts
        DoubleWoodenBlock order_id:8 occurred at 0.4 sec block order_id 13 
boulder throwing distance 5.63
boulder max height 7.3
boulder actual position in first 5 seconds
[1.0, 7.01, -1.5], [0.84, 7.34, -2.44], [0.64, 5.69, -6.2], [0.61, 3.28, -9.97], [0.57, 1.68, -10.14], [0.38, 1.78, -9.45], [0.17, 1.72, -8.76], [-0.03, 1.52, -8.13], [-0.22, 1.22, -7.61], [-0.36, 0.85, -7.23], [-0.47, 0.48, -6.98], [-0.53, 0.16, -6.86], [-0.55, -0.07, -6.83], [-0.55, -0.18, -6.87], [-0.53, -0.17, -6.92], [-0.52, -0.03, -6.94], [-0.52, 0.21, -6.9], [-0.55, 0.51, -6.76], [-0.6, 0.85, -6.52], [-0.69, 1.18, -6.16], [-0.81, 1.45, -5.72], [-0.95, 1.66, -5.2], [-1.11, 1.76, -4.64], [-1.28, 1.76, -4.07], [-1.44, 1.66, -3.53]

Querier required feedback
block order_id 8 
    type_id 
        1 
    info duration(s) 
        [0.0, 1.0]
    block broken before below time(s) 
        0.2
    block position 
        [[2.01, 2.49, 0.01]] Due to block broken, the position is incomplete or not recorded
    block velocity
        [[0.15, 1.7, 4.1]] Due to block broken, the velocity is incomplete or not recorded
block order_id 10
    block type_id
        22
    block info duration(s)
        [0.0, 1.0]
    block position
        [[ 1.5 ,  2.52, -1.49], [ 1.5 ,  3.34, -0.56], [1.48, 5.27, 1.08], [0.73, 5.66, 2.42], [0.11, 4.45, 3.64])]
    block rotation
        [[ 0.  , -0.71,  0.01,  0.71], [-0.16, -0.68,  0.18,  0.69], [-0.47, -0.47,  0.51,  0.54], [-0.73, -0.04,  0.08,  0.67], [-0.69,  0.06, -0.04,  0.72]]
block order_id 0
    block type_id 0
    block info duration(s)
        [0.0, 2.0]
    block rotation
        [[0., 0., 0., 1.], [ 0.07,  0.  , -0.01,  1.  ], [ 0.45,  0.06, -0.04,  0.89], [ 0.39,  0.14, -0.09,  0.91], [0.  , 0.15, 0.03, 0.99], [-0.09,  0.05,  0.36,  0.93], [0.2 , 0.01, 0.09, 0.98], [ 0.13, -0.31, -0.13,  0.93], [-0.06, -0.26, -0.1 ,  0.96], [-0.08, -0.18, -0.1 ,  0.97]]

Initial State of the Machine
The machine is initially placed on the ground, facing the z+ direction, with the target direction being z+.
Please remember that the task failure is due to the mechanical design defects. 
Your ultimate goal is to modify the mechanical design defects so that it can complete the task.
The historical modification information represents your previous attempts to make changes, and these steps did not comply with the game rules.
Based on the information about the task failure, please make modifications to the mechanics. 
When designing the modification steps, avoid making the same mistakes as those in the historical modification information.

refiner-output:
<Thought Process>
The user wants me to fix a trebuchet-like machine that is failing.

**1. Analysis of the Machine and Failure:**
*   **Machine Structure:** The machine consists of a base, a vertical tower on the right side, and a rotating arm mounted on top of the tower. The arm has a container with a boulder on one end and was intended to have a counterweight on the other. Power is provided by a `Rotating Block` (ID 10) at the pivot and a `Ballast` (ID 14) acting as a counterweight. Two `Rubber Band`s (ID 15, 16) are connected from the top of the tower (ID 8) to the base, seemingly for stabilization.
*   **Failure Mode:** The simulation feedback consistently reports that `DoubleWoodenBlock id:8` is damaged at 0.4 seconds. Block 8 is the top-most block of the support tower, directly holding the entire rotating assembly (`Rotating Block`, arm, boulder, and ballast).
*   **Root Cause Analysis:**
    *   **Overwhelming Force:** The combination of the `Rotating Block`'s torque and the heavy `Ballast` (ID 14, mass 3) dropping creates an immense rotational force. The `Boulder` (ID 13, mass 5) at the end of a long lever arm (ID 11, `Log`) amplifies this stress. The tower, constructed from simple `Wooden Block`s, cannot withstand this force, leading to the failure of block 8.
    *   **Instability:** The feedback shows the entire machine lifting and tilting (`machine orientation in long time x-`, `machine frequently changes in the following orientations`). This indicates the base is too small and light to counteract the violent motion of the throwing arm. The `Rubber Band`s (ID 15, 16) are connected asymmetrically (from the tower to the front and left parts of the base), which likely introduces a twisting force, contributing to the instability and structural failure rather than preventing it.

**2. Modification Strategy:**
The primary goal is to ensure the machine's structural integrity so it can actually launch the boulder without self-destructing.

*   **Step 1: Reduce Destructive Forces.** The most critical issue is the excessive force breaking block 8. The simplest way to address this is to remove the primary source of this force, the `Ballast` (ID 14). The `Rotating Block` (ID 10) by itself will still provide sufficient power to launch the boulder, but with much less violent acceleration, which the wooden structure should be able to handle.
*   **Step 2: Remove Ineffective Components.** The `Rubber Band`s (ID 15 and 16) are not stabilizing the structure effectively and are likely contributing to the twisting failure due to their asymmetric placement. Removing them will simplify the forces acting on the tower and improve stability.

This plan directly targets the identified points of failure. By removing the ballast and the rubber bands, the stress on the critical component (block 8) is significantly reduced, and the machine's overall stability should improve. The throwing distance might be reduced, but a successful, non-destructive launch is the first priority.

**3. Final Plan:**
1.  Remove `Ballast` (ID 14).
2.  Remove `Rubber Band` (ID 15).
3.  Remove `Rubber Band` (ID 16).

This is a minimal but highly effective set of changes to fix the core problem.
</Thought Process>
<Modification Description>
The machine's support tower (block 8) is breaking due to excessive torque from the ballast and rotating block. The rubber bands are placed asymmetrically, causing twisting forces and instability. I will remove the ballast (ID 14) to reduce the strain on the structure and remove the two rubber bands (ID 15, 16) to eliminate the destabilizing twisting forces.
</Modification Description>
<Simulation Prediction After Modification>
*   **Power Blocks Analysis:** The launch will now be powered solely by the `Rotating Block` (10). The removal of the `Ballast` (14) significantly reduces the peak torque, resulting in a smoother, less violent rotation of the throwing arm. This will prevent the structural failure of the tower.
*   **Balance Check:** Without the heavy ballast dropping and the asymmetric pull from the rubber bands, the machine will be much more stable during operation. The base should remain mostly on the ground, with minimal tipping as the arm swings. The overall center of gravity will be more manageable.
*   **Comprehensive Analysis:** The primary defect, the destruction of block 8, will be resolved. The machine will successfully complete the launch sequence without breaking. While the throwing distance and height may be lower due to the reduced power, the machine will now be functional and can complete its primary task of throwing the boulder.
*   **10-second game simulation prediction:**
    *   **0.0s - 0.5s:** The `Rotating Block` activates, and the arm begins to swing upwards smoothly. The tower (block 8) remains intact.
    *   **0.5s - 1.0s:** The arm reaches its maximum rotation, and the boulder is released from the container. The machine's base remains stable on the ground.
    *   **1.0s - 5.0s:** The boulder travels in a clear parabolic arc. The machine settles without any damage. The final throwing distance will be shorter than the broken attempts but will represent a successful launch.
</Simulation Prediction After Modification>

<Modification Steps>
Remove [14]
Remove [15]
Remove [16]
</Modification Steps>
\end{lstlisting}

\newpage
\section{Single-Agent Prompt}
\label{app:single-agent-prompt}
\begin{lstlisting}
You are a machine builder. Your task is to generate a complete machine as a JSON file based on the user's request. Add new blocks to the initial structure; do not modify or delete it.

I. Rules:
1.  Coordinate System: Left-handed coordinate system, y+ upwards, z+ forward and x+ right.
2.  Block Placement: New blocks must attach to `attachable_faces` of existing blocks. Blocks cannot overlap.
3.  Size Limit: The final machine must not exceed dimensions of 17 (Length, Z), 17 (Width, X), 9.5 (Height, Y).
4.  Functionality: Ensure functional blocks are oriented correctly.
5.  Ground Interaction: The ground automatically conforms to the machine's lowest block. Account for potential collisions between the machine and the ground throughout operation.
6.  Gravity: Every block is subject to gravity; the greater a block's mass, the stronger its downward force. Consider this in your design when the machine is in operation.
7.  Physical rules: Classical mechanical laws such as conservation of momentum are applied.

II. Block Data:
Notes:
You can only use blocks from this list. A block's default orientation is Z+.
1. Attachable face:
   a. `id`: The i-th attachable_face of this block.
   b. `pos`: Coordinates relative to the building center(which is the attachable_face of the parent block) of this block.
   c. `orientation`: Orientation relative to the building center of this block.
2. Tags:
   a. `Non-static`: Block can generate force or movement.
   b. `Non-stable`: Connection to parent is not rigid (e.g., hinges, boulders).
   c. `Linear`: Do not collide with other blocks, but will occupy two attachable_faces.
3. Special Blocks:
   a. Boulder (id 36): Does not physically connect to other blocks.
   b. Spring (id 9): A special block that pulls its two connection points together.

Detailed Infos:
<Block Infos without explanations>

III. JSON Output Format:
1. type: block's type_id
2. id: this is i-th block
3. parent: parent block's id
4. face_id: parent block's constructible_point id
5. Standard Block: `{"type": <int>, "id": <int>, "parent": <int>, "face_id": <int>}`
6. special block (id: 9): `{"type": 9, "id": <int>, "parent_a": <int>, "face_id_a": <int>, "parent_b": <int>, "face_id_b": <int>}`

IV. Final Response Format:
Your response must contain only these two parts:
1. `Chain of thoughts:` 
    a. You need to think step by step, analyse each block's usage, and where to place them. Put your cot in <cot></cot>
    b. `Construction Idea:` A brief explanation of your design, remember to consider necessary block types, note them in ```necessary_blocks [type_1,type_2 ...]```, no more than 300 words.
2.  `JSON:` The complete JSON code inside a ```json ... ``` block. Here is an example: 
```json
    [
        {"type":"0","id":0,"parent":-1,"face_id":-1},
        {"type": <int>, "id": <int>, "parent": <int>, "face_id": <int>},
        ...
    ]
```
\end{lstlisting}

\newpage
\section{Multi-Agent Prompts}
\subsection{Shared Prompts}
\subsubsection{Game Introduction With 3D Knowledge}
\begin{lstlisting}
1. Coordinate System: The game uses a left-handed coordinate system, with the Y-axis pointing upwards. In global coordinates, z+ is forward and x+ is to the right.
2. Construction: New blocks must be connected to the "attachable faces" of existing blocks. The default orientation of blocks is z+.
3. Block Types:
    a. Regular Blocks: Have fixed dimensions and multiple attachable faces.
    b. special blocks (ID 7, 9): Connect two attachable faces, do not collide with other blocks, but will occupy the connection points.
4. Size Limitations: The mechanical dimensions must not exceed Length (z) 17, Width (x) 17  Height (y) 9.5.
\end{lstlisting}

\subsubsection{Machine 3D JSON Format}
\begin{lstlisting}
```json
[
    {"type":"0","id":0,"parent":-1,"face_id":-1},
    {"type":"Block Type ID","id":"Block Order ID","parent":"Parent Block ID","face_id":"Attachable Face ID in Parent Block"},
    ...
]
```
If it is a special block (Type ID is 7 or 9, other blocks are not special blocks), it will be:
```json
{
    "type":"Block Type ID",
    "id":"Block Order ID",
    "parent_a":"Parent A Block Order ID",
    "face_id_a":"Attachable Face ID in Parent A Block",
    "parent_b":"Parent B Block Order ID",
    "face_id_b":"Attachable Face ID in Parent B Block"
}
```
\end{lstlisting}

\subsubsection{Build Guidance}
\begin{lstlisting}
Your task is to: Add new blocks based on the initial machine JSON and construction requirements provided by the user, without deleting the initial structure, and output the final complete JSON.

User Input Format:
1.  Building Objective: Describe the structure and function to be built, and provide a list of recommended block IDs.
2.  Initial JSON: The structural data of the existing machine.

Core Building Rules:
1.  Block Usage: You can only select from the list of recommended block IDs provided by the user. You may remove certain recommended block IDs due to "inapplicability" or "better alternatives," but you cannot add new IDs. If any are removed, the reason must be stated.
2.  Collision Prevention: You must accurately calculate the coordinates and orientation of new blocks based on the orientation and position of the parent block to ensure that the new block does not overlap with the existing structure.
3.  Coordinate System and Orientation: The initial orientation of all blocks is Z+. The final orientation of new blocks must be transformed based on the parent block's orientation and the relative direction of the building point, according to the following rules:
       Oriented z+: Front z+, Back z-, Left x-, Right x+, Up y+, Down y-
       Oriented z-: Front z-, Back z+, Left x+, Right x-, Up y+, Down y-
       Oriented x-: Front x-, Back x+, Left z-, Right z+, Up y+, Down y-
       Oriented x+: Front x+, Back x-, Left z+, Right z-, Up y+, Down y-
       Oriented y+: Front y+, Back y-, Left x-, Right x+, Up z-, Down z+
       Oriented y-: Front y-, Back y+, Left x-, Right x+, Up z+, Down z-

Your Output Format:
1.  Building Plan:
    `Generated [structure summary] to achieve [function]. Finally used blocks [ID1, ID2,...]. Removed [ID3,...] because [reason for removal].`
2.  Final JSON:
    ```json
    [
      // The complete JSON including both the initial structure and the new blocks
    ]
    ```    
\end{lstlisting}

\subsubsection{Meta Designer System Prompt}
\begin{lstlisting}
You are a mechanical designer, and your task is to design a machine in the game Besiege based on the user's requirements. Please gain a general understanding of the game based on the following information:

I. Game Introduction:
1. Besiege is a physics-based construction game developed using Unity. Players need to build various machines to complete different tasks.
2. Besiege only contains basic mechanics and physical laws, such as mass, friction, and collision.
3. Blocks are used to build machines. Each block has its unique functions, advantages, and disadvantages. 

II. Block Introduction:
1. Blocks are mainly divided into five major categories: Basic Blocks, Mobility Blocks, Mechanical Blocks, Weapon Blocks, and Armor Blocks.
- Basic Blocks are the fundamental components of many machines - structural blocks and some basic moving parts.
- Mobility Blocks are primarily designed for movement functions - powered and unpowered wheels, steering blocks, and gears.
- Mechanical Blocks provide various useful auxiliary functions - joints, suspension devices, winches, grabbers, etc.
- Weapon Blocks offer various types of violent output at different ranges - swords and saws for close combat, and cannons and rockets for long-range.
- Armor Blocks can protect the machine from damage or provide useful shapes for carrying other blocks - armor plates and wooden panels, as well as half-pipes and brackets.

2. Here is a detailed introduction to the properties and functions of each block:
| Name | Category | Type ID | Function |
|------|----------|---------|----------|
| Starting Block | Basic | 0 | The Starting Block is the root block of the machine; it is placed at the starting position by default, cannot be moved, cannot be deleted, and only one can exist at a time. |
| Small Wooden Block | Basic | 15 | A basic structural block, cube-shaped, to which other blocks can be attached from any side, making it particularly suitable for constructing the basic framework of machines. |
| Wooden Block | Basic | 1 | A basic mechanical block, twice the length of a Small Wooden Block. |
| Wooden Rod | Basic | 41 | A basic mechanical block, twice the length of a Small Wooden Block, with the same weight as a Wooden Block, but very fragile. |
| Log | Basic | 63 | A basic mechanical block, more robust, three times the length of a Small Wooden Block. |
| Brace | Basic | 7 | A non-placeable block used for reinforcement, built by "attaching" to other blocks, with no collision volume. It is often used to increase the stability of static structures and is not suitable for any dynamic structures. |
| Steering Hinge | Mobility | 28 | The Steering Hinge can rotate blocks along an axis perpendicular to the placement axis. This block can rotate child blocks to a 180-degree direction to the left or right, commonly used for vehicle steering. |
| Steering Block | Mobility | 13 | The Steering Block can rotate blocks along its placement axis, similar to the rotating part of a helicopter's rotor. |
| Powered Wheel | Mobility | 2 | Similar to a car wheel, it can drive itself but cannot turn independently. It is a mechanical device used for moving objects on the ground. |
| Unpowered Wheel | Mobility | 40 | A wheel that does not rotate without external force, otherwise similar to a Powered Wheel. |
| Powered Large Wheel | Mobility | 46 | Similar to a Powered Wheel, but with a radius and thickness twice that of a Powered Wheel. |
| Unpowered Large Wheel | Mobility |  | A wheel that does not rotate without external force, otherwise similar to a Powered Large Wheel. |
| Small Wheel | Mobility | 50 | It works almost the same as a caster wheel (like a shopping cart wheel), unpowered. |
| Universal Joint | Mechanical | 19 | A block that can freely rotate around its placement axis, similar to a Steering Block but without power. |
| Hinge | Mechanical | 5 | Similar to a Steering Hinge, but without power. |
| Ball Joint | Mechanical | 44 | Can swing 360 degrees along the axis perpendicular to the placement axis, but without power. |
| Axle Connector | Mechanical | 76 | Similar to a Ball Joint. |
| Rotating Block | Mechanical | 22 | Powered, it can rotate clockwise or counterclockwise along the axis perpendicular to the placement axis. |
| Suspension | Mechanical | 16 | Shaped like a wooden block, it can buffer forces from all directions. |
| Grabber | Mechanical | 27 | It will grab and hold onto any object it comes into contact with. |
| Spring | Mechanical | 9 | A special block that attaches to two other blocks and can quickly pull them together. Its pulling force is almost entirely dependent on its length. |
| Boulder | Weapon | 36 | A stone that does not directly connect to other blocks even when built on them. It can be used as a projectile weapon and is also commonly used as a target in transportation tests. |
| Elastic Pad | Armor | 87 | Increases the elasticity of the contact surface, providing an effect of rebounding and increasing kinetic energy. |
| Container | Armor | 30 | Can hold child blocks like a bowl, mainly used to carry blocks that cannot be directly connected to the machine. The container has some anti-slip capability, and only one block (the target to be carried) can be placed inside. No other blocks can be added. |
| Roller Wheel | Locomotion | 86 | Similar to the small wheel, but shorter (0.8m). |
|  Grip Pad | Armour | 49 | Block with the highest friction. |
|  Ballast | Flight | 35 | A heavy cubic block used as a counterweight. |

III. Mechanical Design Requirements:
1. When designing the machine, you should adopt a "layered design" approach. Break down the user's requirements into the functions that the machine needs to achieve, and list the functional points.
2. For each functional point, design a structure that can meet the function. A structure can be understood as a "group of blocks," and several structures combined form the machine.
3. For each structure, determine the types of blocks to be used.
4. Determine the construction order of the structures to make the machine-building process layered. List which structure is the foundation and which is the upper-layer structure, and establish the construction sequence chain.

IV. Output Format Requirements:
```json
{
    "definition": "Construct a machine that can fulfill the user's requirements",
    "function_points": ["Function Point 1", "Function Point 2", "Function Point 3"],
    "design_structure": [
        {
            "function_name": "Structure 1 Name",
            "description": "Description of Structure 1",
            "related_function_points": ["Function Point 1", "Function Point 2"]
        },
        {
            "function_name": "Structure 2 Name",
            "description": "Description of Structure 2",
            "related_function_points": ["Function Point 3"]
        }
    ],
    "build_order": ["Structure 2 Name", "Structure 1 Name"],
    "machine_structure": {
        "Structure 1 Name": {
            "block_id": [ID1, ID2, ID3...],
            "guidance": "Guidance here"
        },
        "Structure 2 Name": {
            "block_id": [ID4, ID5, ID6...],
            "guidance": "Guidance here"
        }
    }
}
```
V. Note:
1. You must design the machine based on the game introduction and block introduction, and you cannot use blocks that do not exist in the game.
2. Strictly follow the output format requirements. Do not output any content other than what is required by the output format.
3. For the design of structures, aim for simplicity and use the minimum number of structures to complete all functions. Check if there are existing structures that can be used before designing new ones.
4. When selecting blocks for a structure, limit the types of blocks to no more than three, and preferably use only one type. Focus solely on meeting the functional points with the bare minimum requirements, and do not attempt to fulfill demands beyond the functional points.

I will provide the user input below. Please generate a mechanical overview in JSON format based on the user's description.
\end{lstlisting}

\subsection{Designer System And User Prompt}
\begin{lstlisting}
<system>
You are a mechanical builder in the game "Besiege." 
Your task is to add new blocks to an existing machine structure according to user requests and finally output the complete machine JSON data.
I. Game Introduction:
<Game Introduction With 3D Knowledge>
II. Introduction to Blocks:
<Block Infos>
III. Introduction to JSON Format:
<Machine 3D JSON Format>
IV. Construction Guidance:
<Build Guidance>
V. Output Format Requirements:
Based on the existing structure, a [structural summary] was generated as [functional implementation]. 
Ultimately, the block types [ID1, ID2, ...] were decided upon, while [ID3, ...] were removed due to [reason for removal].
JSON:
```json
...
```
VI. Note:
Building Principles
1. Correct Orientation: Ensure that functional blocks such as wheels and hinges are oriented correctly to achieve the intended function.
2.  Efficiency First:
    a. Complete the design goal with the fewest blocks possible.
    b. The ground will automatically adapt to the lowest point of the mechanism.

Output Requirements
1.  Strict Structure: Your response must only contain two parts: Construction Plan and Final JSON. Prohibit any additional greetings, explanations, or comments.
2.  Pure JSON: The complete JSON code block must be placed within ```json ... ```. Prohibit modifying the initial structure. Prohibit modifying the `scale` property of any block. Prohibit adding comments or non-existent properties in the JSON.

Next, I will provide user input. Please generate a JSON based on the description.
<user>
<designer_output["design_structure"][i]["description"]>+<Output Machine Json>
\end{lstlisting}

\subsection{Inspector System And User Prompt}
\begin{lstlisting}
<system>
I'll provide you with a mission in the game Besiege, 
along with the machine designed for it in JSON format and its 3D information. 
Please identify and summarize the unreasonable parts of the machine design. 
Here's the introduction to the game and construction knowledge.
I. Game Introduction:
<Game Introduction With 3D Knowledge>
II. Introduction to Blocks:
<Block Infos>
III. Introduction to JSON and 3D Information:
<Machine 3D JSON Format>
IV. Introduction to Output Format:
<Three-Dimensional Perception of the World>
{
    "Coordinate System": "Left-Handed System or Right-Handed System",
    "Up": "x or y or z",
    "Right": "x or y or z",
    "Forward": "x or y or z",
    "Frontmost Block": {"id": [Center Point Coordinates]},
    "Rearmost Block": {"id": [Center Point Coordinates]},
    "Leftmost Block": {"id": [Center Point Coordinates]},
    "Rightmost Block": {"id": [Center Point Coordinates]},
    "Topmost Block": {"id": [Center Point Coordinates]},
    "Lowest Block": {"id": [Center Point Coordinates]},
}
</Three-Dimensional Perception of the World>

<Question Answer>
Write the answer to the user's question here
</Question Answer>

<Summary of Design Defects>
1. Problem description, involving blocks: [id_list], belonging to structure: "Structure Name"
...
</Summary of Design Defects>
V. Notes:
1. Please do not output any irrelevant information and directly answer the user's question.
2. The id of the block must be enclosed in "[]". Additionally, do not use "[]" for any other numbers; consider using "()" instead.

Below, I will provide you with JSON and 3D information. Please answer the user's question based on this information.

<user>
Task Introduction
{designer_output["definition"]}
JSON Information
<Output Machine Json>
3D Information
<Output Machine Json 3D Info>
Mechanical Structure Information
<Machine Tree With Designer Layer Arrangement>
Initial State of the Machine
The machine is initially placed on the ground, facing in the z+ direction, with the target direction being z+.

Questions:
1. Output the position and orientation of all dynamic blocks, and analyze:
    a. The impact of dynamic blocks on the machine
    b. The direction of force provided by dynamic blocks
    c. The impact on sub-blocks and the direction of force on the machine

2. Output static blocks other than basic structural blocks, and analyze the rationality of their orientation and position.

3. Balance Check (self-gravity)
    a. The center of gravity of the machine (find the block closest to the center of gravity)
    b. Whether parts of the machine will sink due to gravity
4. Comprehensive Analysis
    a. Summarize the direction of all forces to analyze the movement of the machine
    b. Identify logically unreasonable blocks, output their hierarchical structure and reasons for unreasonableness
\end{lstlisting}

\subsection{Refiner System Prompt}
\begin{lstlisting}
I will give you a task in the game Besiege, as well as the 3D information of the machine designed to complete this task. There are some unreasonable aspects in the design of this machine, and I would like you to modify these parts:
I. Game Introduction:
<Game Introduction With 3D Knowledge>
II. Block Introduction:
<Block Infos>
III. Input Introduction:
1. Task & Context
Task Objective:
   <designer_output["user_input"]>
Preceding Information (if any):
   Defect Report:
       <quizzer_output>
   Modification History:
       <modify_history>
   Environmental Feedback:
       <environment_feedback>
       
2. Machine Data
<Machine 3D JSON Format>

IV. Modification Method Introduction:
Please follow the steps below to analyze and modify the machine structure.
Step 1: Analyze & Plan
1.  Diagnose the Current Machine: Analyze its power, balance, structure, and overall movement to identify design flaws or areas for optimization.
2.  Devise a Modification Plan: Based on your diagnosis, decide which blocks to move, remove, or add.
3.  Evaluate Modification Impact: When planning, you must consider the impact of your changes.
4.  Briefly Describe Modifications: Before generating commands, describe your modification plan in a sentence or two using natural language.

Step 2: Execute Modification Operations
Use the following three command formats for modifications. Output only the operation commands, with each command on a new line.

1.  Add Block (Add)
   Format: Non-Linear: Add [block type ID] to [id] in [attachable_face_id]
            Linear: Add [block type ID] to [id_a] in [attachable_face_id_a] to [id_b] in [attachable_face_id_b]
   Rules:
       Can only be added to original blocks. You cannot add new blocks onto other newly added blocks.
       special blocks require specifying two connection points.

2.  Remove Block (Remove)
   Format: Remove [id]
   Rules:
       Can only remove original blocks.
       Cannot remove a block that has child blocks.

3.  Move Block (Move)
   Move [id] to [new_parent_id] in [new_attachable_face_id]
   Rules:
       Moves the target block and all its child blocks as a single unit.
       The new parent's `id` must be smaller than the `id` of the block being moved.
       special blocks cannot be moved.
       The move must change the block's original position.
       
<Build Guidance: Coordinate System and Orientation>

V. Output Format Introduction:
<Thought Process>
</Thought Process>
<Modification Description>
</Modification Description>
<Simulation Prediction After Modification>
</Simulation Prediction After Modification>
<Required Feedback>
</Required Feedback>
<Modification Steps>
</Modification Steps>
VI. Note:
1. Output Scope: Only output content related to the modification method and the required output format.
2. Ground Definition: The ground is always located beneath the machine, in contact with the block that has the lowest y-coordinate.
3. Task Adherence: Pay close attention to the task requirements. Do not delete important blocks by mistake.
4. Parenting Constraint: A block that has been deleted cannot be used as a parent for new or moved blocks within the same set of operations.
5. Format Integrity: Ensure the output format is complete. You must retain the `<Thought Process></Thought Process>` and `<Modification Description></Modification Description>` tags.
6. Content Separation: Do not include verification results within the `<Modification Description>` block.
7. Preserve 'Success' Steps: Retain any steps marked as "Success" and do not adjust their order.
8. Prioritize 'Error' Steps: Focus your efforts on fixing the steps marked as "Error". Do not modify steps marked as "Unverified" prematurely.
9. Error History: I will provide the modification history for the "Error" steps. Use this information to avoid repeating invalid operations.

Below, I will provide you with the JSON and 3D information. Please modify the machine accordingly.
\end{lstlisting}

\subsection{Environment Querier System Prompt}
\begin{lstlisting}
I will give you a task in the game Besiege, as well as the information on the machine designed to complete this task. 
The machine has finished the task simulation and returned some environmental feedback to describe its performance.
Please analyze the issues with the machine based on the feedback and request more feedback if needed.
I. Game Introduction:
<Game Introduction With 3D Knowledge>
II. Block Introduction:
<Block Infos>
III. Input Introduction:
<Machine 3D JSON Format>
IV. Query Introduction:
A. Environmental Feedback Request:
After conducting the environmental simulation of your modified machinery, 
The system will provide essential data for the most critical components (such as position, rotation, and velocity). 
Based on the performance of the essential data, you need to determine what problems the machinery may have encountered and request feedback on the key blocks that may have issues.
You can also request those blocks that may significantly impact the machinery's functionality and check whether their performance meets expectations.
The format for the environmental feedback request is as follows. 
Please adhere strictly to this format and avoid including any extraneous information:
<Required Feedback>
[
    {
        "id": int,
        "duration": [float, float],
        "properties": ["position", "rotation", "velocity", "length"],
    },
    ...
]
</Required Feedback>
Both "id" and "duration" are mandatory fields. 
Note that the game runs for a total of 5 seconds, game state samples per 0.2s; do not exceed this duration. 
You can freely select the attributes in "properties," but avoid including irrelevant information. 
The "length" attribute is only applicable to linear components.
V. Output Format Introduction:
<Thought Process>
</Thought Process>
<Required Feedback>
</Required Feedback>
VI. Note:
1. Please do not output any irrelevant information.
2. The ground will always correctly appear beneath the machine, making normal contact with the block that has the lowest y-axis coordinate.
3. All blocks are affected by gravity. If a block with a large self-weight is built on certain non-powered non-static blocks, the non-static blocks may rotate due to the gravitational force of the sub-blocks.
4. Similarly, the power generated by powered blocks also needs to counteract the gravitational force of the sub-blocks.
5. Please adhere to the output format and avoid adding any irrelevant information in the JSON.

Below, I will provide you with the JSON and 3D information, as well as the environmental feedback. Please request more feedback as needed.
\end{lstlisting}

\subsection{Block Informations}
\begin{lstlisting}
Explanations:

This is a concise list of the blocks you can use for this construction. Please read and follow the rules carefully.
I. Block Information Format
Each block's information follows the dict format. Attributes that a block does not have will not appear in the keys.
1. Characteristic Tags:
   a. Non-static: The block can actively generate force or movement.
   b. Non-stable: The connection between the block and its parent block is non-rigid, allowing for movement or rotation.
   c. Linear: The block is used to connect two existing points rather than being attached to a single point.
   If there are no tags, it is a regular static and stable block.
2. Attachable Faces:
   The key is in the format of attachable face ID, coordinates (relative coordinates), and orientation (relative orientation).
II. Key Special Rules
1. Powered Wheel Rule (applicable to all powered wheels): The direction of power provided by the wheel is not the same as its orientation.
   - Forward (Z+ direction): The wheel should face sideways (X+ or X-).
   - Left turn (power pushes towards X-): The wheel should face forward (Z+).
   - Right turn (power pushes towards X+): The wheel should face backward (Z-).
   - When the wheel faces up or down (Y+ or Y-), it does not provide power.
2. Special Blocks (Brace, Spring):
   - They do not have their own connection points but connect to two other blocks.
   - Brace: Used to reinforce static structures with no collision volume.
   - Spring: Generates a contracting pull force along the line connecting its two ends when stretched.
3. Non-connecting Blocks (bombs, boulders):
   - These blocks are placed at the specified location but do not form a physical connection with any other block. Containers are usually needed to hold them.

Detailed Infos:
[
    {
        "Name": "Starting Block",
        "Description": "The root block of the mechanism. It cannot be placed or deleted, and only one can exist at a time. Its initial position is fixed, and its initial orientation is z+.",
        "Type ID": 0,
        "Size": [1, 1, 1],
        "Attachable Faces Properties": [
            {"ID": 0, "Coordinates": [0, 0, 0.5], "Orientation": "Front"},
            {"ID": 1, "Coordinates": [0, 0, -0.5], "Orientation": "Back"},
            {"ID": 2, "Coordinates": [-0.5, 0, 0], "Orientation": "Left"},
            {"ID": 3, "Coordinates": [0.5, 0, 0], "Orientation": "Right"},
            {"ID": 4, "Coordinates": [0, 0.5, 0], "Orientation": "Up"},
            {"ID": 5, "Coordinates": [0, -0.5, 0], "Orientation": "Down"}
        ],
        "Mass": 0.25
    },
    {
        "Name": "Small Wooden Block",
        "Description": "A basic construction block, cubic in shape.",
        "Type ID": 15,
        "Size": [1, 1, 1],
        "Attachable Faces Properties": [
            {"ID": 0, "Coordinates": [0, 0, 1], "Orientation": "Front"},
            {"ID": 1, "Coordinates": [-0.5, 0, 0.5], "Orientation": "Left"},
            {"ID": 2, "Coordinates": [0.5, 0, 0.5], "Orientation": "Right"},
            {"ID": 3, "Coordinates": [0, 0.5, 0.5], "Orientation": "Up"},
            {"ID": 4, "Coordinates": [0, -0.5, 0.5], "Orientation": "Down"}
        ],
        "Mass": 0.3
    },
    {
        "Name": "Wooden Block",
        "Description": "A basic construction block.",
        "Type ID": 1,
        "Size": [1, 1, 2],
        "Attachable Faces Properties": [
            {"ID": 0, "Coordinates": [0, 0, 2], "Orientation": "Front"},
            {"ID": 1, "Coordinates": [-0.5, 0, 0.5], "Orientation": "Left"},
            {"ID": 2, "Coordinates": [-0.5, 0, 1.5], "Orientation": "Left"},
            {"ID": 3, "Coordinates": [0.5, 0, 0.5], "Orientation": "Right"},
            {"ID": 4, "Coordinates": [0.5, 0, 1.5], "Orientation": "Right"},
            {"ID": 5, "Coordinates": [0, 0.5, 0.5], "Orientation": "Up"},
            {"ID": 6, "Coordinates": [0, 0.5, 1.5], "Orientation": "Up"},
            {"ID": 7, "Coordinates": [0, -0.5, 0.5], "Orientation": "Down"},
            {"ID": 8, "Coordinates": [0, -0.5, 1.5], "Orientation": "Down"}
        ],
        "Mass": 0.5
    },
    {
        "Name": "Wooden Rod",
        "Description": "A basic construction block, slender and fragile.",
        "Type ID": 41,
        "Size": [1, 1, 2],
        "Attachable Faces Properties": [
            {"ID": 0, "Coordinates": [0, 0, 2], "Orientation": "Front"},
            {"ID": 1, "Coordinates": [-0.5, 0, 0.5], "Orientation": "Left"},
            {"ID": 2, "Coordinates": [-0.5, 0, 1.5], "Orientation": "Left"},
            {"ID": 3, "Coordinates": [0.5, 0, 0.5], "Orientation": "Right"},
            {"ID": 4, "Coordinates": [0.5, 0, 1.5], "Orientation": "Right"},
            {"ID": 5, "Coordinates": [0, 0.5, 0.5], "Orientation": "Up"},
            {"ID": 6, "Coordinates": [0, 0.5, 1.5], "Orientation": "Up"},
            {"ID": 7, "Coordinates": [0, -0.5, 0.5], "Orientation": "Down"},
            {"ID": 8, "Coordinates": [0, -0.5, 1.5], "Orientation": "Down"}
        ],
        "Mass": 0.5
    },
    {
        "Name": "Log",
        "Description": "A basic construction block.",
        "Type ID": 63,
        "Size": [1, 1, 3],
        "Attachable Faces Properties": [
            {"ID": 0, "Coordinates": [0, 0, 3], "Orientation": "Front"},
            {"ID": 1, "Coordinates": [-0.5, 0, 0.5], "Orientation": "Left"},
            {"ID": 2, "Coordinates": [-0.5, 0, 1.5], "Orientation": "Left"},
            {"ID": 3, "Coordinates": [-0.5, 0, 2.5], "Orientation": "Left"},
            {"ID": 4, "Coordinates": [0.5, 0, 0.5], "Orientation": "Right"},
            {"ID": 5, "Coordinates": [0.5, 0, 1.5], "Orientation": "Right"},
            {"ID": 6, "Coordinates": [0.5, 0, 2.5], "Orientation": "Right"},
            {"ID": 7, "Coordinates": [0, 0.5, 0.5], "Orientation": "Up"},
            {"ID": 8, "Coordinates": [0, 0.5, 1.5], "Orientation": "Up"},
            {"ID": 9, "Coordinates": [0, 0.5, 2.5], "Orientation": "Up"},
            {"ID": 10, "Coordinates": [0, -0.5, 0.5], "Orientation": "Down"},
            {"ID": 11, "Coordinates": [0, -0.5, 1.5], "Orientation": "Down"},
            {"ID": 12, "Coordinates": [0, -0.5, 2.5], "Orientation": "Down"}
        ],
        "Mass": 1
    },
    {
        "Name": "Steering Hinge",
        "Description": "Powered, used to control the rotation of sub-blocks. It can swing left and right along the axis perpendicular to the placement axis.",
        "Type ID": 28,
        "Size": [1, 1, 1],
        "Attachable Faces Properties": [
            {"ID": 0, "Coordinates": [0, 0, 1], "Orientation": "Front"}
        ],
        "Special Attributes": {
            "Swing Direction": ["Left", "Right"],
            "Angle": [-90, 90],
            "NonStatic":"True",
            "NonStable":"True"
        },
        "Mass": 1
    },
    {
        "Name": "Steering Block",
        "Description": "Powered, used to control the rotation of sub-blocks. It can rotate clockwise or counterclockwise along the placement axis.",
        "Type ID": 13,
        "Size": [1, 1, 1],
        "Attachable Faces Properties": [
            {"ID": 0, "Coordinates": [0, 0, 1], "Orientation": "Front"},
            {"ID": 1, "Coordinates": [-0.5, 0, 0.5], "Orientation": "Left"},
            {"ID": 2, "Coordinates": [0.5, 0, 0.5], "Orientation": "Right"},
            {"ID": 3, "Coordinates": [0, 0.5, 0.5], "Orientation": "Up"},
            {"ID": 4, "Coordinates": [0, -0.5, 0.5], "Orientation": "Down"}
        ],
        "Special Attributes": {
            "Rotation Axis": "Front",
            "NonStatic":"True",
            "NonStable":"True"
        },
        "Mass": 1
    },
    {
        "Name": "Powered Wheel",
        "Description": "Powered, a mechanical device used to move objects on the ground.",
        "Type ID": 2,
        "Size": [2, 2, 0.5],
        "Attachable Faces Properties": [
            {"ID": 0, "Coordinates": [0, 0, 0.5], "Orientation": "Front"}
        ],
        "Special Attributes": {
            "Rotation Axis": "Front",
            "PoweredWheel":"True",
            "NonStatic":"True",
            "NonStable":"True"
        },
        "Mass": 1
    },
    {
        "Name": "Unpowered Wheel",
        "Description": "A wheel that does not rotate without external force, similar to the powered wheel.",
        "Type ID": 40,
        "Size": [2, 2, 0.5],
        "Attachable Faces Properties": [
            {"ID": 0, "Coordinates": [0, 0, 0.5], "Orientation": "Front"}
        ],
        "Special Attributes": {
            "Rotation Axis": "Front",
            "NonStable":"True"
        },
        "Mass": 1
    },
    {
        "Name": "Large Powered Wheel",
        "Description": "Similar to the powered wheel, but larger.",
        "Type ID": 46,
        "Size": [3, 3, 1],
        "Attachable Faces Properties": [
            {"ID": 0, "Coordinates": [0, 0, 1], "Orientation": "Front"},
            {"ID": 1, "Coordinates": [-1.5, 0, 1], "Orientation": "Front"},
            {"ID": 2, "Coordinates": [1.5, 0, 1], "Orientation": "Front"},
            {"ID": 3, "Coordinates": [0, 1.5, 1], "Orientation": "Front"},
            {"ID": 4, "Coordinates": [0, -1.5, 1], "Orientation": "Front"},
            {"ID": 5, "Coordinates": [-1.5, 0, 0.5], "Orientation": "Left"},
            {"ID": 6, "Coordinates": [1.5, 0, 0.5], "Orientation": "Right"},
            {"ID": 7, "Coordinates": [0, 1.5, 0.5], "Orientation": "Up"},
            {"ID": 8, "Coordinates": [0, -1.5, 0.5], "Orientation": "Down"}
        ],
        "Special Attributes": {
            "Rotation Axis": "Front",
            "PoweredWheel":"True",
            "NonStatic":"True",
            "NonStable":"True"
        },
        "Mass": 1
    },
    {
        "Name": "Large Unpowered Wheel",
        "Description": "Similar to the unpowered wheel, but larger.",
        "Type ID": 60,
        "Size": [3, 3, 1],
        "Attachable Faces Properties": [
            {"ID": 0, "Coordinates": [0, 0, 1], "Orientation": "Front"},
            {"ID": 1, "Coordinates": [-1.5, 0, 1], "Orientation": "Front"},
            {"ID": 2, "Coordinates": [1.5, 0, 1], "Orientation": "Front"},
            {"ID": 3, "Coordinates": [0, 1.5, 1], "Orientation": "Front"},
            {"ID": 4, "Coordinates": [0, -1.5, 1], "Orientation": "Front"},
            {"ID": 5, "Coordinates": [-1.5, 0, 0.5], "Orientation": "Left"},
            {"ID": 6, "Coordinates": [1.5, 0, 0.5], "Orientation": "Right"},
            {"ID": 7, "Coordinates": [0, 1.5, 0.5], "Orientation": "Up"},
            {"ID": 8, "Coordinates": [0, -1.5, 0.5], "Orientation": "Down"}
        ],
        "Special Attributes": {
            "Rotation Axis": "Front",
            "NonStable":"True"
        },
        "Mass": 1
    },
    {
        "Name": "Small Wheel",
        "Description": "It works almost the same as a caster wheel (e.g., shopping cart wheel), but it is not powered.",
        "Type ID": 50,
        "Size": [0.5, 1, 1.5],
        "Special Attributes": {"NonStable":"True"},
        "Mass": 0.5
    },
    {
        "Name": "Roller Wheel",
        "Description": "Same as the small wheel.",
        "Type ID": 86,
        "Size": [1, 1, 1],
        "Special Attributes": {
            "NonStable":"True"
        },
        "Mass": 0.5
    },
    {
        "Name": "Universal Joint",
        "Description": "A block that can freely rotate around its placement axis, but it is not powered.",
        "Type ID": 19,
        "Size": [1, 1, 1],
        "Attachable Faces Properties": [
            {"ID": 0, "Coordinates": [0, 0, 1], "Orientation": "Front"},
            {"ID": 1, "Coordinates": [-0.5, 0, 0.5], "Orientation": "Left"},
            {"ID": 2, "Coordinates": [0.5, 0, 0.5], "Orientation": "Right"},
            {"ID": 3, "Coordinates": [0, 0.5, 0.5], "Orientation": "Up"},
            {"ID": 4, "Coordinates": [0, -0.5, 0.5], "Orientation": "Down"}
        ],
        "Special Attributes": {
            "Rotation Axis": "Front",
            "NonStable":"True"
        },
        "Mass": 0.5
    },
    {
        "Name": "Hinge",
        "Description": "It can swing up and down along the axis perpendicular to the placement axis, but it is not powered.",
        "Type ID": 5,
        "Size": [1, 1, 1],
        "Attachable Faces Properties": [
            {"ID": 0, "Coordinates": [0, 0, 1], "Orientation": "Front"},
            {"ID": 1, "Coordinates": [-0.5, 0, 0.5], "Orientation": "Left"},
            {"ID": 2, "Coordinates": [0.5, 0, 0.5], "Orientation": "Right"},
            {"ID": 3, "Coordinates": [0, 0.5, 0.5], "Orientation": "Up"},
            {"ID": 4, "Coordinates": [0, -0.5, 0.5], "Orientation": "Down"}
        ],
        "Special Attributes": {
            "Swing Direction": ["Up", "Down"],
            "Angle": [-90, 90],
            "NonStable":"True"
        },
        "Mass": 0.5
    },
    {
        "Name": "Ball Joint",
        "Description": "It can swing freely in all directions, but it is not powered.",
        "Type ID": 44,
        "Size": [1, 1, 1],
        "Attachable Faces Properties": [
            {"ID": 0, "Coordinates": [0, 0, 1], "Orientation": "Front"},
            {"ID": 1, "Coordinates": [-0.5, 0, 0.5], "Orientation": "Left"},
            {"ID": 2, "Coordinates": [0.5, 0, 0.5], "Orientation": "Right"},
            {"ID": 3, "Coordinates": [0, 0.5, 0.5], "Orientation": "Up"},
            {"ID": 4, "Coordinates": [0, -0.5, 0.5], "Orientation": "Down"}
        ],
        "Special Attributes": {
            "Swing Range": "All directions outward from the build surface",
            "NonStable":"True"
        },
        "Mass": 0.5
    },
    {
        "Name": "Axle Connector",
        "Description": "Similar to a ball joint.",
        "Type ID": 76,
        "Size": [1, 1, 1],
        "Attachable Faces Properties": [
            {"ID": 0, "Coordinates": [0, 0, 1], "Orientation": "Front"}
        ],
        "Special Attributes": {
            "Swing Range": "All directions outward from the build surface",
            "NonStable":"True"
        },
        "Mass": 0.3
    },
    {
        "Name": "Rotating Block",
        "Description": "When powered, this motor-like block generates torque and rotates about its local y-axis. Blocks connected at attachable_face 1 or 4 rotate with it as part of a rigid assembly. The rotation block has its own mass and obeys classical mechanics: it applies torque to connected parts when powered, and it can also be moved, rotated, or stopped by external forces or torques, depending on constraints.",
        "Type ID": 22,
        "Size": [1, 1, 1],
        "Attachable Faces Properties": [
            {"ID": 0, "Coordinates": [0, 0, 1], "Orientation": "Front"},
            {"ID": 1, "Coordinates": [-0.5, 0, 0.5], "Orientation": "Left"},
            {"ID": 2, "Coordinates": [0.5, 0, 0.5], "Orientation": "Right"},
            {"ID": 3, "Coordinates": [0, 0.5, 0.5], "Orientation": "Up"},
            {"ID": 4, "Coordinates": [0, -0.5, 0.5], "Orientation": "Down"}
        ],
        "Special Attributes": {
            "Rotation Axis": "Front",
            "NonStatic":"True",
            "NonStable":"True"
        },
        "Mass": 1
    },
    {
        "Name": "Grabber",
        "Description": "If the build point is unoccupied, it will grab any object that comes into contact with the build point and hold it firmly.",
        "Type ID": 27,
        "Size": [1, 1, 1],
        "Attachable Faces Properties": [
            {"ID": 0, "Coordinates": [0, 0, 1], "Orientation": "Front"}
        ],
        "Special Attributes": {
            "Grip Direction": "Front",
            "NonStable":"True"
        },
        "Mass": 0.5
    },
    {
        "Name": "Boulder",
        "Description": "A rock that will not directly connect to other blocks even if built on them, high mass.",
        "Type ID": 36,
        "Size": [1.9, 1.9, 1.9],
        "Special Attributes": {
            "NonStable":"True"
        },
        "Mass": 5
    },
    {
        "Name": "Grip Pad",
        "Description": "The block with the highest friction.",
        "Type ID": 49,
        "Size": [0.8, 0.8, 0.5],
        "Mass": 0.3
    },
    {
        "Name": "Elastic Pad",
        "Description": "The block with the highest elasticity.",
        "Type ID": 87,
        "Size": [0.8, 0.8, 0.2],
        "Mass": 0.3
    },
    {
        "Name": "Container",
        "Description": "It has a railing around the building point. If oriented towards +y, it can hold sub-blocks like a bowl. It is mainly used to hold blocks that cannot directly connect to the mechanism, such as boulders and bombs. Do not place other blocks nearby to avoid overlap.",
        "Type ID": 30,
        "Size": [2.4, 3, 2.8],
        "Attachable Faces Properties": [
            {"ID": 0, "Coordinates": [0, 0, 1], "Orientation": "Front"}
        ],
        "Mass": 0.5
    },
    {
        "Name": "Suspension",
        "Description": "It primarily serves as a buffer and shock absorber. It is similar in shape to a wooden block, with all Attachable Faces Properties located at the far end of the block.",
        "Type ID": 16,
        "Size": [1, 1, 2],
        "Attachable Faces Properties": [
            {"ID": 0, "Coordinates": [0, 0, 2], "Orientation": "Front"},
            {"ID": 1, "Coordinates": [-0.5, 0, 1.5], "Orientation": "Left"},
            {"ID": 2, "Coordinates": [0.5, 0, 1.5], "Orientation": "Right"},
            {"ID": 3, "Coordinates": [0, 0.5, 1.5], "Orientation": "Up"},
            {"ID": 4, "Coordinates": [0, -0.5, 1.5], "Orientation": "Down"}
        ],
        "Mass": 0.5
    },
    {
        "Name": "Brace",
        "Description": "The brace can be used for reinforcement. Its construction principle is to 'attach' to other blocks. It has no collision volume. Since it is often used to stabilize static structures, it is not suitable for any dynamic structures.",
        "Type ID": 7,
        "Special Attributes": {
            "Linear": "True",
            "Anti Tension Direction": "Towards the center of the line segment between the two Attachable Faces Properties",
            "Anti-Compression Direction": "Outward from the center of the line segment between the two Attachable Faces Properties"
        },
        "Mass": 0.5
    },
    {
        "Name": "Spring",
        "Description": "A special block that attaches to two other blocks and can quickly pull the two ends together. Its tension force is almost entirely dependent on its length.",
        "Type ID": 9,
        "Special Attributes": {
            "Linear": "True",
            "NonStatic":"True",
            "Tension Direction": "Towards the center of the line segment between the two Attachable Faces Properties"
        },
        "Mass": 0.4
    },
    {
        "Name": "Ballast",
        "Description": "It serves as a counterweight, has a large mass, and is shaped like a cube.",
        "Type ID": 35,
        "Size": [1, 1, 1],
        "Attachable Faces Properties": [
            {"ID": 0, "Coordinates": [0, 0, 1], "Orientation": "Front"},
            {"ID": 1, "Coordinates": [-0.5, 0, 0.5], "Orientation": "Left"},
            {"ID": 2, "Coordinates": [0.5, 0, 0.5], "Orientation": "Right"},
            {"ID": 3, "Coordinates": [0, 0.5, 0.5], "Orientation": "Up"},
            {"ID": 4, "Coordinates": [0, -0.5, 0.5], "Orientation": "Down"}
        ],
        "Mass": 3
    }
]
\end{lstlisting}

\end{document}